\documentclass{article}

\usepackage[utf8]{inputenc}
\usepackage[T1]{fontenc}
\usepackage{microtype}
\usepackage{graphicx}
\usepackage{xcolor}
\usepackage{nicefrac}
\usepackage{booktabs}
\usepackage{url}
\usepackage{caption}
\usepackage{subfigure}
\usepackage{tikz}
\usepackage{amsmath,amssymb,amsfonts,amsthm}
\usepackage{mathtools}
\usepackage{placeins}
\usepackage{makecell}
\usepackage{todonotes}
\usepackage{abbreviations}
\usepackage{tikz-3dplot} % 마지막에!

\usepackage{hyperref}
\usepackage[capitalize,noabbrev]{cleveref}

\usepackage[capitalize,noabbrev]{cleveref}
\Crefname{assumption}{Assumption}{Assumptions}
\Crefname{claim}{Claim}{Claims}
\Crefname{equation}{Eq.}{Equations}

\usepackage[accepted]{icml2025}
\FloatBarrier

\icmltitlerunning{Linear Bandits with Partially Observable Features}

\begin{document}

\twocolumn[
\icmltitle{Linear Bandits with Partially Observable Features}



\icmlsetsymbol{equal}{*}

\begin{icmlauthorlist}
\icmlauthor{Wonyoung Kim}{equal,cau}
\icmlauthor{Sungwoo Park}{equal,gsds}
\icmlauthor{Garud Iyengar}{columbia}
\icmlauthor{Assaf Zeevi}{columbia}
\icmlauthor{Min-hwan Oh}{gsds}
\end{icmlauthorlist}

\icmlaffiliation{gsds}{Seoul National University, Seoul, Korea}
\icmlaffiliation{cau}{Chung-Ang University, Seoul, Korea}
\icmlaffiliation{columbia}{Columbia University, New York, USA}
% \icmlaffiliation{columbia2}{Columbia Business School, Columbia University, New York, USA}

\icmlcorrespondingauthor{Min-hwan Oh}{minoh@snu.ac.kr}

\icmlkeywords{Linear Bandits, Partially Observable Features, Doubly Robust, Machine Learning, ICML}

\vskip 0.3in
]



\printAffiliationsAndNotice{\icmlEqualContribution} % otherwise use the standard text.

\begin{abstract}
We study the linear bandit problem that accounts for partially observable features. Without proper handling, unobserved features can lead to linear regret in the decision horizon $T$, as their influence on rewards is unknown.
To tackle this challenge, we propose a novel theoretical framework and an algorithm with sublinear regret guarantees.
The core of our algorithm consists of (i) feature augmentation, by appending basis vectors that are orthogonal to the row space of the observed features; and (ii) the introduction of a doubly robust estimator.
Our approach achieves a regret bound of $\tilde{O}(\sqrt{(d + d_h)T})$, where $d$ is the dimension of the observed features and $d_h$ depends on the extent to which the unobserved feature space is contained in the observed one, thereby capturing the intrinsic difficulty of the problem.
Notably, our algorithm requires no prior knowledge of the unobserved feature space, which may expand as more features become hidden.
Numerical experiments confirm that our algorithm outperforms both non-contextual multi-armed bandits and linear bandit algorithms depending solely on observed features.
\end{abstract}

% \begin{table*}[ht]
%     \centering
%     \begin{tabular}{l l}
%     \toprule
%     Feature space & Regret bound \\
%     \midrule
%     span(observed features) $\supseteq$ span(latent features) & $ \tilde{O}(\sqrt{dT}) $ \\
%     span(observed features) $\subseteq$ span(latent features) & $\tilde{O}(\sqrt{KT})$ \\
%     otherwise & $\tilde{O}(\sqrt{(d+d_h)T})$\\
%     \bottomrule
%     \end{tabular}
%     \caption{An overview of regret bound range of our algorithm, \texttt{RoLF}, depending on $d_h \in [0,K-d]$, the dimension of the vector space spanned by the rows of the matrix of unobserved features influencing the reward defined in \cref{eq:d_h}.
%     %Our algorithm incurs regret adaptive to $d_h$, and the regret bound does not exceed that of multi-armed bandit algorithms leveraging UCB, in the worst case.
%     Note that $\tilde{O}$ denotes the big-O notation omitting logarithm factors.}
%     \label{tab:table1}
% \end{table*}

\section{Introduction}
\label{sec:intro}
We consider a linear bandit problem where the learning agent has access to only a \emph{subset} of the features, while the reward is determined using the \emph{complete set} of features, including both observed and unobserved elements.
% ~\footnote{In this paper, we use the terms ``unobserved'' and ``latent'' interchangeably to refer to features that are not visible to the decision-making agent.}.
Conventional linear bandit problems rely on the assumption that the rewards are linear in only observed features, without accounting for the potential presence of unobserved features. 
However, in many real-world applications, rewards are often affected by the unobserved---hence \textit{latent}---features that are not observable by the agent.
For example, in online advertising without personalization---i.e., when serving general users---each advertisement is associated not only with observable features such as its category, format, or display time, but also with unobservable factors such as emotional appeal, creative design quality, and brand familiarity.
These latent factors influence users’ click-through rates, yet they are not directly quantifiable.
Similarly, in clinical trials, treatment outcomes depend not only on observable features like dosage and formulation, but also on unobserved factors such as manufacturing variability and potential side effect risks.
Hence, accurately incorporating latent features is essential for providing precise recommendations.

To address latent features, ~\citet{park2022regret,park2024thompson,kim2023contextual,zeng2024partially} typically assume that the true features follow a specific distribution, such as a Gaussian distribution.
Establishing a sublinear regret bound in the decision horizon without such structural assumptions on the latent features remains a significant challenge and has not been accomplished yet.
Key challenges in the bandit problem with partially observable features arise from the complete lack of information about the latent features---indeed, the agent does not even know whether the features are partially observed.
We tackle these challenges by proposing a novel linear bandit algorithm that is agnostic to partial observability.
Despite having no knowledge of the unobserved features, our algorithm achieves a tighter regret bound than both linear bandit algorithms that consider only observed features and multi-armed bandit (MAB) algorithms that ignore features entirely. 
For brevity, we will refer to linear bandit algorithms relying solely on observed features as ``OLB algorithms'' henceforth.
Specifically, our proposed algorithm achieves a $\sqrt{T}$-rate regret bound, without requiring any prior knowledge of the unobserved features, where $T$ is the decision horizon. 
%and 

The key idea of our proposed algorithm can be summarized in the following two procedures:
(i)~reconstructing feature vectors via feature augmentation to capture the influence of unobserved features on rewards, and (ii) introducing a novel doubly robust (DR) estimator to mitigate information loss due to partial observability.
For (i), we decompose the reward into two additive components: one projected onto the row space spanned by the observed features, and the other projected onto its orthogonal complement.
The former term maximally captures the effect of observed features, while the latter minimizes the impact of unobserved features.
We then \emph{augment} observed features with an orthonormal basis from the complementary space, which is orthogonal to the row space of observed features.
This formulation enables us to reformulate the problem within a conventional linear bandit framework, where the reward function is defined as a dot product of the augmented features and an unknown parameter, without any additional additive term.

However, since these augmented features are not identical to the unobserved features, estimation errors may arise from information loss.
To mitigate such errors, we leverage (ii) a DR estimator, which is widely used in the statistical literature for its robustness to errors caused by missing data.
These two approaches allow our algorithm to compensate for missing information due to partial observability, improving both estimation accuracy and adaptability to the environment.

Our main contributions are summarized as follows:
\vspace{-8pt}
\begin{itemize}
    \item We propose a linear bandit problem with partially observable features.
    Our problem setting is more general and challenging than those in the existing literature on linear bandits with latent features, which often rely on specific structural assumptions governing the relationship between observed and latent features.
    In contrast, our approach assumes no additional structure for the unobserved features beyond the linearity of the reward function, which is commonly adopted in the linear bandit literature~(\cref{sec:preliminaries}).
      
    \item We introduce a novel estimation strategy by (i) augmenting the features that maximally capture the effect of reward projected onto the observed features, while minimizing the impact of unobserved features~(\cref{sec:method}), and (ii) constructing a DR estimator that mitigates errors from unobserved features.
    By integrating augmented features with the DR estimator, we guarantee a $\sqrt{t}$-convergence rate on the rewards for \emph{all} arms in each round $t$~(\cref{thm:est_lasso_consistency}).

    \item We propose an algorithm named \textit{Robust to Latent Features} (\texttt{RoLF}) for a general linear bandit framework with latent features~(\cref{alg:RoLF}).
    The algorithm achieves a regret bound of $\tilde{O}(\sqrt{(d + d_h)T})$\footnote{$\tilde{O}(\cdot)$ is the Big-O notation omitting logarithmic factors.} (\cref{thm:lasso_regret}), where $d_h$ is the number of nonzero coefficients associated with the component of the reward projected onto the orthogonal complement of the row space of observed features~(\cref{subsec:def_dr_lasso}). 
    \texttt{RoLF} requires no prior knowledge or modeling of unobserved features, yet achieves, to the best of our knowledge, a sharper regret bound than OLB and MAB algorithms, as well as existing methods accounting for partial observability or model misspecification within the linear bandit framework.
    
    \item Our numerical experiments confirm that our algorithm consistently outperforms OLB~\citep{li2010contextual-bandit,agrawal2013thompson,kim2019doubly-robust} and MAB~\citep{lattimore2020bandit} algorithms, validating both its practicality and theoretical guarantees~(\cref{sec:experiment}).
\end{itemize}

\section{Related Works}
\label{sec:related_work_part}
While our setting appears similar to prior works on bandit problems with (i) model misspecification and (ii) partial observability, it differs from both lines of research in several aspects, including the nature of the unobserved features and the strategies used to address the problem.

First, our problem setting is more general and challenging than misspecified linear bandit problems, where the reward model deviates from the true reward due to non-linearity~\citep{lattimore2020bandit} or additive deviation terms~\citep{ghosh2017misspecified,bogunovic2021stochastic,he2022nearly}. 
While prior works incorporate cumulative misspecification error into the regret bound, we obtain a regret bound
% a sublinear regret in $T$ 
that remains unaffected by such deviations.
In particular,~\citet{ghosh2017misspecified} employ hypothesis testing for model selection and obtain a regret bound of $O(K\sqrt{T}\log T)$ under high misspecification.
In contrast, our algorithm achieves a regret bound of $O(\sqrt{(d + d_h)T \log T})$ without such tests, while handling partial observability in a unified framework.

Second, in bandit problems with partially observable features, prior works often rely on structural assumptions. 
For example,~\citet{park2022regret,park2024thompson,kim2023contextual} assume the true features are drawn from a Gaussian distribution, while~\citet{zeng2024partially} model them as evolving according to a linear dynamical system with additive Gaussian noise, where observed features are generated via a known linear mapping, also corrupted by Gaussian noise.
These methods typically aim to recover the true features---via decoders~\citep{park2022regret,park2024thompson}, Bayesian oracles~\citep{kim2023contextual}, or Kalman filtering~\citep{zeng2024partially}.
In contrast, our approach makes no structural assumptions about the relationship between observed and unobserved features and does not attempt to recover the latter.
Instead, we mitigate the information loss from partial observability by projecting the latent part of the reward using only the observed features.
A more detailed and comprehensive literature review is provided in~\cref{appendix:related_works}.

\section{Preliminaries}
\label{sec:preliminaries}

\subsection{Notation}
\label{subsec:notation}
For any $n \in \NN$, let $[n]$ denote the set $\{1, 2, \dots, n\}$. 
For a vector $\vb$, we denote its $L_1$, $L_2$ and supremum norms by $\norm{\vb}_1$, $\norm{\vb}_2$, and $\norm{\vb}_\infty$, respectively.
The $L_2$-norm weighted by a positive definite matrix $\Db$ is denoted by $\norm{\vb}_{\Db}$. 
For two vectors $\vb_1$ and $\vb_2$, the inner product is defined as the dot product, i.e., $\dotp{\vb_1}{\vb_2} := \vb_1^\top \vb_2$, and we use both notations interchangeably.
For a matrix $\Mb$, its minimum and maximum eigenvalues are denoted by $\lambda_{\min}(\Mb)$ and $\lambda_{\max}(\Mb)$, respectively.
We let $\R(\Mb)$ denote the row space of $\Mb$, i.e., the subspace spanned by the rows of $\Mb$.

\subsection{Problem Formulation}
\label{subsec:problem_formulation}
In this section, we outline our problem setting and introduce several key assumptions.
Each arm $a\in[K]$ is associated with a true feature vector $\zb_a \in \RR^{d_z}$ that determines its rewards. 
However, the agent can observe only a subset of its elements, with the others remaining unobserved.
Specifically, $\zb_a$ is defined as follows:
\begin{equation}
    \label{eq:z_decompose}
    \zb_a := {\left[\textcolor{blue}{x_{a}^{(1)}}, \cdots, \textcolor{blue}{x_{a}^{(d)}},  \textcolor{red}{u_a^{(1)}}, \cdots, \textcolor{red}{u_a^{(d_u)}}\right]}^{\top}.
\end{equation}

For clarity, we highlight the observed components in blue and the unobserved components in red.
Note that the dimensions of the latent feature vector, $d_u = d_z - d$, and the true feature vector, $d_z$, are both \emph{unknown} to the agent.
As a result, the agent is unaware of how many features, if any, are unobserved, or even whether partial observability exists at all.
This lack of structural information presents a fundamental challenge, as it prevents the agent from explicitly modeling or compensating for the unobserved components when selecting an appropriate learning strategy.

It is worth noting that the setting with fixed observed features\footnote{This assumption is standard in linear bandits with model misspecification \citep{ghosh2017misspecified,lattimore2020learning}, which is a special case of our partially observable feature setting.} includes linear bandits with misspecification error \citep{ghosh2017misspecified,bogunovic2021stochastic,he2022nearly} as special cases.
In \cref{appendix:time_varying}, we present a setting with varying observed features and an algorithm that achieves a $\sqrt{T}$-rate regret bound.
Moreover, if latent features were allowed to change arbitrarily over time, the problem would become non-learnable and thus ill-posed. 
Consequently, assuming fixed features is both natural and well-justified (see \cref{tab:learnability_results} for comparisons).

The reward associated with arm $a$ is defined as the dot product between its true feature vector $\zb_a$ and an unknown parameter $\thetab_{\star} \in \RR^{d_z}$, namely 
\begin{equation}
    \label{eq:reward_definition}
 y_{a,t} = \dotp{\zb_{a}}{\thetab_{\star}} + \epsilon_{t}\qquad \forall\ a\in[K],
\end{equation}
where $a_t\in[K]$ denotes the action selected by the agent at round $t$.
Here, $\epsilon_t$ represents a random noise term that captures the inherent stochasticity in the reward generation process. 
We make the following assumption on $\epsilon_t$, which is commonly adopted in bandit problems. 

% To insert
\begin{table}[ht]
\vspace{-0.3cm}
    \centering
    \caption{Summary of problem settings covered in this paper and the corresponding results. Note that if latent features arbitrarily change over time, the problem itself would become non-learnable, making the problem ill-posed~(see \cref{appendix:time_varying} for details).}
    % \vspace{0.1cm}
    \begin{tabular}{cccc}
      \toprule
        \makecell{\textbf{Observed} \\ \textbf{Features}} & \makecell{\textbf{Unobserved} \\ \textbf{Features}} & \textbf{Learnable?} & \textbf{Results} \\
        % \textbf{Observed Features} & \textbf{Unobserved Features} & \textbf{Learnable?} & \textbf{Results} \\
    \midrule
    Fixed & Fixed & Yes & \cref{thm:lasso_regret} \\
        Varying & Fixed & Yes & \cref{thm:regret_ridge_time_varying} \\
        Varying & Varying & No & - \\
        \bottomrule
    \end{tabular}
    \vspace{0.3cm} 
    \label{tab:learnability_results}
\end{table}
% \vspace{0.3cm}

\begin{assumption}[Sub-Gaussian noise]
\label{asm:sub-gaussian}
Let $\Fcal_{t}$ denote the history at round $t$, represented by a filtration of $\sigma$-algebras, e.g., $\sigma(a_1, y_{a_1,1},\ldots, a_{t-1},y_{a_{t-1}, t-1}, a_t)$.
The reward noise $\epsilon_t$ is conditionally $\sigma$-sub-Gaussian, i.e., for all $\lambda \in \RR$,
\begin{equation*}
    \condexp{\exp(\lambda\epsilon_t)}{\Fcal_{t-1}} \le \exp\left(\frac{\lambda^2\sigma^2}{2}\right).
\end{equation*}
\vspace{-0.6cm}
\end{assumption}
Since $\epsilon_t$ is sampled after $a_t$ is observed, $\epsilon_t$ is $\Fcal_t$-measurable.
Under~\cref{asm:sub-gaussian}, it follows that $\condexp{\epsilon_t}{\Fcal_{t-1}} = 0$, thus $\condexp{y_{a,t}}{\Fcal_{t-1}} = \dotp{\zb_a}{\thetab_{\star}}$.
For brevity, we use $\EE_{t-1}[\cdot]$ to denote $\condexp{\cdot}{\Fcal_{t-1}}$ henceforth.
To eliminate issues of scale for analysis, we assume that the expected reward $|\langle \zb_{a}, \thetab_{\star} \rangle| \le 1$ for all $a\in[K]$, and we do not assume any bound on the norm of $\zb_{a}$ as well.

Let $a_\star := \argmax_{a \in [K]} \dotp{\zb_a}{\thetab_{\star}}$ denote the optimal action, considering both observed and latent features. Moreover, we denote by $\zb_\star$ and $y_{\star,t}$ the true feature vector and the realized reward associated with the optimal action $a_\star$, respectively.
We evaluate the theoretical performance of our algorithm using cumulative regret, which is defined as the expected sum of the differences between the reward of $a_\star$ and $a_t$: 
\begin{align*}
    \regret(T) 
    &= \sum_{t=1}^T \dotp{\thetab_{\star}}{ \zb_\star - \zb_{a_t}} \\
    &= \sum_{t=1}^T \left(\EE_{t-1}[y_{\star,t}] - \EE_{t-1}[y_{a_t, t}]\right)
\label{eq:regret_decomposition}.
\end{align*}
Considering the composition of $\zb_a$ defined in~\cref{eq:z_decompose}, we decompose the parameter vector as $\thetab_{\star} = [({\thetao})^\top,({\thetau})^\top]^\top$, where $\thetao\in\RR^d$ and $\thetau \in \RR^{d_u}$ correspond to the parameters associated with the observed and latent features, respectively.
Adopting this decomposition of $\thetab_\star$, the reward $y_{a_t,t}$ defined in~\cref{eq:reward_definition} can be decomposed into the following three terms: 
\begin{equation}
    y_{a_t,t} = \dotp{\obs{t}}{\thetao} + \epsilon_t + \dotp{\ub_{a_t}}{\thetau}.
    \label{eq:reward_decomposition}
\end{equation}
The last term of~\cref{eq:reward_decomposition} corresponds to the inaccessible portion of the reward.
This reward model is equivalent to that imposed in the linear bandits with misspecification error~\citep{ghosh2017misspecified,lattimore2020learning}.
While the regret bound in~\citet{lattimore2020learning} includes misspecification error that grows linearly with decision horizon, our proposed method (\cref{sec:method}) addresses this misspecification error and achieves a regret bound that is sublinear in $T$.

%\begin{align}
%    \regret(T) 
%    &= \EE\left[\sum_{t=1}^T \left(y_{a_\star, t} - y_{a_t,t}\right)\right]\label{eq:def_regret} \\
%    &= \EE\left[\sum_{t=1}^T \dotp{{\obs{\star}} - {\obs{t}}}{\thetao} + \sum_{t=1}^T \dotp{\ub_{a_\star}- \ub_{a_t}}{\thetau}\right] \label{eq:regret_decomposition}.
%\end{align}
Before presenting our method and algorithm, we first establish a lower bound on the regret of linear bandit algorithms that disregard the unobserved portion of the rewards.
In particular, the following theorem provides lower bounds for two OLB algorithms:~\texttt{OFUL}~\citep{abbasi-yadkori2011improved}~and~\texttt{LinTS}~\citep{agrawal2013thompson}.

\begin{theorem}[Regret lower bound of \texttt{OFUL} and \texttt{LinTS} ignoring latent features]
\label{thm:regret_linear_lower_bound}
Under partial observability, there exists a problem instance where both \texttt{OFUL} and \texttt{LinTS} suffer from cumulative expected regret that grows linearly with $T$ due to their disregard for the unobserved components.
\end{theorem}

\begin{sketchproof}
\vspace{-0.25cm}
Consider a linear bandit problem instance with an action set $\mathcal{A} := \{1,2\}$, where $d = d_u = 1$, implying that $d_z = 2$.
The true feature vectors are given by $\mathcal{Z} := \{[1,3]^\top, [2,19/4]^\top\} \subset \RR^{2}$, with the agent observing only the first element of each vector, while the second element remains unobserved.
In this setting, we take arm 2 to be the optimal action.
However, an estimator relying solely on the observed features is inconsistent, leading \texttt{OFUL} and \texttt{LinTS} to select the suboptimal arm with probability $\Theta(1)$, and thereby incur regret that grows linearly in $T$.
\vspace{-0.25cm}
\end{sketchproof}

\cref{thm:regret_linear_lower_bound} shows that neglecting the latent portion of the reward in decision-making may cause the agent to fail in identifying the optimal action.
The comprehensive proof is deferred to~\cref{appendix:proof_thm1}.
While \cref{thm:regret_linear_lower_bound} specifically considers \texttt{OFUL} and \texttt{LinTS}---which are known to achieve the most efficient regret bounds for UCB- and Thompson sampling-based policies in the linear bandit framework---we additionally present an algorithm-agnostic lower bound based on a different analysis (see~\cref{appendix:regret_lower_bound_general} for details).

\section{Robust Estimation for Partially Observable Features}
\label{sec:method}
% We propose our estimation method to obtain a sublinear regret bound for linear bandits with latent features.
% \cref{subsec:feature_augmentation} introduces the feature vector augmentation to mitigate the information loss incurred by partial observability.
% To further improve the regret bound, we present the DR estimation for a Lasso estimator in~\cref{subsec:def_dr_lasso}.

\subsection{Feature Vector Augmentation with Orthogonal Projection}
\label{subsec:feature_augmentation}
% A stochastic $K$-armed linear bandit problem can be viewed\todo{Added the three following paragraphs.} as a linear regression problem---under conditions such as realizability and sufficient exploration~\citep{foster2018practical,foster2020ucb,simchi-levi2022bypassing}---where 
In the linear bandit framework, accurate estimation of the unknown reward parameter $\thetab_\star$ contributes to optimal decision-making.
However, in our problem setting, the learner lacks access to the latent reward component---namely, the last term in~\cref{eq:reward_decomposition}.
Consequently, without an appropriate mechanism to compensate for the absence of $\thetau$, the agent fails to accurately estimate $\thetab_\star$, resulting in ineffective learning, as demonstrated in~\cref{thm:regret_linear_lower_bound}.

That said, minimizing regret---that is, selecting the optimal arm---does not require recovery of the true reward parameter $\thetab_{\star}$ entirely; rather, it suffices to estimate the $K$ expected rewards $\{ \zb_{a}^\top\thetab_{\star} : a \in [K] \}$.
A straightforward approach is to ignore the features altogether and apply MAB algorithms such as \texttt{UCB1}~\citep{auer2002finite-time}, which are known to achieve a regret bound of $\tilde{O}(\sqrt{KT})$.
However, these algorithms tend to suffer higher regret than feature-based algorithms, particularly when the number of arms is significantly larger than the dimension of the feature vectors, i.e., $K \gg d_z$.
% and $d_u = 0$ (indicating full observability of the features).
% The challenge arises from the fact that we do not know whether latent features exist, which makes it difficult to choose between MAB algorithms and linear bandit algorithms.

To tackle this dilemma, we propose a unified approach that handles partial observability and ensures efficient estimation.
Let us define $\Xb := ( \xb_1, \dots, \xb_K ) \in \RR^{d \times K}$ as the matrix formed by concatenating the observed part of the true features, and $\Ub := (\ub_1^{(u)}, \ldots, \ub_K^{(u)}) \in \RR^{d_u \times K}$ as the matrix formed by concatenating their latent complements for each arm.
Without loss of generality, we assume a set of $K$ vectors $\{\xb_1, \ldots, \xb_K\}$ spans $\RR^d$.\footnote{When $d > K$, we can apply singular value decomposition (SVD) on $\Xb$ to reduce the feature dimension to $\bar{d} \le K$ with $\R(\Xb)=\bar{d}$.}
We denote by $\Pb_{\Xb} := \Xb^\top (\Xb \Xb^\top)^{-1} \Xb$ the projection matrix onto the row space of $\Xb$, denoted by $\R(\Xb)$.
Then the vector of rewards for all arms, $\Yb_t= (y_{1,t}, \ldots, y_{K,t})^\top$, is decomposed as:
\begin{align}
\Yb_t &= \left(\Xb^\top \thetao + \Ub^{\top} \thetau\right) + \epsilon_t \bsym{1}_K \notag  \\
&= \Pb_{\Xb} \left(\Xb^\top \thetao + \Ub^{\top} \thetau\right) \notag \\
&\quad+ (\Ib_K-\Pb_{\Xb}) \left(\Xb^\top \thetao + \Ub^{\top} \thetau\right) + \epsilon_t \bsym{1}_K \notag \\
&= \Xb^\top \left(\thetao + (\Xb\Xb^\top )^{-1}\Xb \Ub^{\top} \thetau \right) \notag \\
&\quad+ (\Ib_K-\Pb_{\Xb}) \Ub^{\top} \thetau + \epsilon_t \bsym{1}_K, \label{eq:reward_decomposition_expanded_matrix}
\end{align}
where the first term corresponds to the reward projected onto $\R(\Xb)$, whereas the second term is the reward projected onto $\R(\Xb)^{\perp}$, the subspace of $\RR^{K}$ orthogonal to $\R(\Xb)$.
We reparametrize $\thetao + (\Xb\Xb^\top )^{-1}\Xb \Ub^{\top} \thetau$ as ${\mub_{\star}^{(o)}}$, representing the parameter associated with the observed features.

To handle the second term in~\cref{eq:reward_decomposition_expanded_matrix}, we consider a set of row vector bases $\{\bb_1^\top, \ldots, \bb_{K-d}^\top\} \in \R(\Xb)^{\perp}$, where~$\bb_i\in\RR^K$ for $i\in[K-d]$.
Given the set, there exist coefficients ${\mu_{\star,1}^{(u)}},\dots,{\mu_{\star,K-d}^{(u)}}\in \RR$ such that the term can be expressed as a linear combination:
\begin{equation}
\label{eq:unobs_reward_projection}
    (\Ib_K-\Pb_{\Xb})\Ub^{\top} \thetau = \sum_{i=1}^{K-d} \mu_{\star,i}^{(u)} \bb_i.
\end{equation}
We define the number of nonzero coefficients as:
\begin{equation}
d_h(\bb_{1}^\top,\ldots,\bb_{K-d}^\top):=|\{i\in[K-d]:\mu_{\star,i}^{(u)}\neq0\}|.
\label{eq:d_h}
\end{equation}
Note that $d_h = 0$ for any basis set~$\{\bb_{1}^\top,\ldots,\bb_{K-d}^\top\}$ when the latent feature space is completely included in the observed feature space, i.e., $\R(\Ub) \subseteq \R(\Xb)$. 
In this case, $(\Ib_K - \Pb_{\Xb})\Ub^\top = \mathbf{0}_{K\times d_u}$, which means that the projected rewards onto $\mathrm{R}(\Xb)^\perp$ can be linearly expressed by the observed features.
If $\R(\Ub) \supseteq \R(\Xb)$, on the other hand, then $d_h=K-d$ for any basis set $\{\bb_{1}^\top,\ldots,\bb_{K-d}^\top\}$.

In other cases, the quantity $d_h$ depends on the choice of the basis $\{\bb_1^\top,\ldots,\bb_{K-d}^\top\}$, and tends to be smaller when a larger portion of $\R(\Ub)$ is included within $\R(\Xb)$.
For any choice of the basis, our algorithm achieves $\tilde{O}(\sqrt{(d+d_h)T})$ regret without prior knowledge of $d_h$, which does not exceed the $\tilde{O}(\sqrt{KT})$ regret bound achieved by MAB algorithms ignoring features.
Further details are provided in~\cref{subsec:analysis}.

Similar to the reparametrization of $\mub_\star^{(o)}$, we denote $\mub_\star^{(u)}:=[\mu_{\star,1}^{(u)}, \ldots, \mu_{\star,K-d}^{(u)}]^\top$.
By concatenating $\mub_\star^{(o)}$ and $\mub_\star^{(u)}$, we define $\mub_\star := [({\mub_{\star}^{(o)}})^\top, ({\mub_{\star}^{(u)}})^\top]^\top \in \RR^{K}$, so that the reward for each $a \in [K]$ can be rewritten as follows:
\begin{align}
y_{a,t} &= \eb_a^\top \Yb \notag \\ 
&= \eb_a^{\top} [\Xb^\top \; \bb_1 \cdots \bb_{K-d}] \mub_{\star} + \epsilon_t \notag \\
&= [{\xb_{a}} \; \eb_a^{\top}\bb_1 \cdots \eb_a^{\top}\bb_{K-d}] \mub_{\star} + \epsilon_t. 
\label{eq:reward_decomposition_expanded}
\end{align}
The decomposition in~\cref{eq:reward_decomposition_expanded} implies that~\cref{eq:reward_decomposition_expanded_matrix} takes the form $\Yb_t = [\Xb^\top \; \bb_1 \cdots \bb_{K-d}] \mub_{\star} + \epsilon_t \bsym{1}_K $.
Note that $\eb_a \in \RR^K$ denotes the $a$-th standard basis vector.
With this modification, the rewards are now represented as a linear function of the augmented feature vectors:~$\tilde{\xb}_a := [\xb_a^\top \; \eb_a^{\top}\bb_1 \cdots \eb_a^{\top}\bb_{K-d}]^{\top} \in \RR^{K}$, \emph{without any misspecification error}.
A toy example illustrating our strategy is shown in~\cref{fig:method}.

\begin{figure}[t]
\vspace{-0.2cm}
\centering
\caption{Illustration comparing OLB algorithms and our approach in estimating rewards of $K=3$ arms.
OLB algorithms find estimates within $\R(\Xb)$ thus accumulating errors from unobserved features.
However, our approach utilizes the projection of the latent reward, $\bb^\top\widehat{\mub}_t^{(u)}$, onto the orthogonal complement of $\R(\Xb)$, thereby enabling reward estimation in $\RR^K$.
% Note that $\widehat{\mub}_t$ is the estimator of the parameter for observed features.
}
\label{fig:method}
\tdplotsetmaincoords{60}{120}
\subfigure[OLB algorithms: Estimation confined to $\mathrm{R}(\Xb)$]
{
    \begin{tikzpicture}[tdplot_main_coords, scale=0.65]
        \tikzset{font=\fontsize{8}{10}\selectfont}
        \draw[thick,->] (0,0,0) -- (3,0,0) node[anchor=north]{$\qb_1$};
        \draw[thick,->] (0,0,0) -- (0,3,0);
        \draw[thick,->] (0,0,0) -- (0,0,3) node[anchor=south]{$\qb_3$};
        \draw[thick,->] (0,0,0) -- (1,2,3) node[anchor=south]{$\EE_{t-1}[\Yb]$};
        \draw[thick,->] (0,0,0) -- (2,1,0);
        \node[anchor=south east] at (1.5, 3.5, 0){$\mathrm{R}(\Xb)$};
        \node[anchor=north] at (0,3.3,0.2){$\qb_2$};
        \node[anchor=west] at (2.7,0.5,-0.2){$\mathbf{X}^\top \widehat{\mub}_t^{(o)}$};
    \end{tikzpicture}
}
\hfill
\subfigure[Ours: Projection onto orthogonal complement]
{
    \begin{tikzpicture}[tdplot_main_coords, scale=0.65]
        \tikzset{font=\fontsize{8}{10}\selectfont}
        \draw[thick,->] (0,0,0) -- (3,0,0) node[anchor=north]{$\qb_1$};
        \draw[thick,->] (0,0,0) -- (0,3,0);
        \draw[thick,->] (0,0,0) -- (0,0,3) node[anchor=south]{$\qb_3$};
        \draw[thick,->] (0,0,0) -- (1,2,3) node[anchor=south]{$\EE_{t-1}[\Yb]$};
        \draw[thick,->] (0,0,0) -- (1,2,0);
        \draw[thick,->] (1,2,3) -- (1,2,0);
        \node[anchor=north] at (0,3.3,0.2){$\qb_2$};
        \node[anchor=south east] at (2.4, 2, 0){$\mathrm{R}(\Xb)$};
        \node[anchor=south west] at (1, 2, 1.5){$\bb^\top \widehat{\mub}_t^{(u)}$};
        \node[anchor=west] at (1.6,1.5,-0.2){$\mathbf{X}^\top \widehat{\mub}_t^{(o)}$};
    \end{tikzpicture}
}
% \vspace{0.1cm}
\end{figure}

In terms of regret minimization, while \texttt{SupLinUCB} \citep{chu2011contextual} achieves a regret bound of $\tilde{O}(\sqrt{dT})$, it is often considered impractical as it computes $\log T$ distinct batches and estimators, requiring knowledge of $T$ and, more critically, discards a significant portion of samples at each parameter update. 
To retain the theoretical efficiency while improving practicality, we adopt the doubly robust (DR) estimation framework, which is known to achieve $\tilde{O}(\sqrt{dT})$~\citep{kim2021doubly,kim2023squeeze}.
Given that $\tilde{\xb}_a \in \RR^{K}$, we propose an efficient algorithm that employs a DR estimator with ridge regularization and obtains a regret bound of $\tilde{O}(\sqrt{KT})$ (see \cref{appendix:dr_ridge}).
However, when $K>d$ and $d_u=0$, the regret is the regret is higher than that of OLB algorithms.
To address this challenge, we propose a novel estimation strategy in the following section that eliminates the dependence on $K$ in the regret bound.

%Note that $\mub_{\star}$ has at most $d + d_h$ non-zero entries.
%\subsection{Doubly Robust Pseudo Rewards with Resampling and Coupling}
%\label{subsec:dr_explained}
%In this section, our goal is to introduce a UCB-based algorithm tailored for linear bandits that achieves a regret bound capped at $\otil(\sqrt{(d+K)T})$. 
%Here, the inclusion of the $K$ term reflects the algorithm's consideration of unique factors associated with each arm. 
%Drawing from the principles of established linear UCB algorithms like \texttt{OFUL}~\citep{abbasiyadkori2011}, we can develop an algorithm with a regret bound of $\otil((d+K)\sqrt{T})$. 
%However, we hypothesize that such algorithms may not be optimal, particularly in scenarios where $K\gg d$, resulting in an inflated regret.

\subsection{Doubly Robust Lasso Estimator}
\label{subsec:def_dr_lasso}
In~\cref{eq:reward_decomposition_expanded}, the parameter $\mub_{\star}$ is sparse, with its sparsity depending on the number of nonzero elements required to represent the inaccessible portion of the reward projected onto $\R(\Xb)^{\perp}$, namely $d_h$ defined in~\cref{eq:d_h}.
Recall from~\cref{eq:unobs_reward_projection} that this projection can be expressed using at most $d_h$ basis vectors, implying that $\mub_{\star}^{(u)}$ has at most $d_h$ nonzero entries.

Let $\check{\mub}^{L}_t$ denote the Lasso estimator of $\mub_{\star}$ using the augmented feature vectors, defined as follows:
\begin{equation}
\begin{split}
\check{\mub}^{L}_{t}&:= \argmin_{\mub} \sum_{\tau=1}^{t} (y_{a_\tau,\tau}-\tilde{\xb}_{a_{\tau}}^{\top}\mub )^2 \\
&\qquad+ 2 \tilde{\sigma}_{\max} \sigma \sqrt{2pt \log \frac{2Kt^2}{\delta}} \|\mub\|_1,
\end{split}
\label{eq:lasso_impute}
\end{equation}
where $p$ is the coupling probability used to define the multinomial distribution for pseudo-action sampling (as defined in~\cref{eq:pseudo_distribution}), and $\delta$ is the confidence parameter of the algorithm.
Note that $\tilde{\sigma}^2_{\max}:=\max_{a\in[K]} \eb_a^\top (\sum_{a\in[K]} \tilde{\xb}_{a}\tilde{\xb}_{a}^\top)\eb_a$, i.e., the largest diagonal entry of the Gram matrix constructed from the augmented feature vectors $ \tilde{\xb}_a$ over $\Acal$.

For the estimator in~\cref{eq:lasso_impute} to correctly identify the zero entries in $\mub_{\star}$, the compatibility condition needs to be satisfied~\citep{geer2009conditions}.
Although the compatibility condition does not, in general, require a positive minimum eigenvalue for the Gram matrix, in our setting, $\lambda_{\min}(t^{-1}\sum_{s=1}^{t}\tilde{\xb}_{a_s}\tilde{\xb}_{a_s}^\top) > 0$. 
Therefore, the compatibility condition is implied without any additional assumption.
However, ensuring a sufficiently large minimum eigenvalue typically requires collecting a large number of exploration samples, which in turn increases regret.
Achieving this with fewer exploration samples remains a key challenge in the bandit literature, as the minimum eigenvalue affects the convergence rate of the estimator and, consequently, the regret bound~\citep{soare2014best-arm,kim2021doubly}.

We introduce a doubly robust (DR) estimator that employs the Gram matrix constructed from the augmented feature vectors over the entire action space, $\sum_{s=1}^t\sum_{a\in[K]}\tilde{\xb}_a\tilde{\xb}_a^\top$, rather than only from the chosen actions, $\sum_{s=1}^t \tilde{\xb}_{a_s}\tilde{\xb}_{a_s}^\top$.
Originating in the missing-data literature~\citep{bang2005doubly}, DR estimation yields consistent estimators if either the imputation or the observation probability model is correct~\citep{kim2021doubly}.
In the bandit setting, only the reward of the chosen arm is observed for each decision round, leaving the other $K-1$ as \emph{missing}.
Hence, DR estimation imputes these $K-1$ missing rewards and incorporates all the associated feature vectors into the estimation process.
Since the observation probabilities are determined by the policy---which is known to the learner---the DR estimator remains robust to errors in the reward estimation.
While~\citet{kim2019doubly-robust}~proposed a DR Lasso estimator under IID features satisfying the compatibility condition, we propose a novel DR Lasso estimator that does not rely on such assumptions.

We improve the DR estimation by incorporating resampling and coupling methods.
For each $t$, let $\Ecal_t \subseteq [t]$ denote the set of exploration rounds such that for any $\tau\in\Ecal_t$, the action $a_{\tau}$ is sampled uniformly over $[K]$.
The set $\Ecal_t$ is constructed recursively: starting with $\Ecal_0 = \emptyset$, we define 
\begin{equation*}
    % \label{eq:def_exploration}
    \Ecal_t =
    \begin{cases}
        \Ecal_{t-1} \cup \{t\} & \mathrm{if}\quad |\Ecal_t| \le C_e \log\dfrac{2Kt^2}{\delta}, \\
        \Ecal_{t-1} &\mathrm{otherwise}.
    \end{cases}
\vspace{-0.2cm}
\end{equation*}

Here, $C_e$ is defined as $(8K)^3 \tilde{\sigma}_{\min}^{-2} \tilde{\sigma}_{\max}^2 (1 - p)^{-2}$, where $\tilde{\sigma}_{\min}^2 := \lambda_{\min}(\sum_{a \in [K]} \tilde{\xb}_a \tilde{\xb}_a^\top)$ denotes the minimum eigenvalue of the Gram matrix constructed from the augmented feature vectors over the \emph{entire} action space.

At each round $t \notin \Ecal_t$, the algorithm selects $a_t$ according to an $\epsilon_t$-greedy strategy: with probability $\epsilon_t = t^{-1/2}$, it selects $\widehat{a}_t = \argmax_{a \in [K]} \tilde{\xb}_a^\top \widehat{\mub}_{t-1}^{L}$; with $1 - \epsilon_t$, it selects an action uniformly at random from $[K]\setminus\{ \widehat{a}_t\}$.
Given $a_t$, a \emph{pseudo-action} $\tilde{a}_t$ is sampled from a multinomial distribution:
\begin{equation}
\begin{split}
&\phi_{a_t,t} := \PP(\tilde{a}_t=a_t\mid a_t) = p,\\
&\phi_{k,t} := \PP(\tilde{a}_t=k\mid a_t) = \frac{1-p}{K-1},
\end{split}
\label{eq:pseudo_distribution}
\end{equation}
for all $k\in[K] \setminus \{a_t\}$, where $p\in(1/2,1)$ is the coupling probability specified by the algorithm.
To couple the policies for the actual action $a_t$ and the pseudo-action $\tilde{a}_t$, we repeatedly resample both of them until they match.

With the pseudo-actions coupled with the actual actions, we construct unbiased pseudo-rewards for all $a\in[K]$ as:
\begin{equation}
\tilde{y}_{a,t}:={\tilde{\xb}_a}^{\top}\check{\mub}_t^L+\frac{\II(\tilde{a}_{t}=a)}{\phi_{a,t}}\big(y_{a,t}-\tilde{\xb}_a^{\top}\check{\mub}_t^L\big),
\label{eq:pseudoY}
\end{equation}
where $\check{\mub}_t^L$ is the imputation estimator that fills in the missing rewards of unselected arms at round $t$, as defined in~\cref{eq:lasso_impute}.

As shown in~\cref{eq:pseudoY}, the pseudo-reward includes an inverse probability term. 
Hence, when DR estimation is applied under the $\epsilon_t$-greedy policy with $\epsilon_t = t^{-1/2}$, its variance can grow unbounded over time.
To mitigate this issue, we propose a resampling and coupling strategy: by coupling the $\epsilon_t$-greedy policy with the multinomial distribution~(\cref{eq:pseudo_distribution}), we ensure that each inverse probability weight $\phi_{a,t}^{-1}$, for $a\in[K]$, remains bounded by $O(K)$.
This strategy effectively stabilizes the variance of the pseudo-rewards.

Moreover, one can show that the resampling succeeds with high probability for each round. 
Let $\Mcal_{t}$ denote the event that the pseudo-action $\tilde{a}_t$ matches the chosen action $a_t$ within a specified number of resamples. 
For a given $\delta^\prime\in(0,1)$, we set the maximum number of resamples to $\rho_{t}:=\log((t+1)^{2}/\delta^\prime)/\log(1/(1-p))$; then $\Mcal_{t}$ occurs with probability at least $1-\delta^\prime/(t+1)^{2}$.
Resampling allows the algorithm to further explore the action space to balance regret minimization with accurate reward estimation.
% This coupling yields a lower bound on the observation probability, which in turn reduces the variance of the DR pseudo-rewards in \cref{eq:pseudoY}.
% Let $\Mcal_{t}$ denote an event where $\tilde{a}_t= a_t$ within a specified number of resamples. 
% For a given $\delta^\prime\in(0,1)$, we set the maximum number of resamples to $\rho_{t}:=\log((t+1)^{2}/\delta^\prime)/\log(1/p)$ ensuring that $\Mcal_{t}$ occurs with probability at least $1-\delta^\prime/(t+1)^{2}$.
% This resampling allows the algorithm to further explore the action space in order to balance regret minimization with accurate reward estimation.
% % By the coupling strategy, $\epsilon_t$-greedy policy is replaced with a multinomial distribution over the arms, denoted by~$\phi_{1,t} ,\ldots, \phi_{K,t}$. 
% When the DR estimation is employed under the $\epsilon_t$-greedy policy, the pseudo-reward~(\cref{eq:pseudoY}) involves inverse probability term $\epsilon_t^{-1} := \sqrt{t}$, which can cause the variance of the pseudo-reward to grow unbounded over time.
% To mitigate this issue, we couple the $\epsilon_t$-greedy policy with the multinomial distribution~\cref{eq:pseudo_distribution}, which ensures that each inverse probability weight $\phi_{a,t}^{-1}$, for $a\in[K]$, satisfies $O(K)$.

% and is defined based on the main estimator used to estimate $\mub_{\star}$. 

For $a \neq \tilde{a}_t$, i.e., an arm that is \emph{not} selected in the round $t$, we impute the missing rewards using $\tilde{\mathbf{x}}_k^\top \check{\mub}_t^L$.
For $a=\tilde{a}_t$, in contrast, the term $\II(\tilde{a}_t=a)y_{a,t}/\phi_{a,t}$ calibrates the predicted reward to ensure the unbiasedness of the pseudo-rewards for all arms.
Given that $\EE_{\tilde{a}_t}[\II(\tilde{a}_t = a)] = \PP(\tilde{a}_t = a) = \phi_{a,t}$, taking the expectation over $\tilde{a}_t$ on both sides of~\cref{eq:pseudoY} gives $\EE_{\tilde{a}_t}[\tilde{y}_{a,t}]= \EE_{t-1}[y_{a,t}] = \tilde{\xb}_a^{\top}{{\mub}_{\star}}$ for all $a\in [K]$.
Although the estimate $\tilde{\xb}_a^\top\check{\mub}_t$ may have a large error, it is multiplied by the mean-zero random variable $(1-\II(\tilde{a}_t=a)/\phi_{a,t})$, which makes the resulting pseudo-rewards defined in~\cref{eq:pseudoY} robust to errors of $\tilde{\xb}_a^\top\check{\mub}_t$.
The pseudo-rewards can only be computed---and thus can only be used---when the pseudo-action $\tilde{a}_t$ matches the actual action $a_t$, that is, when the event $\Mcal_t$ occurs. 
Since $\Mcal_t$ occurs with high probability, we are able to compute pseudo-rewards for almost all rounds.

Incorporating the pseudo-rewards, $\tilde{y}_{a,t}$, into estimation, we define our DR Lasso estimator as follows:
\begin{equation}
\begin{split}
\widehat{\mub}^{L}_t &:= \argmin_{\mub}  \sum_{\tau=1}^{t}\II (\Mcal_{\tau})\sum_{a\in[K]} \left(\tilde{y}_{a,\tau}-\tilde{\xb}_a^\top {\mub} \right)^2 \\ 
&\qquad + \frac{4\sigma \tilde{\sigma}_{\max}}{p} \sqrt{2t\log\frac{2Kt^{2}}{\delta}}\|\mub\|_1,
\end{split}
\label{eq:lasso_main}
\end{equation}
and note that our estimator uses the Gram matrix aggregated over all actions at each round, $\sum_{\tau=1}^{t} \sum_{a=1}^{K} \tilde{\xb}_{a} \tilde{\xb}_{a}^\top$, instead of the one built only from the chosen actions, $\sum_{\tau=1}^{t} \tilde{\xb}_{a_\tau} \tilde{\xb}_{a_\tau}^\top$.
The following theorem guarantees convergence of the estimator, across all arms, to the true parameter, after a sufficient number of exploration rounds.

\begin{theorem}[Consistency of the DR Lasso estimator]
\label{thm:est_lasso_consistency}
Let $d_h$ denote the dimension of the projected latent rewards defined in \cref{eq:d_h}.
Then with probability at least $1-3\delta$,
\begin{equation*}
\max_{a\in[K]}\vert \tilde{\xb}_{a}^{\top}(\widehat{\mub}^{L}_t - \mub_{\star})\vert\le\frac{8\sigma \tilde{\sigma}_{\max}}{p\tilde{\sigma}_{\min}} \sqrt{\frac{2(d+d_{h})}{t}\log\frac{2Kt^{2}}{\delta}},
\label{eq:lasso_est_consistency}
\end{equation*}
for all rounds $t$ such that $t\ge |\Ecal_t|$.
\end{theorem}

Although the DR Lasso estimator leverages $K$-dimensional feature vectors, its error bound depends only logarithmically on $K$. 
Such fast convergence is typically achieved under classical regularity conditions, including the compatibility condition and the restrictive minimum eigenvalue condition~\citep{geer2009conditions,bhlmann2011statistics}.
Existing Lasso-based bandit approaches 
\citep{kim2019doubly-robust,bastani2020online,oh2021sparsity-agnostic,ariu2022thresholded,chakraborty2023thompson,lee2025lasso} generally impose these conditions directly on the feature vectors.
Our approach, on the other hand, does not require such assumptions, as the augmented features are orthogonal vectors lying in $\R(\Xb)^\perp$.
Moreover, their average Gram matrix satisfies $\lambda_{\min}(\sum_{a\in[K]}\tilde{\xb}_a\tilde{\xb}_a^{\top})\ge \min\{1,\lambda_{\min}(\sum_{a\in[K]}\xb_a\xb_a^{\top} )\}$. 
Thus, the convergence rate scales only as $\sqrt{\log K}$ with respect to $K$.
The consistency is proved by bounding the two components of the error in the pseudo-rewards~(\cref{eq:pseudoY}): (i) the noise of the reward and (ii) the error of the imputation estimator $\check{\mub}_t$. 
Since (i) is sub-Gaussian, it can be bounded using martingale concentration inequalities. 
For (ii), the imputation error $\tilde{\xb}_a^\top (\check{\mub}^{L}_t - \mub_{\star} )$ is multiplied by the mean-zero random variable $\left(1-{\II(\tilde{a}_t=a)}/{\phi_{a,t}}\right)$ and thus the magnitude of the error can be bounded by $\|\check{\mub}^L_t - \mub_{\star}\|_1/\sqrt{t}$.
The detailed proof is deferred to~\cref{appendix:proof_thm2}.

%------------------------------------
% Algorithm
%------------------------------------
\begin{algorithm}[t]
    \caption{Robust to Latent Feature~(\texttt{RoLF})}
    \label{alg:RoLF}
\begin{algorithmic}[1]
    \STATE \textbf{INPUT:} Observed features $\{\xb_a: a\in[K]\}$, coupling probability $p\in(1/2,1)$, confidence parameter $\delta>0$.
    \STATE Initialize $\widehat{\mub}_0 = \mathbf{0}_K$, exploration phase $\Ecal_0=\emptyset$, and exploration factor $C_{\mathrm{e}}:=(8K)^3 \tilde{\sigma}^{-2}_{\min} \tilde{\sigma}_{\max}^2 (1-p)^{-2}$.
    \STATE Find orthogonal basis $\{\bb_1^\top,\ldots,\bb_{K-d}^\top\}\subseteq\mathrm{R}(\Xb)^{\perp}$ and construct $\{\tilde{\xb}_a:a\in[K]\}$.
    \FOR{$t=1,\ldots,T$}
    \IF{$|\Ecal_t| \le  C_{\mathrm{e}} \log (2Kt^2/\delta)$}
    \STATE Randomly sample $\widehat{a}_t$ uniformly over $[K]$ and $\Ecal_t = \Ecal_{t-1}\cup \{t\}$.
    \ELSE
    \STATE Compute $\widehat{a}_t := \arg\max_{a\in[K]}\tilde{\xb}_a^\top \widehat{\mub}^{L}_{t-1}$.
    \ENDIF
    \WHILE{$\tilde{a}_t \neq a_t$ and $\text{count} \leq \rho_t$}
    \STATE Sample $a_t$ with $\PP(a_t = \widehat{a}_t)=1-(t^{-1/2})$ and $\PP(a_t =k)=t^{-1/2}/(K-1),\; \forall k\neq\widehat{a}_t$.
    \STATE Sample $\tilde{a}_t$ according to ~\cref{eq:pseudo_distribution}.
    \STATE $\text{count} = \text{count}+ 1$.
    \ENDWHILE
    %\ENDIF
    \STATE Play $a_t$ and observe $y_{a_t,t}$.
    \IF{$\tilde{a}_t \neq a_t$}
    \STATE Set $\widehat{\mub}^{L}_{t}:=\widehat{\mub}^{L}_{t-1}$.
    \ELSE
    % \STATE Compute $\widehat{\mub}^{R}_t$ in~\cref{eq:def_ridge_DR} ($\widehat{\mub}^{L}_t$ in~\cref{eq:lasso_main} for Lasso) with pseudo-rewards $\tilde{y}_{a,t}$ in~\cref{eq:pseudoY} and the imputation estimator $\check{\mub}^{R}_t$ in~\cref{eq:ridge_impute} ($\check{\mub}^{L}_t$ in~\cref{eq:lasso_impute} for Lasso).
    \STATE Update $\widehat{\mub}^{L}_t$ following~\cref{eq:lasso_main} with $\tilde{y}_{a,t}$ and update $\check{\mub}^{L}_t$ following~\cref{eq:lasso_impute}.
    \ENDIF
    \ENDFOR
   % \REPEAT
   % \STATE Initialize $noChange = true$.
   % \FOR{$i=1$ {\bfseries to} $m-1$}
   % \IF{$x_i > x_{i+1}$}
   % \STATE Swap $x_i$ and $x_{i+1}$
   % \STATE $noChange = false$
   % \ENDIF
   % \ENDFOR
   % \UNTIL{$noChange$ is $true$}
\end{algorithmic}
\end{algorithm}
%------------------------------------
% Algorithm
%------------------------------------
% \vspace{-0.5cm}
\section{Proposed Algorithm and Regret Analysis}

\label{sec:algorithm}
% The details of~\cref{alg:RoLF} are presented in~\cref{subsec:algorithm}, and the upper bound on the regret incurred by the algorithm is provided in~\cref{subsec:analysis}.

\subsection{Robust to Latent Features (\texttt{RoLF}) Algorithm}
\label{subsec:algorithm}
\begin{table}[ht]
\vspace{-0.3cm}
    \centering
    \caption{An overview of regret bound range of our algorithm, \texttt{RoLF}, depending on $d_h \in [0,K-d]$, the number of nonzero elements to effectively capture the unobserved part of reward projected onto $R(\Xb)^\perp$.
    %Our algorithm incurs regret adaptive to $d_h$, and the regret bound does not exceed that of multi-armed bandit algorithms leveraging UCB, in the worst case.
    Note that $\tilde{O}$ denotes the big-O notation omitting logarithm factors.}
    \vspace{0.1cm}
    \begin{tabular}{c c}
    \toprule
    \textbf{Feature space} & \textbf{Regret bound} \\
    \midrule
    span(observed) $\supseteq$ span(latent) & $ \tilde{O}(\sqrt{dT}) $ \\
    span(observed) $\subseteq$ span(latent) & $\tilde{O}(\sqrt{KT})$ \\
    otherwise & $\tilde{O}(\sqrt{(d+d_h)T})$\\
    \bottomrule
    \end{tabular}
    \vspace{0.2cm} 
    \label{tab:regret_overview}
\end{table}

In this section, we present our algorithm \texttt{RoLF}  in~\cref{alg:RoLF}.
In the initialization step, given the observed features, our algorithm constructs a set of orthogonal basis vectors $\{\bb_1^\top, \dots, \bb_{K-d}^\top\}\subseteq\R(\Xb)^{\perp}$ to augment each observed feature vector.
The algorithm first selects a candidate action $\widehat{a}_t$---either uniformly at random or greedily based on the estimated reward---and then resamples $a_t$ and $\tilde{a}_t$ until they match or the maximum number of trials allowed for each round, $\rho_{t}:=\log((t+1)^{2}/\delta^\prime)/\log(1/(1-p))$, is reached.
Once the resampling phase ends, the algorithm selects $a_t$ and the corresponding reward $y_{a_t, t}$ is observed.
If $a_t$ and $\tilde{a}_t$ match within $\rho_{t}$, then both the imputation and the main estimators are updated; otherwise, neither is updated.

Our proposed algorithm does not require knowledge of the dimension of the unobserved features $d_u$, nor the dimension of the reward component projected onto $\R(\Xb)^\perp$.
Although we present the algorithm for fixed feature vectors, the algorithm is also applicable to time-varying feature vectors.
In such cases, we estimate the bias caused by unobserved features by augmenting the standard basis. 
For further details, refer to~\cref{appendix:time_varying}.
\vspace{-0.2cm}

\subsection{Regret Analysis}
\label{subsec:analysis}
% We analyze the regret bound of \texttt{RoLF} using the Lasso estimator defined in~\cref{eq:lasso_main}, as stated in the following theorem:
\begin{theorem}[Regret bound for Lasso \texttt{RoLF}]
\label{thm:lasso_regret}
Let $d_h$ denote the number of nonzero coefficients in the representation of the projected latent reward as defined in~\cref{eq:d_h}. 
Then for any $\delta\in(0,1)$ and $p\in(1/2,1)$, with probability at least $1-6\delta$, the cumulative regret of \texttt{RoLF} is bounded by
\begin{equation*}
\label{eq:regret_lasso}
    \begin{split}
\regret(T) 
& \le 2\cdot 8^3 K^3(1-p)^{-2}\log\frac{2KT^{2}}{\delta} \\ 
&\quad + \frac{4\sqrt{T}}{K-1}+ 2\sqrt{2T \log \frac{2}{\delta}}+4\delta \\
&\quad +\frac{32\sigma \tilde{\sigma}_{\max}}{p \tilde{\sigma}_{\min}} \sqrt{2(d+d_{h})T\log\frac{2KT^{2}}{\delta}}.
    \end{split}
\end{equation*}
\end{theorem}
%The comprehensive proofs for the respective theorems are provided in \cref{appendix:proof_thm4} and \cref{appendix:proof_thm5}.

To the best of our knowledge,~\cref{thm:lasso_regret} provides the first regret bound that converges faster than $\tilde{O}(\sqrt{KT})$, specifically for algorithms that account for unobserved features without relying on any structural assumptions.
Assuming $\norm{\tilde{\xb}_{a}}_{\infty} \le 1$ (instead of $\norm{\tilde{\xb}_{a}}_{2} \le 1$), $\tilde{\sigma}^2_{\min}$ and $\tilde{\sigma}_{\max}^2$ are constant independent of $d$ or $K$. 
Note that the number of rounds required for the exploration phase is $O(K^3 \log KT)$, which grows only logarithmically with the time horizon $T$.
The factor $K^3$ is irreducible, as the algorithm needs information about all $K$ biases from the missing features.
By employing the Gram matrix of the \emph{augmented} feature vectors over the action space $\Acal$, $\sum_{a=1}^{K}\tilde{\xb}_{a}\tilde{\xb}_{a}^\top$, when combined with DR estimation, we reduce the required exploration phase time from $O(K^4 \log KT)$ to $O(K^3 \log KT)$, thereby improving the sample complexity by a factor of $K$.

As demonstrated in~\cref{thm:est_lasso_consistency}, the last term on the right-hand side of the regret bound is proportional to $\sqrt{d + d_h}$ rather than $\sqrt{K}$, resulting in an overall regret bound of $O(\sqrt{(d+d_h)T\log KT})$.
The specific value of $d_h$ is determined by the relationship between $\mathrm{R}(\Xb)$ and $\mathrm{R}(\Ub)$, as discussed in~\cref{subsec:feature_augmentation}.
\cref{tab:regret_overview} summarizes how the regret bound varies under different relationships between these subspaces.
The formal proof of~\cref{thm:lasso_regret} is deferred to~\cref{appendix:proof_thm3}.

\section{Numerical Experiments}
\label{sec:experiment}
\subsection{Experimental Setup}
In this experiment, we simulate and compare our algorithms,~\cref{alg:RoLF} and~\cref{alg:RoLF_ridge} (\cref{appendix:dr_ridge}), with baseline OLB algorithms: \texttt{LinUCB}~\citep{li2010contextual-bandit} and \texttt{LinTS}~\citep{agrawal2013thompson}. 
These algorithms adopt UCB and Thompson sampling strategies, respectively, assuming a linear reward model based on observed features.
We also include \texttt{DRLasso}~\citep{kim2019doubly-robust} since our algorithm employs DR estimation with a Lasso estimator, and \texttt{UCB($\delta$)}~\citep{lattimore2020bandit}, an MAB algorithm that ignores features, to assess the effectiveness of using feature information, even under partial observability.

% The detailed experimental environment, including exact values of all variables, is provided in~\cref{appendix:exp_setup}.
% For the simulation environment, we generate the true features $\zb_a$ for each arm $a \in [K]$ from $\mathcal{N}(\mathbf{0}, \mathbf{I}_{d_z})$ and construct the observed features $\xb_a$ by subsampling $d$ elements from $\zb_a$. 
% Orthogonal basis vectors $\{\mathbf{b}_1^\top, \dots, \mathbf{b}_{K-d}^\top\}$ are derived via singular value decomposition (SVD) on the observed feature matrix $\mathbf{X}$, ensuring orthogonality to $\R(\mathbf{X})$.
% These basis vectors are linearly concatenated to $\Xb$ to form the augmented feature matrix. 
% The reward parameter $\thetab_\star \in \mathbb{R}^{d_z}$ is sampled from the uniform distribution $\mathrm{Unif}(-1/2, 1/2)$, and the rewards are generated following the definition~\cref{eq:reward_definition}.
% Finally, we set the coupling probability $p = 0.6$ (see~\cref{eq:pseudo_distribution}), the confidence parameter $\delta = 10^{-4}$, and the decision horizon $T = 1200$.

% We set the coupling probability $p$, a hyperparameter used in the sampling distribution of $\tilde{a}_t$, is set to $0.6$ (see~\cref{eq:pseudo_distribution}). 
% The confidence parameter $\delta$, which is also a hyperparameter, is set to $10^{-4}$, and the total decision horizon is $T = 1200$. 

To provide comprehensive results, we consider two scenarios: one with partial observability and one without.
For each scenario, we conduct experiments under three cases, classified by the relationship between the row spaces of the observed and unobserved features, $\mathrm{R}(\Xb)$ and $\mathrm{R}(\Ub)$, which determines the value of $d_h$:
(i) Case 1, the general case where neither $\mathrm{R}(\Xb)$ nor $\mathrm{R}(\Ub)$ fully contains the other;
(ii) Case 2, $\mathrm{R}(\Ub) \subseteq \mathrm{R}(\Xb)$, where the row space spanned by the unobserved features lies entirely within that spanned by the observed features, thus $d_h = 0$;
(iii) Case 3, $\mathrm{R}(\Xb) \subseteq \mathrm{R}(\Ub)$, where the row space spanned by the observed feature space is fully contained within that spanned by the unobserved features, implying $d_h = K - d$.
Note that in Scenario 2, Case 3 is excluded because $\R(\Ub) = \emptyset$ implies $\R(\Xb) = \emptyset$, which violates our problem setup.

In the simulation, after constructing the true features $\zb_a\in\RR^{d_z}$ and observed features $\xb_a\in\RR^{d}$, we apply singular value decomposition (SVD) to the observed feature matrix $\Xb$ to derive orthogonal basis vectors $\{\mathbf{b}_1^\top, \dots, \mathbf{b}_{K-d}^\top\}$ orthogonal to $\R(\Xb)$.
These vectors are then concatenated with $\Xb$ to form the augmented feature matrix. 
The reward parameter $\thetab_\star \in \mathbb{R}^{d_z}$ is sampled from $\mathrm{Unif}(-1/2, 1/2)$, and the rewards are generated following~\cref{eq:reward_definition}.
For the hyperparamters in our algorithms, the coupling probability $p$ and the confidence parameter $\delta$, are set to $0.6$ and $10^{-4}$, respectively.
The total decision horizon is $T = 1200$. 
Throughout the experiments, we fix the number of arms at $K = 30$ and the dimension of the true features at $d_z = 35$, ensuring $d_z \geq d$ to cover both scenarios.
Further details on the experimental setup are provided in~\cref{appendix:exp_setup}.
\vspace{-0.1cm}

\subsection{Experimental Results}

\begin{figure*}[t]
\setcounter{subfigure}{0} 
    \centering     %%% not \center
    \subfigure[Case 1 (General Case)]
    {\label{fig:1a}\includegraphics[width=0.32\linewidth]{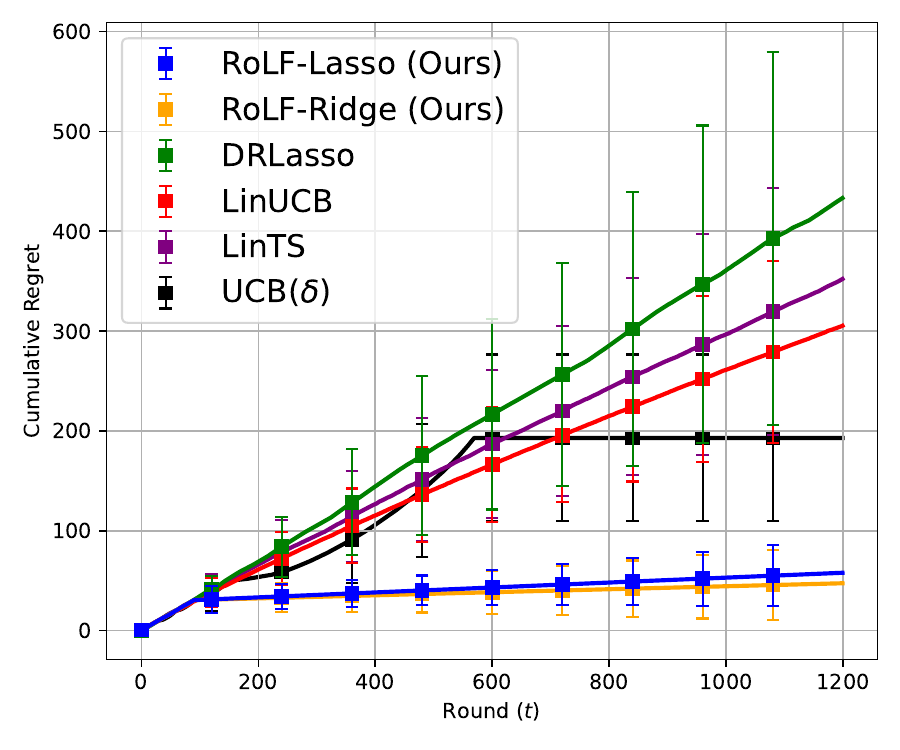}}
    \subfigure[Case 2 ($\mathrm{R}(\Ub) \subseteq \mathrm{R}(\Xb)$)]
    {\label{fig:1b}\includegraphics[width=0.32\linewidth]
    {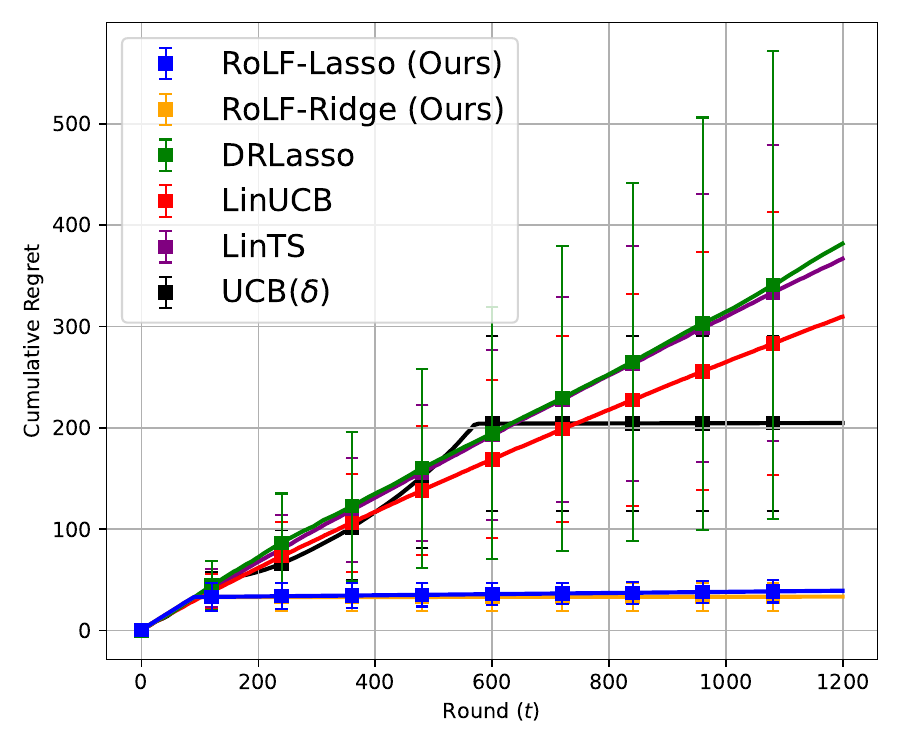}}
    \subfigure[Case 3 ($\mathrm{R}(\Xb) \subseteq \mathrm{R}(\Ub)$)]
    {\label{fig:1c}\includegraphics[width=0.32\linewidth]
{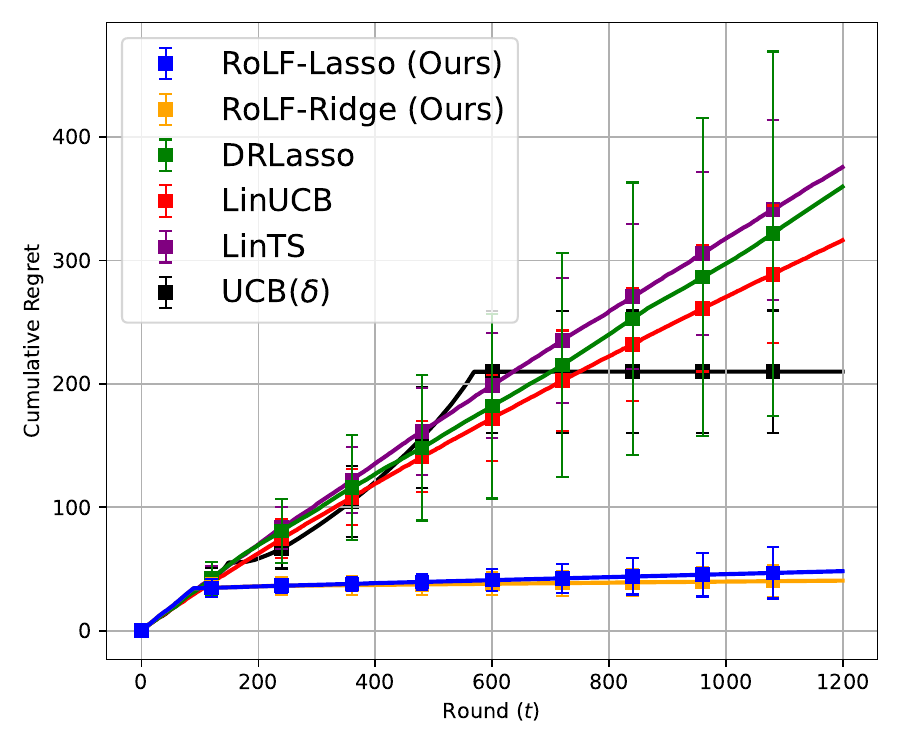}}
    \caption{Cumulative regrets of the algorithms with partial observability (Scenario 1).
    % The proposed \texttt{RoLF} algorithm performs better on the partially observable features on various dimension of observable feature vectors.
    }
        \label{fig:result1}
\end{figure*}

\begin{figure*}[t]
    \centering     %%% not \center
    \subfigure[Case 1 (General Case)]
        {\label{fig:2a}\includegraphics[width=0.32\linewidth]
        {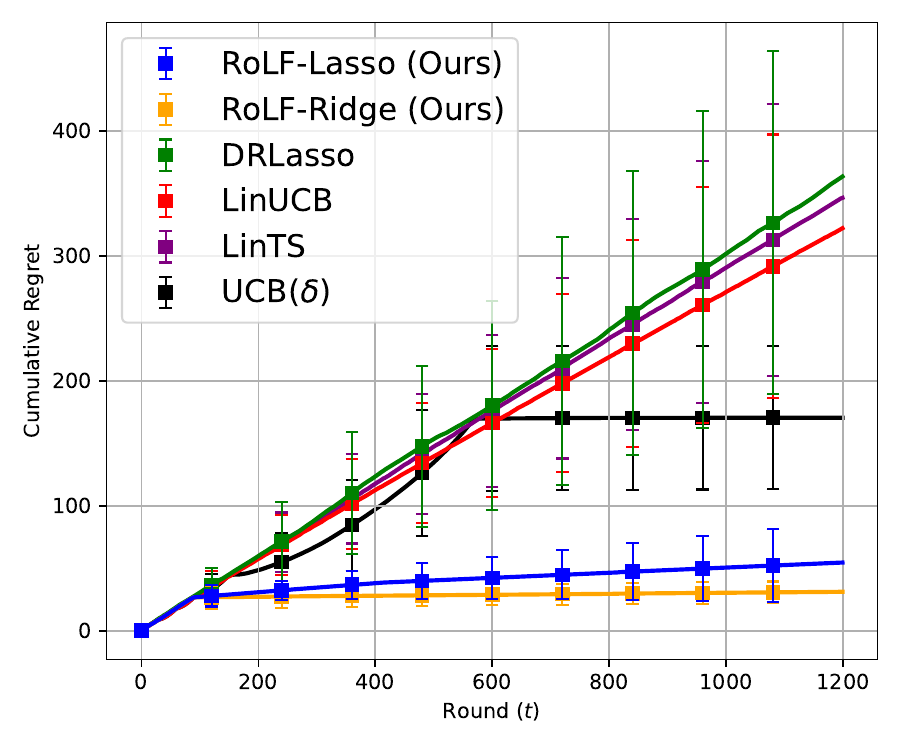}}
    \subfigure[Case 2 ($\mathrm{R}(\Ub) \subseteq \mathrm{R}(\Xb)$)]
        {\label{fig:2b}\includegraphics[width=0.32\linewidth]
        {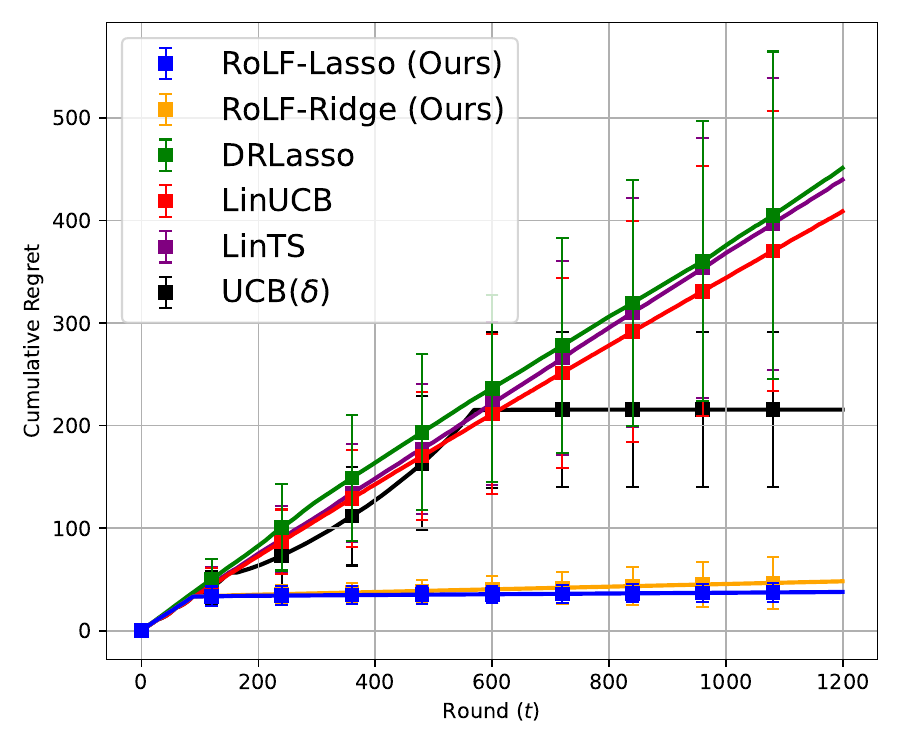}}
    \caption{Cumulative regrets of the algorithms without partial observability (Scenario 2).
    % The proposed \texttt{RoLF} algorithm performs better on the partially observable features on various dimension of observable feature vectors.
    }
        \label{fig:result2}
\end{figure*}

\subsubsection{Scenario 1}
In this scenario, $d$ is set to $\lfloor d_z / 2 \rfloor$, so that the agent observes only about half of the full feature dimension.
From~\cref{fig:result1}, we observe that the baseline OLB algorithms---\texttt{LinUCB}, \texttt{LinTS}, and \texttt{DRLasso}---perform worse than our algorithms in terms of both the level and robustness of cumulative regret.
This pattern is consistently observed across all three cases, implying that linear bandit algorithms fail to identify the optimal arm in the presence of unobserved components, as they cannot capture the portion of the reward associated with the unobserved features.
However, our algorithms show almost the same performance regardless of the relationship between $\mathrm{R}(\Xb)$ and $\mathrm{R}(\Ub)$, indicating the robustness to variations in feature observability structure.

Note that for all cases our algorithms---\texttt{RoLF-Lasso}~(\cref{alg:RoLF}) and \texttt{RoLF-Ridge}~(\cref{alg:RoLF_ridge})---exhibit a sharp decline in the growth rate of cumulative regret after a certain number of rounds.
This behavior is primarily due to the forced-exploration phase built into the algorithms, which ensures sufficient coverage of the action space in the early stages. 
Following this phase, the resampling and coupling strategies further enhance the efficiency of the DR estimation, leading to slower regret accumulation over time.
This pattern is commonly shown from experimental results of other bandit algorithms employing the forced-exploration strategy~\citep{goldenshluger2013linear,hao2020high-dimensional,chakraborty2023thompson}.

Furthermore, in Case 2~(\cref{fig:1b}), the cumulative regret grows more slowly than in other cases.
This behavior is explained by the relationship $\mathrm{R}(\Ub)\subseteq \mathrm{R}(\Xb)$, implying the reward components projected onto $\mathrm{R}(\Xb)^\perp$ can be fully expressed by the observed features.
As a result, the augmented features fully capture the underlying reward structure, enabling faster convergence relative to the other scenarios.
\vspace{-0.1cm}
\subsubsection{Scenario 2}
In Scenario 2, we set $d = d_z = 2K$, implying that no latent features remain and allowing us to evaluate our algorithms under the condition $d > K$.
As discussed in~\cref{subsec:feature_augmentation}, we apply SVD to reduce the dimensionality of the observed features before constructing the augmented feature set.
\cref{fig:result2}~shows that our algorithms perform well even without unobserved features, both in terms of the level and robustness of the cumulative regret.
Moreover, by incorporating dimensionality reduction, our algorithms remain effective even when the feature matrix is not full rank. 
Lastly, similar to~\cref{fig:result1}, cumulative regret in Case 2 converges slightly more slowly than in Case 1.

Meanwhile, the baseline OLB algorithms continue to struggle in identifying the optimal arm throughout the horizon.
For \texttt{LinUCB} and \texttt{LinTS}, this behavior may be attributed to the curse of dimensionality, where regret scales linearly with the feature dimension $d$---a well-known limitation of linear bandit algorithms~\citep{oswal2020linear,tran2024symmetric}.
Additionally, since the true rewards are bounded by 1, the resulting small reward gaps between arms may further hinder the identification of the optimal arm.
In the case of \texttt{DRLasso}, similar behavior arises from the fixed feature setting: the algorithm uses an ``averaged'' context vector across the action space at each round for parameter estimation.
Consequently, under the fixed-feature setting, this strategy becomes effectively equivalent to using a single fixed vector throughout the entire learning horizon, thereby limiting the expressiveness of the estimation.

\section{Conclusion}
\label{sec:conclusion}
In this work, we addressed the problem of partially observable features within the linear bandit framework.
We showed that conventional algorithms that ignore unobserved features may suffer linear regret due to information loss, and introduced \texttt{RoLF}, a novel algorithm that accounts for latent features using only observed data without requiring prior knowledge.
Our algorithm achieves a tighter regret bound than existing methods, and this improvement is supported by our numerical experiments.

For future work, from the perspective that our feature augmentation strategy reformulates the problem as another linear bandit problem without model misspecification, extending this approach to other reward models, e.g., generalized linear models, would be an interesting direction.
Additionally, although we assumed the latent reward component to be linear in the unobserved features, we can relax this assumption by viewing the latent reward component as an exogenous factor that interferes with the learning process.
This perspective allows for modeling the latent reward using a general function class without structural assumptions.

% % The outcomes in \cref{fig:result1} reveal that both versions of our algorithm, the Lasso and the Ridge, outperform in terms of regret and robustness. 
% % Specifically, we see that conventional linear bandit algorithms show significant dependence on the number of observable elements. 
% % As fewer features become observable, the performances of \texttt{LinUCB} and \texttt{LinTS} decline significantly and exhibit high variability across different environments. 
% % However, all three graphs in \cref{fig:result1} demonstrate that our algorithm remains robust regardless of the number of observable elements. 
% % In fact, our algorithm’s cumulative regret, both in terms of its absolute value and variability, is minimally affected by the number of observable features.

%%%%%%%%%%%%%%%%%%%%%%%%%%%%%%%%%%%%%%%%%%%%%%%%%%%%%%%%%%%%%%%%%%%%%%%%%%%%%%%
%%%%%%%%%%%%%%%%%%%%%%%%%%%%%%%%%%%%%%%%%%%%%%%%%%%%%%%%%%%%%%%%%%%%%%%%%%%%%%%
% Code of Conduct
%%%%%%%%%%%%%%%%%%%%%%%%%%%%%%%%%%%%%%%%%%%%%%%%%%%%%%%%%%%%%%%%%%%%%%%%%%%%%%%
%%%%%%%%%%%%%%%%%%%%%%%%%%%%%%%%%%%%%%%%%%%%%%%%%%%%%%%%%%%%%%%%%%%%%%%%%%%%%%%

% \newpage

\section*{Impact Statement}

This paper presents work whose goal is to advance the field of Machine Learning. There are many potential societal consequences of our work, none which we feel must be specifically highlighted here.

\section*{Acknowledgements}
This work was supported by the National Research Foundation of Korea~(NRF) grant funded by the Korea government~(MSIT) (RS-2022-NR071853; RS-2023-00222663; RS-2025-16070886), 
by a grant of Korean ARPA-H Project through the Korea Health Industry Development Institute (KHIDI), funded by the Ministry of Health \& Welfare, Republic of Korea (RS-2024-00512375),
by AI-Bio Research Grant through Seoul National University, and by the Institute of Information \& Communications Technology Planning \& Evaluation (IITP) grant funded by the Korea government (MSIT) (RS-2025-02263754; RS-2021-II211341, Artificial Intelligence Graduate School Program, Chung-Ang University). Garud Iyengar's research was partially supported by ONR grant N000142312374 and NSF grant EFMA-2132142.
\bibliography{references}

\begin{thebibliography}{42}
\providecommand{\natexlab}[1]{#1}
\providecommand{\url}[1]{\texttt{#1}}
\expandafter\ifx\csname urlstyle\endcsname\relax
  \providecommand{\doi}[1]{doi: #1}\else
  \providecommand{\doi}{doi: \begingroup \urlstyle{rm}\Url}\fi

\bibitem[Abbasi-yadkori et~al.(2011)Abbasi-yadkori, P\'{a}l, and Szepesv\'{a}ri]{abbasi-yadkori2011improved}
Abbasi-yadkori, Y., P\'{a}l, D., and Szepesv\'{a}ri, C.
\newblock Improved algorithms for linear stochastic bandits.
\newblock In \emph{Advances in Neural Information Processing Systems}, volume~24. Curran Associates, Inc., 2011.

\bibitem[Abe \& Long(1999)Abe and Long]{abe1999associative}
Abe, N. and Long, P.~M.
\newblock Associative reinforcement learning using linear probabilistic concepts.
\newblock In \emph{Proceedings of the Sixteenth International Conference on Machine Learning}, ICML '99, pp.\  3–11. Morgan Kaufmann Publishers Inc., 1999.

\bibitem[Agrawal \& Goyal(2013)Agrawal and Goyal]{agrawal2013thompson}
Agrawal, S. and Goyal, N.
\newblock Thompson sampling for contextual bandits with linear payoffs.
\newblock In \emph{Proceedings of the 30th International Conference on Machine Learning}, volume~28 of \emph{Proceedings of Machine Learning Research}, pp.\  127--135. PMLR, 2013.

\bibitem[Ariu et~al.(2022)Ariu, Abe, and Proutiere]{ariu2022thresholded}
Ariu, K., Abe, K., and Proutiere, A.
\newblock Thresholded lasso bandit.
\newblock In \emph{Proceedings of the 39th International Conference on Machine Learning}, volume 162 of \emph{Proceedings of Machine Learning Research}, pp.\  878--928. PMLR, 2022.

\bibitem[Auer(2002)]{auer2002using}
Auer, P.
\newblock Using confidence bounds for exploitation-exploration trade-offs.
\newblock \emph{Journal of Machine Learning Research}, 3:\penalty0 397--422, 2002.

\bibitem[Auer et~al.(2002)Auer, Cesa-Bianchi, and Fischer]{auer2002finite-time}
Auer, P., Cesa-Bianchi, N., and Fischer, P.
\newblock Finite-time analysis of the multiarmed bandit problem.
\newblock \emph{Machine Learning}, 47:\penalty0 235--256, 2002.

\bibitem[Bang \& Robins(2005)Bang and Robins]{bang2005doubly}
Bang, H. and Robins, J.~M.
\newblock Doubly robust estimation in missing data and causal inference models.
\newblock \emph{Biometrics}, 61\penalty0 (4):\penalty0 962--973, 2005.

\bibitem[Bastani \& Bayati(2020)Bastani and Bayati]{bastani2020online}
Bastani, H. and Bayati, M.
\newblock Online decision making with high-dimensional covariates.
\newblock \emph{Operations Research}, 68\penalty0 (1):\penalty0 276--294, 2020.

\bibitem[Bogunovic et~al.(2021)Bogunovic, Losalka, Krause, and Scarlett]{bogunovic2021stochastic}
Bogunovic, I., Losalka, A., Krause, A., and Scarlett, J.
\newblock Stochastic linear bandits robust to adversarial attacks.
\newblock In \emph{Proceedings of The 24th International Conference on Artificial Intelligence and Statistics}, volume 130 of \emph{Proceedings of Machine Learning Research}, pp.\  991--999. PMLR, 2021.

\bibitem[B{\"u}hlmann \& van~de Geer(2011)B{\"u}hlmann and van~de Geer]{bhlmann2011statistics}
B{\"u}hlmann, P. and van~de Geer, S.
\newblock \emph{Statistics for High-Dimensional Data: Methods, Theory and Applications}.
\newblock Springer Berlin Heidelberg, 2011.

\bibitem[Chakraborty et~al.(2023)Chakraborty, Roy, and Tewari]{chakraborty2023thompson}
Chakraborty, S., Roy, S., and Tewari, A.
\newblock Thompson sampling for high-dimensional sparse linear contextual bandits.
\newblock In \emph{Proceedings of the 40th International Conference on Machine Learning}, volume 202 of \emph{Proceedings of Machine Learning Research}, pp.\  3979--4008. PMLR, 2023.

\bibitem[Chu et~al.(2011)Chu, Li, Reyzin, and Schapire]{chu2011contextual}
Chu, W., Li, L., Reyzin, L., and Schapire, R.
\newblock Contextual bandits with linear payoff functions.
\newblock In \emph{Proceedings of the Fourteenth International Conference on Artificial Intelligence and Statistics}, volume~15 of \emph{Proceedings of Machine Learning Research}, pp.\  208--214. PMLR, 2011.

\bibitem[Dani et~al.(2008)Dani, 9, Hayes, and Kakade]{dani2008stochastic}
Dani, V., 9, ., Hayes, T., and Kakade, S.~M.
\newblock Stochastic linear optimization under bandit feedback.
\newblock \emph{21st Annual Conference on Learning Theory - COLT 2008, Helsinki, Finland}, pp.\  355--366, 2008.

\bibitem[Dimakopoulou et~al.(2019)Dimakopoulou, Zhou, Athey, and Imbens]{dimakopoulou2019balanced}
Dimakopoulou, M., Zhou, Z., Athey, S., and Imbens, G.
\newblock Balanced linear contextual bandits.
\newblock \emph{Proceedings of the AAAI Conference on Artificial Intelligence}, 33\penalty0 (01):\penalty0 3445--3453, 2019.

\bibitem[Ghosh et~al.(2017)Ghosh, Ray~Chowdhury, and Gopalan]{ghosh2017misspecified}
Ghosh, A., Ray~Chowdhury, S., and Gopalan, A.
\newblock Misspecified linear bandits.
\newblock \emph{Proceedings of the AAAI Conference on Artificial Intelligence}, 31\penalty0 (1), 2017.

\bibitem[Goldenshluger \& Zeevi(2013)Goldenshluger and Zeevi]{goldenshluger2013linear}
Goldenshluger, A. and Zeevi, A.
\newblock {A linear response bandit problem}.
\newblock \emph{Stochastic Systems}, 3\penalty0 (1):\penalty0 230 -- 261, 2013.

\bibitem[Hao et~al.(2020)Hao, Lattimore, and Wang]{hao2020high-dimensional}
Hao, B., Lattimore, T., and Wang, M.
\newblock High-dimensional sparse linear bandits.
\newblock In \emph{Advances in Neural Information Processing Systems}, volume~33, pp.\  10753--10763. Curran Associates, Inc., 2020.

\bibitem[He et~al.(2022)He, Zhou, Zhang, and Gu]{he2022nearly}
He, J., Zhou, D., Zhang, T., and Gu, Q.
\newblock Nearly optimal algorithms for linear contextual bandits with adversarial corruptions.
\newblock In \emph{Advances in Neural Information Processing Systems}, volume~35, pp.\  34614--34625. Curran Associates, Inc., 2022.

\bibitem[Kim \& Paik(2019)Kim and Paik]{kim2019doubly-robust}
Kim, G.-S. and Paik, M.~C.
\newblock Doubly-robust lasso bandit.
\newblock In \emph{Advances in Neural Information Processing Systems}, volume~32. Curran Associates, Inc., 2019.

\bibitem[Kim et~al.(2023{\natexlab{a}})Kim, Yun, Jeong, Nam, Shin, and Combes]{kim2023contextual}
Kim, J.-H., Yun, S.-Y., Jeong, M., Nam, J., Shin, J., and Combes, R.
\newblock Contextual linear bandits under noisy features: Towards bayesian oracles.
\newblock In \emph{Proceedings of The 26th International Conference on Artificial Intelligence and Statistics}, volume 206 of \emph{Proceedings of Machine Learning Research}, pp.\  1624--1645. PMLR, 2023{\natexlab{a}}.

\bibitem[Kim et~al.(2021)Kim, Kim, and Paik]{kim2021doubly}
Kim, W., Kim, G.-S., and Paik, M.~C.
\newblock Doubly robust thompson sampling with linear payoffs.
\newblock In \emph{Advances in Neural Information Processing Systems}, volume~34, pp.\  15830--15840. Curran Associates, Inc., 2021.

\bibitem[Kim et~al.(2023{\natexlab{b}})Kim, Lee, and Paik]{kim2023double}
Kim, W., Lee, K., and Paik, M.~C.
\newblock Double doubly robust thompson sampling for generalized linear contextual bandits.
\newblock \emph{Proceedings of the AAAI Conference on Artificial Intelligence}, 37\penalty0 (7):\penalty0 8300--8307, 2023{\natexlab{b}}.

\bibitem[Kim et~al.(2023{\natexlab{c}})Kim, Paik, and Oh]{kim2023squeeze}
Kim, W., Paik, M.~C., and Oh, M.-H.
\newblock Squeeze all: Novel estimator and self-normalized bound for linear contextual bandits.
\newblock In \emph{Proceedings of The 26th International Conference on Artificial Intelligence and Statistics}, volume 206 of \emph{Proceedings of Machine Learning Research}, pp.\  3098--3124. PMLR, 2023{\natexlab{c}}.

\bibitem[Kim et~al.(2024)Kim, Iyengar, and Zeevi]{kim2024doubly}
Kim, W., Iyengar, G., and Zeevi, A.
\newblock A doubly robust approach to sparse reinforcement learning.
\newblock In \emph{Proceedings of The 27th International Conference on Artificial Intelligence and Statistics}, volume 238 of \emph{Proceedings of Machine Learning Research}, pp.\  2305--2313. PMLR, 2024.

\bibitem[Lai \& Robbins(1985)Lai and Robbins]{lai1985asymptotically}
Lai, T. and Robbins, H.
\newblock Asymptotically efficient adaptive allocation rules.
\newblock \emph{Advances in Applied Mathematics}, pp.\  4--22, 1985.

\bibitem[Lattimore \& Szepesv{\'a}ri(2020)Lattimore and Szepesv{\'a}ri]{lattimore2020bandit}
Lattimore, T. and Szepesv{\'a}ri, C.
\newblock \emph{Bandit algorithms}.
\newblock Cambridge University Press, 2020.

\bibitem[Lattimore et~al.(2020)Lattimore, Szepesvari, and Weisz]{lattimore2020learning}
Lattimore, T., Szepesvari, C., and Weisz, G.
\newblock Learning with good feature representations in bandits and in {RL} with a generative model.
\newblock In \emph{Proceedings of the 37th International Conference on Machine Learning}, volume 119 of \emph{Proceedings of Machine Learning Research}, pp.\  5662--5670. PMLR, 2020.

\bibitem[Lee et~al.(2025)Lee, Hwang, and Oh]{lee2025lasso}
Lee, H., Hwang, T., and Oh, M.-h.
\newblock Lasso bandit with compatibility condition on optimal arm.
\newblock In \emph{International Conference on Learning Representations}, 2025.

\bibitem[Li et~al.(2010)Li, Chu, Langford, and Schapire]{li2010contextual-bandit}
Li, L., Chu, W., Langford, J., and Schapire, R.~E.
\newblock A contextual-bandit approach to personalized news article recommendation.
\newblock In \emph{Proceedings of the 19th International Conference on World Wide Web}, WWW '10, pp.\  661–670. Association for Computing Machinery, 2010.

\bibitem[Oh et~al.(2021)Oh, Iyengar, and Zeevi]{oh2021sparsity-agnostic}
Oh, M.-H., Iyengar, G., and Zeevi, A.
\newblock Sparsity-agnostic lasso bandit.
\newblock In \emph{Proceedings of the 38th International Conference on Machine Learning}, volume 139 of \emph{Proceedings of Machine Learning Research}, pp.\  8271--8280. PMLR, 2021.

\bibitem[Oswal et~al.(2020)Oswal, Bhargava, and Nowak]{oswal2020linear}
Oswal, U., Bhargava, A., and Nowak, R.
\newblock Linear bandits with feature feedback.
\newblock In \emph{The Thirty-Fourth {AAAI} Conference on Artificial Intelligence, {AAAI} 2020, The Thirty-Second Innovative Applications of Artificial Intelligence Conference, {IAAI} 2020, The Tenth {AAAI} Symposium on Educational Advances in Artificial Intelligence, {EAAI} 2020, New York, NY, USA, February 7-12, 2020}, pp.\  5331--5338. {AAAI} Press, 2020.

\bibitem[Park \& Faradonbeh(2022)Park and Faradonbeh]{park2022regret}
Park, H. and Faradonbeh, M. K.~S.
\newblock A regret bound for greedy partially observed stochastic contextual bandits.
\newblock In \emph{Decision Awareness in Reinforcement Learning Workshop at ICML 2022}, 2022.

\bibitem[Park \& Faradonbeh(2024)Park and Faradonbeh]{park2024thompson}
Park, H. and Faradonbeh, M. K.~S.
\newblock Thompson sampling in partially observable contextual bandits.
\newblock \emph{arXiv preprint arXiv:2402.10289}, 2024.

\bibitem[Robbins(1952)]{robbins1952some}
Robbins, H.~E.
\newblock Some aspects of the sequential design of experiments.
\newblock \emph{Bulletin of the American Mathematical Society}, 58:\penalty0 527--535, 1952.

\bibitem[Rusmevichientong \& Tsitsiklis(2010)Rusmevichientong and Tsitsiklis]{rusmevichientong2010linearly}
Rusmevichientong, P. and Tsitsiklis, J.~N.
\newblock Linearly parameterized bandits.
\newblock \emph{Mathematics of Operations Research}, 35\penalty0 (2):\penalty0 395--411, 2010.

\bibitem[Soare et~al.(2014)Soare, Lazaric, and Munos]{soare2014best-arm}
Soare, M., Lazaric, A., and Munos, R.
\newblock Best-arm identification in linear bandits.
\newblock In \emph{Advances in Neural Information Processing Systems}, volume~27. Curran Associates, Inc., 2014.

\bibitem[Tennenholtz et~al.(2021)Tennenholtz, Shalit, Mannor, and Efroni]{tennenholtz2021bandits}
Tennenholtz, G., Shalit, U., Mannor, S., and Efroni, Y.
\newblock Bandits with partially observable confounded data.
\newblock In \emph{Proceedings of the Thirty-Seventh Conference on Uncertainty in Artificial Intelligence}, volume 161 of \emph{Proceedings of Machine Learning Research}, pp.\  430--439. PMLR, 2021.

\bibitem[Tran et~al.(2024)Tran, Ta, Mandal, and Tran-Thanh]{tran2024symmetric}
Tran, N.~P., Ta, T.~A., Mandal, D., and Tran-Thanh, L.
\newblock Symmetric linear bandits with hidden symmetry.
\newblock In \emph{Advances in Neural Information Processing Systems}, volume~37, pp.\  128699--128733. Curran Associates, Inc., 2024.

\bibitem[Tropp(2012)]{tropp2012user-friendly}
Tropp, J.~A.
\newblock User-friendly tail bounds for sums of random matrices.
\newblock \emph{Foundations of Computational Mathematics}, 12\penalty0 (4):\penalty0 389--434, 2012.

\bibitem[Tropp(2015)]{tropp2015introduction}
Tropp, J.~A.
\newblock An introduction to matrix concentration inequalities.
\newblock \emph{Found. Trends Mach. Learn.}, 8\penalty0 (1–2):\penalty0 1–230, 2015.

\bibitem[van~de Geer \& B{\"u}hlmann(2009)van~de Geer and B{\"u}hlmann]{geer2009conditions}
van~de Geer, S.~A. and B{\"u}hlmann, P.
\newblock {On the conditions used to prove oracle results for the Lasso}.
\newblock \emph{Electronic Journal of Statistics}, 3:\penalty0 1360 -- 1392, 2009.

\bibitem[Zeng et~al.(2025)Zeng, Bhatt, Koppel, and Ganesh]{zeng2024partially}
Zeng, S., Bhatt, S., Koppel, A., and Ganesh, S.
\newblock Partially observable contextual bandits with linear payoffs.
\newblock In \emph{ICASSP 2025 - 2025 IEEE International Conference on Acoustics, Speech and Signal Processing (ICASSP)}, 2025.

\end{thebibliography}
\bibliographystyle{icml2025}

%%%%%%%%%%%%%%%%%%%%%%%%%%%%%%%%%%%%%%%%%%%%%%%%%%%%%%%%%%%%%%%%%%%%%%%%%%%%%%%
%%%%%%%%%%%%%%%%%%%%%%%%%%%%%%%%%%%%%%%%%%%%%%%%%%%%%%%%%%%%%%%%%%%%%%%%%%%%%%%
% APPENDIX
%%%%%%%%%%%%%%%%%%%%%%%%%%%%%%%%%%%%%%%%%%%%%%%%%%%%%%%%%%%%%%%%%%%%%%%%%%%%%%%
%%%%%%%%%%%%%%%%%%%%%%%%%%%%%%%%%%%%%%%%%%%%%%%%%%%%%%%%%%%%%%%%%%%%%%%%%%%%%%%
\newpage
\appendix
\onecolumn

\section{Related Works}
\label{appendix:related_works}
In bandit problems, the learning agent learns only from the outcomes of chosen actions, leaving unchosen alternatives unknown~\citep{robbins1952some}. 
This constraint requires a balance between exploring new actions and exploiting actions learned to be good, known as the exploration-exploitation tradeoff. 
Efficiently managing this tradeoff is crucial for guiding the agent towards the optimal policy.
To address this, algorithms based on optimism in the face of uncertainty~\citep{lai1985asymptotically} employ the Upper Confidence Bound (UCB) strategy, which encourages the learner to select actions with the highest sum of estimated reward and uncertainty.
This approach adaptively balances exploration and exploitation, and has been widely and studied in the context of linear bandits~\citep{abe1999associative,auer2002using,dani2008stochastic,rusmevichientong2010linearly}.
Notable examples include \texttt{LinUCB}~\citep{li2010contextual-bandit,chu2011contextual} and \texttt{OFUL}~\citep{abbasi-yadkori2011improved}, which are known for their practicality and performance guarantees. 
However, existing approaches differ from ours in two key aspects: (i) they assume that the learning agent can observe the entire feature vector related to the reward, and (ii) their algorithms have regret that scales linearly with the dimension of the observed feature vector, i.e., $\tilde{O}(d\sqrt{T})$. 

In contrast, we develop an algorithm that achieves a sublinear regret bound by employing the doubly robust (DR) technique, thereby avoiding the linear dependence on the dimension of the feature vectors.
The DR estimation in the framework of linear contextual bandits was first introduced by~\citet{kim2019doubly-robust}~and~\citet{dimakopoulou2019balanced}, and subsequent studies improve the regret bound in this problem setting by a factor of $\sqrt{d}$~\citep{kim2021doubly,kim2023double}.
A recent application~\citep{kim2023squeeze} achieves a regret bound of order $O(\sqrt{dT\log T})$ under IID features over rounds.
However, the extension to non-stochastic or non-IID features remains an open question.
To address this issue, we develop a novel analysis that applies the DR estimation to non-stochastic features, achieving a regret bound sublinear with respect to the dimension of the augmented feature vectors. 
Furthermore, we extend DR estimation to handle sparse parameters, thereby further improving the regret bound to be sublinear in the reduced dimension.

Our problem is more general and challenging than misspecified linear bandits, where the assumed reward model fails to accurately reflect the true reward, such as when the true reward function is non-linear~\citep{lattimore2020bandit}, or when a deviation term is added to the reward model~\citep{ghosh2017misspecified,bogunovic2021stochastic,he2022nearly}. 
While our work assumes that the misspecified (or inaccessible) portion of the reward is linearly related to certain unobserved features, misspecified linear bandit problems can be reformulated as a special case of our framework.
While the regret bounds in \citet{lattimore2020bandit}, \citet{bogunovic2021stochastic} and \citet{he2022nearly} incorporate the sum of misspecification errors that may accumulate over the decision horizon, our work establishes a regret bound that is sublinear in the decision horizon $T$ and is not affected by misspecification errors.
\citet{ghosh2017misspecified} proposed a hypothesis test to decide between using linear bandits or MAB, demonstrating an $O(K\sqrt{T}\log T)$ regret bound when the total misspecification error exceeds $\Omega(d\sqrt{T})$.
In contrast, our algorithm achieves an $O(\sqrt{(d+d_h)T \log T})$ regret bound without requiring hypothesis tests for misspecification or partial observability.

Last but not least, our problem appears similar to the literature addressing bandit problems with partially observable features~\citep{tennenholtz2021bandits,park2022regret,park2024thompson,kim2023contextual,zeng2024partially}.
In particular,~\citet{park2022regret,park2024thompson,kim2023contextual} assume that the true features follow a specific distribution, typically Gaussian. \citet{zeng2024partially} further assume that the true features evolve according to a linear dynamical system with additive Gaussian noise.
\citet{park2022regret,park2024thompson} and~\citet{zeng2024partially} construct the observed features as emissions from the true features via a known linear mapping, also corrupted by additive Gaussian noise, whereas~\citet{kim2023contextual} first corrupt the true features with Gaussian noise and then generate the observed features by masking elements of the corrupted features following an unknown Bernoulli distribution.
In addition, all of these approaches aim to recover the true features:~\citet{park2022regret,park2024thompson} introduce a known decoder mapping from the observed features to the corresponding latent features;~\citet{kim2023contextual} leverage a Bayesian oracle strategy for estimation; and~\citet{zeng2024partially} estimate the true features using a Kalman filter.
In contrast, our setting imposes no structural assumptions on either the observed or latent features, making the problem more general and challenging than those addressed in the aforementioned works.
Furthermore, our approach does not attempt to recover any information related to latent features. 
Instead, we compensate for the lack of reward information due to unobserved features, in the sense that we project the inaccessible portion of the reward onto the orthogonal complement of the row space spanned by the observed feature vectors.

On the other hand,~\citet{tennenholtz2021bandits} assume that partially observed features are available as an offline dataset, and leverage the dataset to recover up to $L$ dimensions of the $d$-dimensional true features, where $L\le d$ is the dimension of the partially observed features.
This setting is different from ours in which partial observability arises naturally and no offline access is available.
In their framework, the correlation between the observed and unobserved features is used to model the relationship between the estimator based on the observed features and that based on the true features, which requires further estimation of the unknown correlation.
In contrast, our method is agnostic to such correlation, which makes our approach more practical.
Moreover, we exploit the observed features via feature augmentation and DR estimation, resulting in a faster convergence rate of regret compared to their UCB-based approach.

\section{Detailed Experimental Setup}
\label{appendix:exp_setup}
% For the simulation environment, we generate the true features $\zb_a$ for each arm $a \in [K]$ from $\mathcal{N}(\mathbf{0}, \mathbf{I}_{d_z})$ and construct the observed features $\xb_a$ by subsampling $d$ elements from $\zb_a$. 
% Orthogonal basis vectors $\{\mathbf{b}_1^\top, \dots, \mathbf{b}_{K-d}^\top\}$ are derived via singular value decomposition (SVD) on the observed feature matrix $\mathbf{X}$, ensuring orthogonality to $\R(\mathbf{X})$.
For both scenarios, the features---including the true features $\zb_a$, observed features $\xb_a$, and unobserved features $\ub_a$---are constructed differently based on the relationship between the row space spanned by the observed features, $\R(\Xb)$, and the row space spanned by the unobserved features, $\R(\Ub)$.
In Case 1, the general case, the true features $\zb_a$ for each arm $a \in [K]$ are sampled from $\mathcal{N}(\mathbf{0}, \mathbf{I}_{d_z})$, and the observed features $\xb_a$ are obtained by truncating the first $d$ elements of $\zb_a$, following the definition given in~\cref{eq:z_decompose}.

In Cases 2 and 3, on the other hand, the features are generated in a way to explicitly reflect the inclusion relationship between $\R(\Xb)$ and $\R(\Ub)$.
Particularly, in Case 2, where $\R(\Ub)\subseteq \R(\Xb)$, the observed features $\xb_a$ are sampled from $\mathcal{N}(\mathbf{0}, \mathbf{I}_{d})$ for each $a\in[K]$.
Then, we generate a coefficient matrix $\Cb_{\ub}\in\RR^{d_u\times d}$, where each element is sampled from $\mathrm{Unif}(-1, 1)$, and construct $\ub_a$ by computing $\ub_a = \Cb_{\ub}\xb_a$.
This construction ensures that $\R(\Ub)$ lies within $\R(\Xb)$.
In Case 3, where $\R(\Xb) \subseteq \R(\Ub)$, we reverse the process by sampling $\ub_a \sim \mathcal{N}(\mathbf{0}, \mathbf{I}_{d_u})$, generating $\Cb_{\xb} \in \RR^{d \times d_u}$ from $\mathrm{Unif}(-1, 1)$, and we set $\xb_a = \Cb_{\xb} \ub_a$, thereby ensuring $\R(\Xb)\subseteq \R(\Ub)$.
In both Case 2 and Case 3, the true features $\zb_a$ are formed by concatenating $\xb_a$ and $\ub_a$.

After the construction of the features, the orthogonal basis vectors $\{\mathbf{b}_1^\top, \dots, \mathbf{b}_{K-d}^\top\}$ are derived via singular value decomposition (SVD) on the observed feature matrix $\mathbf{X}$, ensuring orthogonality to $\R(\mathbf{X})$.
These basis vectors are linearly concatenated to $\Xb$ to form the augmented feature matrix. 
The reward parameter $\thetab_\star \in \mathbb{R}^{d_z}$ is sampled from the uniform distribution $\mathrm{Unif}(-1/2, 1/2)$, and the rewards are generated via dot products following the definition~\cref{eq:reward_definition}.
The coupling probability $p$, a hyperparameter used in the sampling distribution of $\tilde{a}_t$, is set to $0.6$ (see~\cref{eq:pseudo_distribution}). 
The confidence parameter $\delta$, which is also a hyperparameter, is set to $10^{-4}$, and the total decision horizon is $T = 1200$. 

Throughout the experiments, we fix the number of arms at $K = 30$ and the dimensionality of the true features $d_z = 35$.
Furthermore, to accommodate both partial and full observability, we set $d_z \geq d$.
Specifically, in Scenario 1, $d$ is set to $\lfloor d_z / 2 \rfloor = 17$, indicating that only about half of the full feature space is observable to the agent.
In Scenario 2, on the other hand, we set $d = 2K = 60$ and $d_z = d$, implying that latent features are absent.
The setup of the second scenario also allows us to evaluate our algorithm in the case where $d > K$.
For each of the three structural cases (i.e., the relationships between $\mathrm{R}(\Xb)$ and $\mathrm{R}(\Ub)$) considered under both scenarios, we conduct five independent trials using different random seeds.
The results are presented in terms of the sample mean and one standard deviation of the cumulative regret.

\section{Robust to Latent Feature Algorithm with Ridge Estimator}
\label{appendix:dr_ridge}

\begin{algorithm}[ht]
\caption{Robust to Latent Feature with Ridge Estimator~(\texttt{RoLF-Ridge}) }
\label{alg:RoLF_ridge}
\begin{algorithmic}[1]
\STATE \textbf{INPUT:} Observed features $\{\xb_a: a\in[K]\}$, coupling probability $p\in(1/2,1)$, confidence parameter $\delta>0$.
\STATE Initialize $\widehat{\mub}_0 = \mathbf{0}_K$, exploration phase $\Ecal_t=\emptyset$ and exploration factor $C_{\mathrm{e}}:=32(1-p)^{-2}K^2$.
%\STATE Initialize $\widehat{\mub}_0 = \mathbf{0}_K$, the exploration phase $\Ecal_t=\emptyset$ and the exploration factor $C_{\mathrm{e}}:=(\sqrt{16K}+1)^{2}2(1-p)^{-2}K^{2}\tilde{\sigma}_{\max}^{2}\tilde{\sigma}_{\min}^{-2}$.
\STATE Find orthogonal basis $\{\bb_1^\top,\ldots,\bb_{K-d}^\top\}\subseteq\mathrm{R}(\Xb)^{\perp}$ and construct $\{\tilde{\xb}_a:a\in[K]\}$.
\FOR{$t=1,\ldots,T$}
\IF{$|\Ecal_t| \le  C_{\mathrm{e}} \log (2Kt^2/\delta)$}
\STATE Randomly sample $a_t$ uniformly over $[K]$ and $\Ecal_t = \Ecal_{t-1}\cup \{t\}$.
\ELSE
\STATE Compute $\widehat{a}_t := \arg\max_{a\in[K]}\tilde{\xb}_a^\top \widehat{\mub}^{R}_{t-1}$.
\ENDIF
\WHILE{$\tilde{a}_t \neq a_t$ and $\text{count} \leq \rho_t$}
\STATE Sample $a_t$ with $\PP(a_t = \widehat{a}_t)=1-(t^{-1/2})$ and $\PP(a_t =k)=t^{-1/2}/(K-1),\; \forall k\neq\widehat{a}_t$.
\STATE Sample $\tilde{a}_t$ according to ~\cref{eq:pseudo_distribution}.
\STATE $\text{count} = \text{count}+ 1$.
\ENDWHILE
\STATE Play $a_t$ and observe $y_{a_t,t}$.
\IF{$\tilde{a}_t \neq a_t$}
\STATE Set $\widehat{\mub}^{R}_{t}:=\widehat{\mub}^{R}_{t-1}$.
\ELSE
% \STATE Compute $\widehat{\mub}^{R}_t$ in~\cref{eq:def_ridge_DR} ($\widehat{\mub}^{L}_t$ in~\cref{eq:lasso_main} for Lasso) with pseudo-rewards $\tilde{y}_{a,t}$ in~\cref{eq:pseudoY} and the imputation estimator $\check{\mub}^{R}_t$ in~\cref{eq:ridge_impute} ($\check{\mub}^{L}_t$ in~\cref{eq:lasso_impute} for Lasso).
\STATE Update $\widehat{\mub}_t^R$ following~\cref{eq:def_ridge_DR} with $\tilde{y}_{a,t}$ and update $\check{\mub}_t^R$ following~\cref{eq:ridge_impute}.
\ENDIF
\ENDFOR
\end{algorithmic}
\end{algorithm}

Our Doubly robust (DR) ridge estimator is defined as follows:
\begin{equation}
\widehat{\mub}^{R}_t := \left(\sum_{\tau=1}^{t}\II(\Mcal_{\tau})\sum_{a\in[K]}\tilde{\xb}_{a}\tilde{\xb}_{a}^\top + \Ib_K \right)^{-1} \left(\sum_{\tau=1}^{t} \II(\Mcal_{\tau}) \sum_{a\in[K]} \tilde{\xb}_{a} \tilde{y}_{a,\tau}\right),
\label{eq:def_ridge_DR}
\end{equation}
where $\tilde{y}_{a,\tau}$ is the DR pseudo reward:
\begin{equation*}
\tilde{y}_{a,t}:={\tilde{\xb}_{a}}^{\top}\check{\mub}_t^R+\frac{\II(\tilde{a}_{t}=a)}{\phi_{a,t}}\big(y_{a,t}-\tilde{\xb}_{a}^{\top}\check{\mub}_t^R\big),
\end{equation*}
and the imputation estimator $\check{\mub}^{R}_t$ is defined as
\begin{equation}
\check{\mub}^{R}_{t}:=\left(\sum_{\tau=1}^{t} \tildeobs{\tau}\tildeobs{\tau}^{\top} + p \Ib_K\right)^{-1} \left(\sum_{\tau=1}^{t} \tildeobs{\tau}y_{a_\tau,\tau} \right).
\label{eq:ridge_impute}
\end{equation}

The following theorem shows that this Ridge estimator is consistent, meaning it converges to the true parameter ${\mub}_{\star}$ with high probability as the agent interacts with the environment.

\begin{theorem}[Consistency of the DR Ridge estimator]
\label{thm:est_ridge_consistency}
% Let $\Vb_{t}:= t\sum_{a\in[K]}\tilde{\xb}_{a}\tilde{\xb}_{a}^{\top}$ denote the Gram matrix at round $t$.
%Let $\Vb_{t}:= \sum_{\tau=1}^t\sum_{a\in[K]}\tilde{\xb}_{a}\tilde{\xb}_{a}^{\top}$ denote the Gram matrix at round $t$.
For each $t$, let $\Ecal_t \subseteq [t]$ denote an exploration phase such that for each $\tau\in\Ecal_t$ the action $a_{\tau}$ is sampled uniformly over $[K]$. 
Assume that $\|\mub_{\star}\|_{\infty}\le 1$ and $\|\tilde{\xb}_a\|_{\infty} \le 1$ for all $a\in[K]$.
Then with probability at least $1-3\delta$,
\begin{equation*}
\max_{a\in[K]}|\tilde{\xb}_{a}^\top(\widehat{\mub}^{R}_t-{\mub}_{\star})| \le\frac{2}{\sqrt{t}}\left(\frac{\sigma}{p}\sqrt{K\log\frac{t+1}{\delta}}+\sqrt{K}\right),
\end{equation*}
for all rounds $t$ such that $|\Ecal_t| \ge 32 (1-p)^{-2}K^{2}\log(2Kt^2/\delta)$.
\end{theorem}

With $|\Ecal_t| = O(K^2 \log K t)$ exploration rounds, the DR Ridge estimator achieves an $O(\sqrt{K/t})$ convergence rate over all $K$ rewards.
This is possible because the DR pseudo-rewards defined in~\cref{eq:pseudoY} impute the missing rewards for all arms $a\in[K]$ using $\tilde{\xb}_{a}^\top\check{\mub}_t$, based on the samples collected during the exploration phase, $\Ecal_t$.
With this convergence guarantee, we establish a regret bound for \texttt{RoLF-Ridge}, which is an adaptation of~\cref{alg:RoLF} using the Ridge estimator.

\begin{theorem}[Regret bound for Ridge \texttt{RoLF}]
\label{thm:ridge_regret}
Suppose $\|\mub_{\star}\|_{\infty}\le 1$ and $\|\zb_a\|_{\infty} \le 1$ for all $a\in[K]$.
For $\delta\in(0,1)$, with probability at least $1-4\delta$, the cumulative regret of the proposed algorithm using the DR Ridge estimator is bounded by
\begin{equation*}
\regret(T)\le \frac{32K^{2}}{(1-p)^{2}}\log\frac{2dT^{2}}{\delta}+2 \sqrt{2T \log \frac{2}{\delta}} + \frac{4 \sqrt{T}}{K-1} + 4\delta+8\sqrt{KT}\left(\frac{\sigma}{p}\sqrt{\log\frac{T}{\delta}}+1\right),
\end{equation*}
\end{theorem}

The first and second terms come from the distribution of $a_t$, which is a combination of the $1-t^{-1/2}$-greedy policy and resampling up to~$\rho_t:=\log((t+1)^2/\delta)/\log(1/p)$ trials.
The third term is determined by the size of the exploration set, $\Ecal_t$, while the last term arises from the estimation error bounded by the DR estimator as described in~\cref{thm:est_ridge_consistency}.
The hyperparameter $p\in(1/2,1)$ balances the size of the exploration set in the third term and the estimation error in the last term.
Overall, the regret is $O(\sqrt{KT \log T})$, which shows a significant improvement compared to the regret lower bound in~\cref{thm:regret_linear_lower_bound} for any linear bandit algorithm that does not account for unobserved features and unobserved rewards.
% \sw{(As I have found of, the regret of \texttt{UCB} for MAB is $O(\sqrt{KT\log T})$... what about we just skip this algorithm?)}

\section{A Modified Algorithm for Time-Varying Observed Features}
\label{appendix:time_varying}
In this section, we define the problem of linear bandits with partially observable features under a setting where the observed features vary over time, describe our proposed method, and provide theoretical guarantees.

\subsection{Problem Formulation}

Let $\xb_{1,t},\ldots,\xb_{K,t}$ denote the observed features and $\ub_{1},\ldots,\ub_{K}$ denote the unobserved features.
Now the observed features arbitrarily change over $t$ but the unobserved features are fixed over time.
When the algorithm selects an arm $a_t$, the reward is 
\begin{equation*}
    y_{a_t,t}= \langle \xb_{a_t,t}, \thetab_{\star}^{(o)} \rangle + \langle \ub_{a_t},\thetab_{\star}^{(u)}\rangle + \epsilon_t,
\end{equation*}
where $\epsilon_t$ is Sub-Gaussian noise that follows \cref{asm:sub-gaussian}.
The expected reward of each arm is stable over time, where MAB algorithms without using features are applicable for achieving a $\tilde{O}(\sqrt{KT})$ regret bound.
When the observed features vary over time, the expected reward of each arm $\EE[y_{a_t,t}]=\langle \xb_{a_t,t}, \thetab_{\star}^{(o)} \rangle + \langle \ub_{a_t},\thetab_{\star}^{(u)}\rangle$ also arbitrarily changes over time, and MAB algorithms suffer regret linear in $T$.
To our knowledge, there is no other work that addresses this challenging setting.

\subsection{Proposed Method: Orthogonal Basis Augmentation}
We address the problem by augmenting standard basis $\eb_1,\ldots,\eb_K$ in $\RR^K$ to estimate bias caused by the unobserved features.
Let $\tilde{\xb}_{a,t} := \eb_a^\top [\Xb_t \; \eb_1 \; \cdots \; \eb_K] \in \RR^{d+K}$ and let $\Delta_{a} := \langle \ub_a, \theta_{\star}^{(u)} \rangle$ denote the bias that stems from the latent features.
Then,
\begin{align*}
y_{a,t} &= \langle \xb_{a,t}^\top, \thetab_{\star}^{(o)} \rangle + \langle \ub_{a,t}^\top ,\thetab_{\star}^{(u)} \rangle + \epsilon_{t}
\\ &= \langle \eb_a^\top [\Xb_t\; \eb_1\; \ldots\; \eb_K],  [\thetab_{\star}^{(o)}\; \Delta_1\; \ldots\; \Delta_K]\rangle + \epsilon_{t}.
\end{align*}
Therefore, applying the \texttt{RoLF-Ridge} algorithm to the augmented features $\tilde{\xb}_{a,t} := \eb_a^\top [\Xb_t\; \eb_1\; \ldots\; \eb_K]$ yields the following regret bound.

\begin{theorem}[Regret bound for \texttt{Ridge-RoLF-V} with time-varying observed features]
\label{thm:regret_ridge_time_varying}
If the observed features are vary over time, then for any $\delta\in(0,1)$, with probability at least $1-4\delta$, the cumulative regret of the proposed algorithm \texttt{Ridge-RoLF-V} using DR Ridge estimator is bounded by
\begin{equation*}
\begin{split}
\regret(T)\le &4\delta+\frac{2\sqrt{T}}{d+K-1}+\frac{32(K+d)^{2}}{(1-p)^{2}}\log\frac{2(K+d)T^{2}}{\delta}+8\sqrt{(d+K)T}\left(\frac{\sigma}{p}\sqrt{\log\frac{T^2}{\delta}}+1\right).
\end{split}
\end{equation*}
\end{theorem}

The proof follows similar arguments given in \cref{thm:ridge_regret} and is therefore omitted.
The resulting regret bound achieves a rate of $\tilde{O}(\sqrt{(d+K)T})$, which, to the best of our knowledge, is the first sublinear regret guarantee for partially observable linear bandits (as well as misspecified linear bandits) with arbitrarily time-varying observed features.

\section{Missing Proofs}
\label{appendix:proofs}
\subsection{Proof of \cref{thm:regret_linear_lower_bound}}
\label{appendix:proof_thm1}
Throughout this paper, we consider a bandit problem where the agent observes only a subset of the reward-generating feature vector and cannot access or estimate the unobserved portion.
If the agent uses online decision-making algorithms that rely solely on observed features, as defined in \cref{def:obs_dependent_policy}, the resulting issue can be interpreted as a model misspecification.
Therefore, in this theorem, we present a problem instance where ``misspecified'' algorithms, considering only observed features, may incur regret that grows linearly in $T$.

Following the statement of~\cref{thm:regret_linear_lower_bound}, we assume that $d = d_u = 1$, which means $d_z = 2$.
Given the true feature set $\mathcal{Z} = \{[1,3]^\top, [2,19/4]^\top\}$, let the first element of each vector is observed to the agent; while the second element remains unobserved.
This results in $\xb_1 =x_1= 1$, $\xb_2 =x_2= 2$, $\ub_1 =u_1= 3$ and $\ub_2=u_2 = 19/4$.
We set the true parameter as $\thetab_{\star}\in\RR^2 = [2, -1]^\top$, meaning $\thetao = \theta_\star^{(o)} = 2$ and $\thetau= \theta_\star^{(u)} = -1$.
Using the reward function from \cref{subsec:problem_formulation} and considering \cref{asm:sub-gaussian}, the expected reward for each arm is given by
\begin{equation*}
    \gamma_i:=\mathbb{E}[y_i]= \mathbf{z}_i^\top \theta_\star = x_i\theta_\star^{(o)} + u_i\theta_\star^{(u)}\quad \forall\ i\in\{1,2\}.
\end{equation*}

Plugging the values in, the true mean reward for each arm is directly computed as $\gamma_1 = 2-3 = -1$ and $\gamma_2 = 4-19/4 = -3/4$, which satisfies the assumption that its absolute value does not exceed 1~(\cref{subsec:problem_formulation}), and since $\gamma_1 < \gamma_2$, arm 2 is the optimal action.
We further assume that the total learning horizon $T > 256\sigma^2\max\{\log2, \log(1/\delta')\}$ for any $\delta'\in(0,1/2)$, where $\sigma$ is the sub-Gaussian parameter of the reward noise.

For brevity, we denote the latent reward components as $g_1 := u_1 \theta_\star^{(u)}$ and $g_2 := u_2 \theta_\star^{(u)}$, yielding $g_1 = -3$ and $g_2 = -19/4$.
Since $\gamma_2 \ne 2\gamma_1$ and $\lvert g_i \rvert \ge 3 > 0$ for all $i \in \{1, 2\}$, our problem setup satisfies the ``large deviation'' criterion in Definition 1 and Theorem 2 of~\citet{ghosh2017misspecified}, with $l = 3$ and $\beta = 0$.
Applying the theorem, it follows that \texttt{OFUL}~\citep{abbasi-yadkori2011improved} suffers linear regret in this problem instance, i.e., $\Omega(T)$.
Motivated by this result, we further show that \texttt{LinTS}~\citep{agrawal2013thompson} is similarly affected.

For each round $t\in[T]$, \texttt{LinTS} estimates the true parameter using the ridge estimator, given by:
\begin{align}
    \widehat{\thetab}_t 
    &= (\Xb_t^\top \Xb_t + \lambda \Ib_d)^{-1}(\Xb_t^\top \Yb_t) \notag \\
    &= (\Xb_t^\top \Xb_t + \lambda \Ib_d)^{-1}(\Xb_t^\top (\Xb_t\thetao + \gb_t+ \epsb_t)) \notag \\
    &= \thetao - \lambda \Vb_t^{-1}\thetao + \Vb_t^{-1}\Xb_t^\top \gb_t+ \Vb_t^{-1}\Xb_t^\top\epsb_t, \label{eq:thm1_ridge_decompose}
\end{align}
where $\Xb_t:=(\xb_{a_1}^\top,\dots,\xb_{a_t}^\top)\in\RR^{t\times d}$ is a matrix containing features chosen up to round $t$, $\Yb_t := (y_{a_1},\dots,y_{a_t})\in\RR^t$ is a vector of observed rewards, and $\epsb_t := (\epsilon_1,\dots, \epsilon_t)\in\RR^t$ contains noise attached to each reward. Unlike a typical ridge estimator, here the term $\gb_t:= (g_{a_1},\dots, g_{a_t})\in\RR^{t}$, the vector containing the latent portion of observed rewards is introduced due to model misspecification.
Note that $\Vb_t := (\Xb_t^\top \Xb_t + \lambda \Ib_d)\succ 0$.

For this problem instance, since $d = 1$,~\cref{eq:thm1_ridge_decompose} is equivalent to:
\begin{equation*}
    \widehat{\theta}_t = \theta_\star^{(o)} - \frac{\theta_\star^{(o)}}{\sum_{\tau=1}^tx_{a_\tau}^{2} + 1} + \frac{\sum_{\tau=1}^tx_{a_\tau}g_{a_\tau}}{\sum_{\tau=1}^tx_{a_\tau}^{2} + 1} + \frac{\sum_{\tau=1}^t x_{a_\tau}\epsilon_{\tau}}{\sum_{\tau=1}^t x_{a_\tau}^{2} + 1},
\end{equation*}
where we assume $\lambda = 1$.
Note that we also denote $\widehat{\thetab}_t$ and $\thetao$ by $\widehat{\theta}_t$ and $\theta_\star^{(o)}$, respectively, since both are scalars. 
Hence, the estimation error is computed as:
\begin{equation}
\widehat{\theta}_t - \theta_\star^{(o)} = -\frac{\theta_\star^{(o)}}{\sum_{\tau=1}^t x_{a_\tau}^{2} + 1} + \frac{\sum_{\tau=1}^t x_{a_\tau}g_{a_\tau}}{\sum_{\tau=1}^t x_{a_\tau}^{2} + 1} + \frac{\sum_{\tau=1}^t x_{a_\tau}\epsilon_{\tau}}{\sum_{\tau=1}^t x_{a_\tau}^{2} + 1}.
\label{eq:thm1_ridge_est_error}
\end{equation}
Let $ N_1 $ and $ N_2 $ denote the number of times arms 1 and 2 have been played up to round $ t $, respectively. 
This implies that $N_1 + N_2 = t$.
Then, for the numerator of the second term, since
\begin{equation*}
\sum_{\tau=1}^t x_{a_\tau}g_{a_\tau} = (\underbrace{g_1+\cdots+g_1}_{N_1} + \underbrace{2g_2+\cdots+2g_2}_{N_2}) = g_1 N_1 + 2g_2 N_2,
\end{equation*}
we can observe that
\begin{align*}
    \sum_{\tau=1}^t x_{a_\tau}g_{a_\tau}
    &= g_1 N_1 + 2g_2 N_2 \\
    &\ge \underline{g}N_1 + 2\underline{g}N_2 \\
    &= \underline{g}N_1 + 2\underline{g}(t - N_1) \\
    &= 2\underline{g}t - \underline{g}N_1 \\
    &\ge \underline{g}t \qquad\qquad (\because N_1 \le t) \\
    &= -\frac{19}{4}t,
\end{align*}
where $\underline{g} = \min\{g_1,g_2\} = -19/4$, which implies that $\sum_{\tau=1}^t x_{a_\tau}g_{a_\tau} = \Theta(t)$.
For the denominator, $\sum_{\tau=1}^t x_{a_\tau}^{2} + 1$, which grows at a rate of $O(t)$, implying that the second term of the right-hand side in~\cref{eq:thm1_ridge_est_error} is $\Theta(1)$, and that $\widehat{\theta}_t$ is not consistent since it does not converge to $\theta_\star^{(o)}$ as $t\to\infty$.

For arm 2, which is optimal, to be selected in round $ t+1 $ under \texttt{LinTS}, the condition $ x_2 \tilde\theta_t \ge x_1 \tilde\theta_t $ must hold, where $ \tilde\theta_t \sim \mathcal{N}\left( \widehat{\theta}_t, v^2(\sum_{\tau=1}^t x_{a_\tau}^2 + 1)^{-1} \right) $.
Given the assumptions that $ x_1 = 1 $ and $ x_2 = 2 $, arm 2 is selected whenever $ \tilde\theta_t \ge 0 $. Thus, for arm 1 to be chosen, we require $ \tilde\theta_t < 0 $. 
We will show that the probability of $ \tilde\theta_t < 0 $ does not diminish sufficiently to be ignored, even when the agent plays for a sufficiently large amount of time.
To clarify, let us define the two events $E_{\tilde\theta} := \{\tilde\theta_t \ge 0\}$ and $E_{\hat\theta} := \{\widehat{\theta}_t \ge 0\}$.
We revisit~\cref{eq:thm1_ridge_decompose} as follows:
\begin{align}
    \widehat{\theta}_t &= \frac{\theta_\star^{(o)}\sum_{\tau=1}^t x_{a_\tau}^2}{\sum_{\tau=1}^t x_{a_\tau}^{2} + 1} + \frac{\sum_{\tau=1}^t x_{a_\tau}g_{a_\tau}}{\sum_{\tau=1}^t x_{a_\tau}^{2} + 1} + \frac{\sum_{\tau=1}^t x_{a_\tau}\epsilon_{\tau}}{\sum_{\tau=1}^t x_{a_\tau}^{2} + 1} \notag \\
    &\le \theta_{\star}^{(o)} +\frac{g_1 N_1 + 2g_2 N_2}{N_1+4N_2 + 1} + \frac{\sum_{\tau=1}^t x_{a_\tau}\epsilon_{\tau}}{N_1+4N_2 + 1} \qquad\qquad\quad (\because \theta_\star^{(o)} > 0)\notag \\
    % &\le \theta_{\star}^{(o)} +\frac{g_1 N_1 + 2g_2 N_2}{N_1+4N_2 + 1} + \frac{\max_{a\in\{1,2\}} x_a}{N_1+4N_2+1}\sum_{\tau=1}^{t}\epsilon_\tau \notag \\
    &\le \theta_{\star}^{(o)} +\frac{g_1 N_1 + 2g_2 N_2}{N_1+4N_2 + 1} + \frac{2}{N_1+4N_2+1}\sum_{\tau=1}^{t}\epsilon_\tau, \label{eq:thm1_ridge_decompose_bound}
\end{align}
where the last inequality holds since $\sum_{\tau=1}^t x_{a_\tau}\epsilon_{\tau} \le \max_{a\in\{1,2\}}x_a\sum_{\tau=1}^t \epsilon_\tau$ and $\max_{a\in\{1,2\}} x_a = 2$.
For the second term of~\cref{eq:thm1_ridge_decompose_bound}, for $t \ge 19$,
\begin{align*}
    \frac{g_1 N_1 + 2g_2 N_2}{N_1+4N_2 + 1} 
    & = \frac{-3N_1-\frac{19}{2}N_2}{N_1+4N_2+1} \\
    &\le \frac{-\frac{19}{8}(N_1+4N_2)}{N_1+4N_2+1} \\
    &= -\frac{19}{8} + \frac{19/8}{t+3N_2+1} \\
    &\le -\frac{19}{8} + \frac{19}{8t},
\end{align*}
which is upper bounded by $-{9}/{4}$. This bound is followed by:
\begin{equation}
\begin{split}
    \widehat{\theta}_t
    &\le \theta_{\star}^{(o)} +\frac{g_1 N_1 + 2g_2 N_2}{N_1+4N_2 + 1} + \frac{2}{N_1+4N_2+1}\sum_{\tau=1}^{t}\epsilon_\tau \\
    &\le 2-\frac{9}{4} + \frac{2}{t+3N_2+1}\sum_{\tau=1}^t\epsilon_\tau \\
    % &= -\frac{1}{4}+ \frac{2}{t+3N_2+1}\sum_{\tau=1}^t\epsilon_\tau \\
    &\le -\frac{1}{4}+ \frac{2}{t}\sum_{\tau=1}^t\epsilon_\tau.
\end{split}
\label{eq:ridge_upper_bound}
\end{equation}
Thus, we have the following: 
\begin{equation*}
    \PP(\widehat{\theta}_t > 0) \le \PP\left(-\frac{1}{4}+ \frac{2}{t}\sum_{\tau=1}^t\epsilon_\tau> 0\right) = \PP\left(\frac{2}{t}\sum_{\tau=1}^t\epsilon_\tau> \frac{1}{4}\right).
\end{equation*}
Since $\epsilon_\tau$ is an IID sub-Gaussian random variable for all $\tau\in[t]$, by applying Hoeffding inequality we obtain: $\PP(\widehat{\theta}_t > 0) \le \exp\left(-t/128\sigma^2\right)$.
Given this, we now bound the probability of the event $E_{\tilde{\theta}}$:
\begin{align}
    \mathbb{P}(E_{\tilde\theta}) &= \mathbb{P}(E_{\tilde\theta}\cap E_{\hat\theta}) + \mathbb{P}(E_{\tilde\theta}\cap E_{\hat\theta}^c) \notag \\
&= \mathbb{P}(E_{\tilde\theta}\mid E_{\hat\theta})\cdot \mathbb{P}(E_{\hat\theta}) + \mathbb{P}(E_{\tilde\theta} \mid  E_{\hat\theta}^c) \cdot \mathbb{P}(E_{\hat\theta}^c) \notag \\
&= \mathbb{P}(\tilde\theta_t \ge 0\mid \widehat{\theta}_t \ge 0)\cdot \mathbb{P}(\widehat{\theta}_t \ge 0) + \mathbb{P}(\tilde\theta_t \ge 0\mid \widehat{\theta}_t < 0)\cdot \mathbb{P}(\widehat{\theta}_t < 0) \notag \\
&\le \exp\left(-\frac{t}{128\sigma^2}\right) + \mathbb{P}(\tilde\theta_t \ge 0\mid \widehat{\theta}_t < 0). \label{eq:thm1_prob_bound}
\end{align}
Since the second term of~\cref{eq:thm1_prob_bound} is calculated under a Gaussian distribution, thus its value does not exceed $1/2$ for all $t\in[T]$, it follows that $\PP(E_{\tilde\theta}^c) \ge 1/2 - \exp\left(-{t}/128\sigma^2\right).$
Note that the total decision horizon $T > 256\sigma^2\log(1/\delta^\prime)$, thus for any $t > 128\sigma^2\log(1/\delta^\prime)$, we have $\mathbb{P}(E_{\tilde\theta}^c) \ge 1/2 - \delta^\prime$.
This implies that for rounds beyond $T/2$, the suboptimal arm will be played at least $(1/2-\delta^\prime)T/2$ times for any $\delta^\prime\in(0,1/2)$, thus incurring
\begin{equation*}
    \EE[\regret_{\texttt{LinTS}}(T)] \ge (\gamma_2-\gamma_1)\left(\frac{1}{2}-\delta^\prime\right)\frac{T}{2} = \frac{1}{4}\left(\frac{1}{2}-\delta^\prime\right)\frac{T}{2}.
\end{equation*}

For \texttt{OFUL}, we also present another analysis that does not require the assumption that the suboptimal arm is played for initial $t$ rounds, which is taken in Theorem 2 of~\citet{ghosh2017misspecified}.
The optimal arm, arm 2, is selected when 
\begin{equation}
x_2 \widehat{\theta}_t + \frac{x_2}{\sqrt{1 + \sum_{\tau=1}^{t-1}x_{a_\tau}^2 }} > x_1 \widehat{\theta}_t + \frac{x_1}{\sqrt{1 + \sum_{\tau=1}^{t-1}x_{a_\tau}^2 }},
\label{eq:OFUL_condition}
\end{equation}
where $\widehat{\theta}_t$ is the same ridge estimator as in \texttt{LinTS}.
The inequality \cref{eq:OFUL_condition} is equivalent to $\widehat{\theta}_t > (1 + \sum_{\tau=1}^{t-1}x_{a_\tau}^2 )^{-1/2}$, which implies $\widehat{\theta}_t > 1/\sqrt{2t}$.
By \cref{eq:ridge_upper_bound}, 
\begin{align*}
\PP\left(\widehat{\theta}_t > \frac{1}{\sqrt{2t}} \right) &\le \PP\left( -\frac{1}{4}+ \frac{2}{t}\sum_{\tau=1}^t\epsilon_\tau > \frac{1}{\sqrt{2t}}\right) \\
&= \PP\left(\frac{2}{t}\sum_{\tau=1}^t\epsilon_\tau > \frac{1}{\sqrt{2t}}+\frac{1}{4}\right) \\
&\le \PP\left(\frac{2}{t}\sum_{\tau=1}^t\epsilon_\tau > \frac{1}{4}\right) \le \exp\left(-\frac{t}{128\sigma^2}\right).
\end{align*}
Thus, for $t \ge 128\sigma^2 \log 2$, the probability of selecting arm 2 is less than $1/2$ and for $ T > 256 \sigma^2 \log 2$,
\[
\EE[\regret_{\texttt{OFUL}}(T)] \ge (\gamma_2-\gamma_1)\cdot\frac{1}{2} \cdot \frac{T}{2} = \frac{T}{16},
\]
and the algorithm suffers expected regret linear in $T$.
\hfill \qed

\subsection{Proof of \cref{thm:est_lasso_consistency}}
\label{appendix:proof_thm2}
Let $\Vb_{t}:=\sum_{\tau=1}^{t}\sum_{a\in[K]}\mathbf{\tilde{x}}_{a}\mathbf{\tilde{x}}_{a}^{\top}$.
Then
\begin{equation*}
\max_{a\in[K]} |\tilde{\xb}_{a}^\top (\widehat{\mub}^{L}_t-{\mub}_{\star})| \le \sqrt{\sum_{a\in[K]} |\tilde{\xb}_{a}^\top (\widehat{\mub}^{L}_t-{\mub}_{\star})|^2} = t^{-1/2} \|\widehat{\mub}^{L}_t-{\mub}_{\star}\|_{\Vb_t}.
\end{equation*}
To use Lemma~\ref{lem:lasso}, we prove a bound for $\Vert \sum _{\tau=1}^{t} \sum_{a\in[K]} \left(\tilde{y}_{a,\tau} - \tilde{\xb}_{a}^\top  \mub_{t}\right) \tilde{\xb}_{a} \Vert_{\infty}$.
By definition of $\tilde{y}_{a,\tau}$,
\begin{align}
&\left\Vert \sum_{\tau=1}^{t}\sum_{a\in[K]}\left(\tilde{y}_{a,\tau}-\tilde{\xb}_{a}^{\top}\mub_{\star}\right)\tilde{\xb}_{a}\right\Vert _{\infty} \notag \\
&=\left\Vert \sum_{\tau=1}^{t}\sum_{a\in[K]}\left(1-\frac{\II(\tilde{a}_{\tau}=a)}{\phi_{a,\tau}}\right)\mathbf{\tilde{x}}_{a}\mathbf{\tilde{x}}_{a}^{\top}\left(\check{\mub}_{t}^{L}-\mub_{\star}\right)+\frac{\II(\tilde{a}_{\tau}=a)}{\phi_{a,\tau}}\left(y_{a,\tau}-\tilde{\xb}_{a}^{\top}\mub_{\star}\right)\tilde{\xb}_{a}\right\Vert _{\infty} \notag \\
&\le\left\Vert \sum_{\tau=1}^{t}\sum_{a\in[K]}\left(1-\frac{\II(\tilde{a}_{\tau}=a)}{\phi_{a,\tau}}\right)\mathbf{\tilde{x}}_{a}\mathbf{\tilde{x}}_{a}^{\top}\left(\check{\mub}_{t}^{L}-\mub_{\star}\right)\right\Vert _{\infty}+\left\Vert \sum_{\tau=1}^{t}\sum_{a\in[K]}\frac{\II(\tilde{a}_{\tau}=a)}{\phi_{a,\tau}}\left(y_{a,\tau}-\tilde{\xb}_{a}^{\top}\mub_{\star}\right)\mathbf{\tilde{x}}_{a}\right\Vert _{\infty}. \label{eq:bound_pseudo_martingale}
\end{align}
With probability at least $1-\delta/(\tau+1)^2$, the event $\Mcal_{\tau}$ happens for all $ \tau \ge 1$ and we obtain a pair of matching sample $\tilde{a}_{\tau}$ and $a_{\tau}$. 
Under $\Mcal_\tau$, the second term in~\cref{eq:bound_pseudo_martingale} is equal to,
\begin{align*}
\left\Vert \sum_{\tau=1}^{t}\sum_{a\in[K]}\frac{\II(\tilde{a}_{\tau}=a)}{\phi_{a,\tau}}\left(y_{a,\tau}-\tilde{\xb}_{a}^{\top}\mub_{\star}\right)\mathbf{\tilde{x}}_{a}\right\Vert _{\infty}
&= \left\Vert \sum_{\tau=1}^{t}\sum_{a\in[K]}\frac{\II(a_{\tau}=a)}{\phi_{a,\tau}}\left(y_{a,\tau}-\mathbf{\tilde{x}}_{a}^{\top}\mub_{\star}\right)\mathbf{\tilde{x}}_{a}\right\Vert _{\infty} \\
&= \frac{1}{p}\left\Vert \sum_{\tau=1}^{t}\epsilon_{\tau}\mathbf{\tilde{x}}_{a_{\tau}}\right\Vert _{\infty}.
\end{align*}
Because $\norm{\vb}_\infty = \max_{i\in[d]}\abs{\eb_i^\top \vb}$ for any $\vb\in\RR^{d}$,
\begin{equation*}
\frac{1}{p}\left\Vert \sum_{\tau=1}^{t}\epsilon_{\tau}\mathbf{\tilde{x}}_{a_{\tau}}\right\Vert _{\infty}=\frac{1}{p}\max_{a\in[K]}\left\vert \sum_{\tau=1}^{t}\epsilon_{\tau}\eb_{a}^{\top}\mathbf{\tilde{x}}_{a_{\tau}}\right\vert.
\end{equation*}
Applying Lemma~\ref{lem:exp_bound}, with probability at least $1-\delta/t^2$,
\begin{align*}
\max_{a\in[K]}\left\vert \sum_{\tau=1}^{t}\epsilon_{\tau}\eb_{a}^{\top}\mathbf{\tilde{x}}_{a_{\tau}}\right\vert
&\le\max_{a\in[K]}\sigma\sqrt{2\sum_{\tau=1}^{t}\left(\eb_{a}^{\top}\mathbf{\tilde{x}}_{a_{\tau}}\right)^{2}\log\frac{2Kt^{2}}{\delta}} \\
&=\max_{a\in[K]}\sigma\sqrt{2\eb_{a}^{\top}\left(\sum_{\tau=1}^{t}\mathbf{\tilde{x}}_{a_{\tau}}\mathbf{\tilde{x}}_{a_{\tau}}^{\top}\right)\eb_{a}\log\frac{2Kt^{2}}{\delta}}\\
&\le\max_{a\in[K]}\sigma\sqrt{2\eb_{a}^{\top}\Vb_t\eb_{a}\log\frac{2Kt^{2}}{\delta}}\\
&=\sigma \tilde{\sigma}_{\max} \sqrt{2 t \log\frac{2Kt^{2}}{\delta}},
\end{align*}
where the last equality follows by the definition $\tilde{\sigma}^2_{\max} = \max_{a\in[K]} \eb_a^\top (\sum_{a\in[K]} \tilde{\xb}_{a} \tilde{\xb}_{a}^\top ) \eb_a $.
Thus,
\begin{equation}
\frac{1}{p} \left\Vert \sum_{\tau=1}^{t}\epsilon_{\tau}\mathbf{\tilde{x}}_{a_{\tau}}\right\Vert _{\infty} \le  \frac{\sigma \tilde{\sigma}_{\max}}{p}\sqrt{2t\log\frac{2Kt^{2}}{\delta}}. 
\label{eq:est_error_norm}
\end{equation}

Now we turn to the first term in~\cref{eq:bound_pseudo_martingale}.
Define $\Ab_{t}:=\sum_{\tau=1}^{t}\sum_{a\in[K]}\II(\tilde{a}_{\tau}=a)\phi_{a,\tau}^{-1} \mathbf{\tilde{x}}_{a}\mathbf{\tilde{x}}_{a}^{\top}$.
Then the first term is rearranged as
\begin{equation}
\left\Vert \sum_{\tau=1}^{t}\sum_{a\in[K]}\left(1-\frac{\II(\tilde{a}_{\tau}=a)}{\phi_{a,\tau}}\right)\mathbf{\tilde{x}}_{a}\mathbf{\tilde{x}}_{a}^{\top}\left(\check{\mub}_{t}^{L}-\mub_{\star}\right)\right\Vert _{\infty}=\left\Vert \left(\Vb_{t}-\Ab_{t}\right)\left(\check{\mub}_{t}^{L}-\mub_{\star}\right)\right\Vert _{\infty}.
\label{eq:est_1st_term}
\end{equation}
Since $\norm{\vb}_\infty = \max_{i\in[d]}\abs{\eb_i^\top \vb}$ for any $\vb\in\RR^{d}$,
\begin{align*}
\left\Vert \left(\Vb_{t}-\Ab_{t}\right)\left(\check{\mub}_{t}^{L}-\mub_{\star}\right)\right\Vert _{\infty}=&\max_{a\in[K]}|\eb_{a}^{\top}\left(\Vb_{t}-\Ab_{t}\right)\left(\check{\mub}_{t}^{L}-\mub_{\star}\right)|\\
\le&\max_{a\in[K]}\left\Vert \eb_{a}^{\top}\left(\Vb_{t}-\Ab_{t}\right)\Ab_{t}^{-1/2}\right\Vert _{2}\!\!\left\Vert \check{\mub}_{t}^{L}-\mub_{\star}\right\Vert_{\Ab_t}.
\end{align*}
Because $\check{\mub}_{t}^{L}$ is a minimizer of \cref{eq:lasso_impute}, by Lemma~\ref{lem:lasso} and \cref{eq:est_error_norm},
\begin{equation*}
\left\Vert \check{\mub}_{t}^{L}-\mub_{\star}\right\Vert _{\Ab_t}\le \frac{4\sigma \tilde{\sigma}_{\max}}{p} \sqrt{\frac{2t(d+d_{h})\log(2Kt^2/\delta)}{\lambda_{\min}\left(\Ab_{t}\right)}}.
\end{equation*}
By \cref{cor:Gram_matrix}, with $\epsilon \in (0,1)$ to be determined later, for $t\ge8\epsilon^{-2}(1-p)^{-2}K^{2}\log(2Kt^2/\delta)$, with probability at least $1-\delta/t^2$, 
\begin{equation}
\left\Vert \Ib_{K}-\Vb_{t}^{-1/2}\Ab_{t}\Vb_{t}^{-1/2}\right\Vert_{2} \le \epsilon.
\label{eq:lasso_Gram_bound}
\end{equation}
\cref{eq:lasso_Gram_bound} implies $(1-\epsilon) \Ib_K \preceq \Vb_{t}^{-1/2}\Ab_{t}\Vb_{t}^{-1/2}$ and $(1-\epsilon) \Vb_t \preceq \Ab_t$.
Thus,
\begin{equation*}
\left\Vert \check{\mub}_{t}^{L}-\mub_{\star}\right\Vert _{\Ab_t}\le  \frac{4\sigma \tilde{\sigma}_{\max}}{p} \sqrt{\frac{2t(d+d_{h})\log(2Kt^2/\delta)}{(1-\epsilon) \lambda_{\min}\left(\Vb_{t}\right)}}=\frac{4\sigma \tilde{\sigma}_{\max}}{p\tilde{\sigma}_{\min}}\sqrt{\frac{2(d+d_{h})\log(2Kt^2/\delta)}{1-\epsilon}},
\end{equation*}
and \cref{eq:est_1st_term} is bounded by,
\begin{align}
\left\Vert \left(\Vb_{t}-\Ab_{t}\right)\left(\check{\mub}_{t}^{L}-\mub_{\star}\right)\right\Vert _{\infty}\ &\le \max_{i\in[K]}\left\Vert e_{i}^{\top}\left(\Vb_{t}-\Ab_{t}\right)\Ab_{t}^{-1/2}\right\Vert _{2}\frac{4\sigma \tilde{\sigma}_{\max}}{p\tilde{\sigma}_{\min}}\sqrt{\frac{2(d+d_{h})\log(2Kt^2/\delta)}{1-\epsilon}} \notag \\
&\le \max_{i\in[K]}\left\Vert e_{i}^{\top}\left(\Vb_{t}-\Ab_{t}\right)\Vb_{t}^{-1/2}\right\Vert _{2}\frac{4\sigma \tilde{\sigma}_{\max}}{p(1-\epsilon)\tilde{\sigma}_{\min}}\sqrt{2(d+d_{h})\log\frac{2Kt^{2}}{\delta}}. \label{eq:est_1st_term_2}
\end{align}
For the first term in~\cref{eq:est_1st_term_2}, we have that
\begin{align*}
\max_{i\in[K]}\left\Vert \eb_{i}^{\top}\left(\Vb_{t}-\Ab_{t}\right)\Vb_{t}^{-1/2}\right\Vert _{2} 
&= \max_{i\in[K]}\left\Vert \eb_{i}^{\top}\Vb_t^{1/2} \Vb_t^{-1/2}\left(\Vb_{t}-\Ab_{t}\right)\Vb_{t}^{-1/2}\right\Vert _{2}\\
&\le \tilde{\sigma}_{\max}\sqrt{t}  \left \Vert \Ib_K- \Vb_{t}^{-1/2}\Ab_{t}\Vb_{t}^{-1/2}\right\Vert_{2} \\
& \le \tilde{\sigma}_{\max}\sqrt{t}\epsilon,
\end{align*}
where the last inequality holds due to \cref{eq:lasso_Gram_bound}.
Combining this result with~\cref{eq:est_1st_term_2}, we obtain that
\begin{equation*}
\left\Vert \left(\Vb_{t}-\Ab_{t}\right)\left(\check{\mub}_{t}^{L}-\mub_{\star}\right)\right\Vert _{\infty} \le \frac{4\epsilon \sigma \tilde{\sigma}^2_{\max}}{p(1-\epsilon)\tilde{\sigma}_{\min}}\sqrt{2t(d+d_{h})\log\frac{2Kt^{2}}{\delta}}.
\end{equation*}
Setting $\epsilon = K^{-1/2} \tilde{\sigma}_{\min} /8\tilde{\sigma}_{\max}$ yields that $1-\epsilon \ge 1/2$ and that
\begin{align*}
\left\Vert \left(\Vb_{t}-\Ab_{t}\right)\left(\check{\mub}_{t}^{L}-\mub_{\star}\right)\right\Vert _{\infty} &\le \frac{\sigma \tilde{\sigma}_{\max}}{p}\sqrt{2t\frac{d+d_{h}}{K}\log\frac{2Kt^{2}}{\delta}} \\
&\le \frac{\sigma \tilde{\sigma}_{\max}}{p}\sqrt{2t\log\frac{2Kt^{2}}{\delta}}.
\end{align*}
Now we conclude that for $t \ge 8^3 K^3 \tilde{\sigma}^{-2}_{\min} \tilde{\sigma}_{\max}^2 (1-p)^{-2} \log(2Kt^2/\delta)$,
\begin{equation*}
    \left\Vert \sum _{\tau=1}^{t} \sum_{a\in[K]} (\tilde{y}_{a,\tau} - \tilde{\xb}_{a}^\top  \mub_{t}) \tilde{\xb}_{a} \right\Vert_{\infty} \le \frac{\sigma \tilde{\sigma}_{\max}}{p}\sqrt{2t\log\frac{2Kt^{2}}{\delta}},
\end{equation*}
by~\cref{lem:lasso} and taking a union bound on the both terms in~\cref{eq:bound_pseudo_martingale}, with probability at least $1-2\delta/t^2$,
\begin{equation*}
    \Vert\widehat{\mub}_t^{L} - \mub_{\star}\Vert_{\Vb_t} \le \frac{8\sigma \tilde{\sigma}_{\max}}{p} \sqrt{\frac{2t(d+d_{h})\log(2Kt^2/\delta)}{\lambda_{\min}\left(\Vb_{t}\right)}}=\frac{8\sigma \tilde{\sigma}_{\max}}{p\tilde{\sigma}_{\min}} \sqrt{2(d+d_{h})\log\frac{2Kt^{2}}{\delta}},
\end{equation*}
which completes the proof.  \hfill\qed

\subsection{Proof of~\cref{thm:lasso_regret}}
\label{appendix:proof_thm3}
Since we have shown that the reward defined in~\cref{eq:reward_definition} is equivalent to its form obtained via the projection and augmentation strategy in~\cref{eq:reward_decomposition_expanded}, we henceforth calculate the regret bound based on~\cref{eq:reward_decomposition_expanded}.
Under the assumption that the expected reward is bounded by 1~(\cref{subsec:problem_formulation}), the instantaneous regret $\EE_{t-1}[y_{\star,t}]-\EE_{t-1}[y_{a_t,t}]$ is bounded above by $2$ for any $t\in[T]$, and that the number of rounds for the exploration phase satisfies $|\Ecal_{T}|\le8^3 K^3(1-p)^{-2}\log(2KT^2/\delta)$.
Given these bounds, the cumulative regret is bounded as:
\begin{align}
\regret(T)
&= \sum_{t=1}^T (\EE_{t-1}[y_{\star,t}]-\EE_{t-1}[y_{a_t,t}]) \notag \\
&= \sum_{t\in \Ecal_T} (\EE_{t-1}[y_{\star,t}]-\EE_{t-1}[y_{a_t,t}]) + \sum_{t\in[T]\setminus \Ecal_T} (\EE_{t-1}[y_{\star,t}]-\EE_{t-1}[y_{a_t,t}]) \notag \\
&\le 2\cdot 8^3 K^3(1-p)^{-2}\log\frac{2KT^{2}}{\delta}+\sum_{t\in [T]\setminus \Ecal_T} (\EE_{t-1}[y_{\star,t}]-\EE_{t-1}[y_{a_t,t}]) \notag \\
&\begin{aligned}
& = 2\cdot 8^3 K^3(1-p)^{-2}\log\frac{2KT^{2}}{\delta}+\sum_{t\in [T]\setminus \Ecal_T}\left\{ \II(a_{t}=\widehat{a}_{t})(\EE_{t-1}[y_{\star,t}]-\EE_{t-1}[y_{a_t,t}])\right\} \\
&\quad +\sum_{t\in [T]\setminus \Ecal_T}\left\{ \II(a_t\neq\widehat{a}_t)(\EE_{t-1}[y_{\star,t}]-\EE_{t-1}[y_{a_t,t}])\right\} \label{eq:lasso_regret_decomposition}.
\end{aligned}
\end{align}
We first consider the second term of~\cref{eq:lasso_regret_decomposition}.
By \cref{thm:est_lasso_consistency}, on the event $\{a_{t}=\widehat{a}_{t}\}$, 
\begin{align*}
\EE_{t-1}[y_{\star,t}]-\EE_{t-1}[y_{a_{t},t}]
&= \tilde{\xb}_{a_{\star}}^{\top}{\mub}_{\star}-\tilde{\xb}_{\widehat{a}_{t}}^{\top}{\mub}_{\star}\\
&= \tilde{\xb}_{a_\star}^\top \mub_\star + \tilde{\xb}_{a_\star}^\top \widehat{\mub}_{t-1}^{L} - \tilde{\xb}_{a_\star}^\top \widehat{\mub}_{t-1}^{L} + \tilde{\xb}_{\widehat{a}_t}^\top \widehat{\mub}_{t-1}^{L} - \tilde{\xb}_{\widehat{a}_t}^\top \widehat{\mub}_{t-1}^{L} -\tilde{\xb}_{\widehat{a}_{t}}^{\top}{\mub}_{\star}\\
&\le \left\vert\tilde{\xb}_{a_\star}^\top(\mub_\star - \widehat{\mub}_{t-1}^{L})\right\vert + \left\vert{\tilde{\xb}_{\widehat{a}_t}^\top(\mub_\star-\widehat{\mub}_{t-1}^{L})}\right\vert +\tilde{\xb}_{a_{\star}}^{\top}\widehat{\mub}_{t-1}^{L}-\tilde{\xb}_{\widehat{a}_{t}}^{\top}\widehat{\mub}_{t-1}^{L} \\
&\le 2\max_{a\in[K]}\left\vert\tilde{\xb}_{a}^{\top}({\mub}_{\star}-\widehat{\mub}_{t-1}^{L})\right\vert+\tilde{\xb}_{a_{\star}}^{\top}\widehat{\mub}_{t-1}^{L}-\tilde{\xb}_{\widehat{a}_{t}}^{\top}\widehat{\mub}_{t-1}^{L}\\
&\le 2\max_{a\in[K]}\left\vert\tilde{\xb}_{a}^{\top}({\mub}_{\star}-\widehat{\mub}_{t-1}^{L})\right\vert\\
&\le  \frac{16\sigma\tilde{\sigma}_{\max}}{p\tilde{\sigma}_{\min}}\sqrt{\frac{2(d+d_{h})}{t}\log\frac{2Kt^{2}}{\delta}},
\end{align*}
with probability at least $1-2\delta/t^2$ for each $t\in [T]\setminus \Ecal_T$.
Summing over $t$ and applying a union bound, we obtain that, with probability at least $1-4\delta$,
\begin{equation}
 \sum_{t\in[T]\setminus\Ecal_{T}}\left\{ \II(a_{t}=\widehat{a}_{t})\left(\EE_{t-1}[y_{\star,t}]-\EE_{t-1}[y_{a_{t},t}]\right)\right\} \le\frac{32\sigma \tilde{\sigma}_{\max}}{p \tilde{\sigma}_{\min}} \sqrt{2(d+d_{h})T\log\frac{2KT^{2}}{\delta}}.
 \label{eq:lasso_regret_second_term}
\end{equation}
For the last term of~\cref{eq:lasso_regret_decomposition},
\begin{align}
    &\sum_{t\in [T]\setminus \Ecal_T}\left\{ \II(a_t\neq\widehat{a}_t)\left(\EE_{t-1}[y_{\star,t}]-\EE_{t-1}[y_{a_{t},t}]\right)\right\} \notag \\
    &\le 2\left[\sum_{t\in [T]} \II(a_t\neq\widehat{a}_t)-\PP\left(a_{t}\neq\widehat{a}_{t}\right) + \PP\left(a_{t}\neq\widehat{a}_{t}\right)\right] \qquad(\because \EE_{t-1}[y_{\star,t}]-\EE_{t-1}[y_{a_{t},t}] \le 2) \notag \\
    & \le 2 \sqrt{2T \log \frac{2}{\delta}} %\Bigg(\sw{\text{Shouldn't this be } \sqrt{2T\log \frac{2}{\delta}}}\Bigg) 
    + \frac{4 \sqrt{T}}{K-1} + 4\delta, \label{eq:lasso_regret_final_term}
\end{align}
where the first term in~\cref{eq:lasso_regret_final_term} holds with probability at least $1-\delta$ by Hoeffding's inequality, and the remaining terms follow from~\cref{lem:exploration_prob_bound}.
Taking a union bound over~\cref{eq:lasso_regret_second_term},~\cref{eq:lasso_regret_final_term}, and $\Mcal_t$, we obtain, with probability at least $1-6\delta$,
\begin{align*}
\regret(T) 
& \le 2\cdot 8^3 K^3(1-p)^{-2}\log\frac{2KT^{2}}{\delta}+\frac{4\sqrt{T}}{K-1} \\
& \quad+ 2{\sqrt{2T \log \frac{2}{\delta}}} + 4\delta +\frac{32\sigma \tilde{\sigma}_{\max}}{p \tilde{\sigma}_{\min}} \sqrt{2(d+d_{h})T\log\frac{2KT^{2}}{\delta}},
\end{align*}
which concludes the proof. \hfill\qed

\subsection{Proof of \cref{thm:est_ridge_consistency}}
\label{appendix:proof_thm4}
Let $\tilde{\Vb}_{t}:=\sum_{\tau=1}^{t}\II(\mathcal{M}_{\tau})\sum_{a\in[K]}\mathbf{\tilde{x}}_{a}\mathbf{\tilde{x}}_{a}^{\top}+\Ib_{K}$
and $\Vb_{t}:=\sum_{\tau=1}^{t}\sum_{a\in[K]}\mathbf{\tilde{x}}_{a}\mathbf{\tilde{x}}_{a}^{\top}+\Ib_{K}$.
By definition of $\widehat{\mub}_{t}^{R}$ presented in~\cref{eq:def_ridge_DR},
\begin{equation*}
\tilde{\xb}_{a}^{\top}(\widehat{\mub}_{t}^{R}-{\mub}_{\star})=\tilde{\xb}_{a}^{\top}\tilde{\Vb}_t^{-1}\left\{ \sum_{\tau=1}^{t}\II(\Mcal_{\tau})\sum_{a\in[K]}\tilde{\xb}_{a}\left(\tilde{y}_{a,\tau}-\tilde{\xb}_{a}^{\top}{\mub}_{\star}\right)-{\mub}_{\star}\right\} .
\end{equation*}
By definition of the pseudo-rewards,
\begin{equation*}
\tilde{y}_{a,\tau}-\tilde{\xb}_{a}^{\top}{\mub}_{\star}=\left(1-\frac{\II(\tilde{a}_{\tau}=a)}{\phi_{a,t}}\right)\tilde{\xb}_{a}^{\top}\left(\check{\mub}_t^{R}-{\mub}_{\star}\right)+\frac{\II(\tilde{a}_{\tau}=a)}{\phi_{a,\tau}}\epsilon_{\tau}.
\end{equation*}
Let $\tilde{\Ab}_{t}:=\sum_{\tau=1}^{t}\II(\mathcal{M}_{\tau})\sum_{a\in[K]}\frac{\II(\tilde{a}_{\tau}=a)}{\phi_{a,t}}\mathbf{\tilde{x}}_{a}\mathbf{\tilde{x}}_{a}^{\top}+\Ib_{K}$
and $\Ab_{t}:=\sum_{\tau=1}^{t}\sum_{a\in[K]}\frac{\II(\tilde{a}_{\tau}=a)}{\phi_{a,t}}\mathbf{\tilde{x}}_{a}\mathbf{\tilde{x}}_{a}^{\top}+\Ib_{K}$
Then,
\begin{equation}
\tilde{\xb}_{a}^{\top}(\widehat{\mub}_{t}^R-{\mub}_{\star})=\tilde{\xb}_{a}^{\top}\tilde{\Vb}_{t}^{-1}\left\{ \left(\tilde{\Vb}_{t}-\tilde{\Ab}_{t}\right)\left(\check{\mub}_t^{R}-{\mub}_{\star}\right)+\sum_{\tau=1}^{t}\II(\Mcal_{\tau})\sum_{a\in[K]}\frac{\II(\tilde{a}_{\tau}=a)}{\phi_{a,\tau}}\tilde{\xb}_{a}\epsilon_{\tau}-{\mub}_{\star}\right\}.
\label{eq:est_err_1}
\end{equation}
By definition of the imputation estimator $\check{\mub}_t$,
%\tilde{\xb}_{a_{\tau}}\tilde{\xb}_{a_{\tau}}^{\top}
\begin{align*}
\check{\mub}_t^{R}-{\mub}_{\star}
&= \left(\sum_{\tau=1}^{t}\tilde{\xb}_{a_\tau}\tilde{\xb}_{a_\tau}^\top+p\Ib_{K}\right)^{-1} \left(\sum_{\tau=1}^{t}\tilde{\xb}_{a_\tau}\epsilon_{\tau}-p{\mub}_{\star}\right) \\
&=\left(\sum_{\tau=1}^{t}\frac{1}{\phi_{a_{\tau},\tau}}\tilde{\xb}_{a_\tau}\tilde{\xb}_{a_\tau}^{\top}+\Ib_{K}\right)^{-1}\left(\sum_{\tau=1}^{t}\frac{1}{\phi_{a_{\tau},\tau}}\tilde{\xb}_{a_\tau}\epsilon_{\tau}-{\mub}_{\star}\right)\\
&=\left(\sum_{\tau=1}^{t}\sum_{a\in[K]}\frac{\II(\tilde{a}_{\tau}=a)}{\phi_{a,\tau}}\tilde{\xb}_{a_\tau}\tilde{\xb}_{a_\tau}^{\top}+\Ib_{K}\right)^{-1}\left(\sum_{\tau=1}^{t}\frac{1}{p}\tilde{\xb}_{a_\tau}\epsilon_{\tau}-{\mub}_{\star}\right),
\end{align*}
%(\sw{To guarantee equality for the third equation, we need the coupling event $\cap_{\tau=1}^t \Mcal_\tau$})
where the second equality holds because $\phi_{a_{\tau},\tau}=p$ and the coupling event $\cap_{\tau=1}^{t}\Mcal_\tau$.
Thus,
\begin{equation*}
\check{\mub}_t^{R}-{\mub}_{\star} = \Ab_t^{-1}\left(\sum_{\tau=1}^{t}\frac{1}{p}\tilde{\xb}_{a_\tau}\epsilon_{\tau}-{\mub}_{\star}\right).
\end{equation*}
%Under the coupling event $\cap_{\tau=1}^{t}\Mcal_{\tau}$, (\sw{do we have to consider the probability of this event?})
Plugging in~\eqref{eq:est_err_1},
\[
\tilde{\xb}_{a}^{\top}(\widehat{\mub}_{t}^R-{\mub}_{\star})
= \tilde{\xb}_{a}^{\top}\tilde{\Vb}_{t}^{-1}\left\{ \left(\tilde{\Vb}_{t}-\tilde{\Ab}_{t}\right)\Ab_{t}^{-1}\left(\sum_{\tau=1}^{t}\frac{1}{p}\tilde{\xb}_{a_{\tau}}\epsilon_{\tau}-{\mub}_{\star}\right)+\sum_{\tau=1}^{t}\II(\Mcal_{\tau})\sum_{a\in[K]}\frac{\II(\tilde{a}_{\tau}=a)}{\phi_{a,\tau}}\tilde{\xb}_{a}\epsilon_{\tau}-{\mub}_{\star}\right\} 
\]
Under the coupling event, $\tilde{\Ab}_t=\Ab_t$, $\tilde{\Vb}_t=\Vb_t$ and $\sum_{a\in[K]}\phi_{a,\tau}^{-1}\II(\tilde{a}_\tau=a)\tilde{\xb}_a = {\tilde{\xb}_{a_\tau}}/{p}$. 
It follows that
\begin{align*}
\tilde{\xb}_{a}^{\top}(\widehat{\mub}_{t}^R-{\mub}_{\star})
& =\tilde{\xb}_{a}^{\top}\Vb_{t}^{-1}\left\{ \left(\Vb_{t}-\Ab_{t}\right)\Ab_{t}^{-1}+\Ib_{K}\right\} \left(\sum_{\tau=1}^{t}\frac{1}{p}\tilde{\xb}_{a_{\tau}}\epsilon_{\tau}-{\mub}_{\star}\right)\\
& =\tilde{\xb}_{a}^{\top}\Vb_{t}^{-1/2}\left(\Vb_{t}^{1/2}\Ab_{t}^{-1}\Vb_{t}^{1/2}\right)\Vb_{t}^{-1/2}\left(\sum_{\tau=1}^{t}\frac{1}{p}\tilde{\xb}_{a_{\tau}}\epsilon_{\tau}-{\mub}_{\star}\right).
\end{align*}
Taking absolute value on both sides, by Cauchy-Schwarz inequality,
\begin{equation}
\max_{a\in[K]}\abs{\tilde{\xb}_{a}^{\top}(\widehat{\mub}_{t}^R-{\mub}_{\star})}\le\max_{a\in[K]}\|\tilde{\xb}_{a}\|_{\Vb_{t}^{-1}}\|\Vb_{t}^{1/2}\Ab_{t}^{-1}\Vb_{t}^{1/2}\|_{2}\left\Vert \sum_{\tau=1}^{t}\frac{1}{p}\tilde{\xb}_{a_{\tau}}\epsilon_{\tau}-{\mub}_{\star}\right\Vert _{\Vb_{t}^{-1}}.
\label{eq:ridge_consistency_bound}
\end{equation}
\cref{cor:Gram_matrix} implies that, for $t\ge 8\epsilon^{-2}(1-p)^{-2}K^2\log(2Kt^2/\delta)$, with $\epsilon$ to be determined later, \[
\Ib_{K}-\Vb_{t}^{-1/2}\Ab_{t}\Vb_{t}^{-1/2}\preceq\epsilon\Ib_{K},
\]
with probability at least $1-\delta$ for all $t\ge 1$.
Rearranging terms gives
\begin{equation*}
\Vb_{t}^{1/2}\Ab_{t}^{-1}\Vb_{t}^{1/2}\preceq(1-\epsilon)^{-1}\Ib_{K}.
\end{equation*}
Combining this with~\cref{eq:ridge_consistency_bound},
\begin{align*}
\max_{a\in[K]}\abs{\tilde{\xb}_{a}^{\top}(\widehat{\mub}_{t}^R-{\mub}_{\star})}&\le \frac{\max_{a\in[K]}\|\tilde{\xb}_{a}\|_{\Vb_{t}^{-1}}}{1-\epsilon}\left\Vert \sum_{\tau=1}^{t}\frac{1}{p}\tilde{\xb}_{a_{\tau}}\epsilon_{\tau}-{\mub}_{\star}\right\Vert _{\Vb_{t}^{-1}}\\
&\le \frac{\max_{a\in[K]}\|\tilde{\xb}_{a}\|_{\Vb_{t}^{-1}}}{1-\epsilon}\left(\frac{1}{p}\left\Vert \sum_{\tau=1}^{t}\tilde{\xb}_{a_{\tau}}\epsilon_{\tau}\right\Vert _{\Vb_{t}^{-1}}+\left\Vert {\mub}_{\star}\right\Vert _{\Vb_{t}^{-1}}\right),
\end{align*}
where the last inequality follows from the triangle inequality.
Note that $\Vb_t$ is deterministic, both $\Vb_t$ and $\sum_{\tau=1}^{t} \tilde{\xb}_{a_\tau} \tilde{\xb}_{a_\tau}^\top + \Ib_K$ are positive definite, and $\Vb_t \succeq \sum_{\tau=1}^{t} \tilde{\xb}_{a_\tau} \tilde{\xb}_{a_\tau}^\top + \Ib_K$.
Then, by Lemma 9 of~\citep{abbasi-yadkori2011improved},
with probability at least $1-\delta$,
\begin{align*}
\left\Vert \sum_{\tau=1}^{t}\tilde{\xb}_{a_{\tau}}\epsilon_{\tau}\right\Vert _{\Vb_{t}^{-1}}&\le \left\Vert \sum_{\tau=1}^{t}\tilde{\xb}_{a_{\tau}}\epsilon_{\tau}\right\Vert _{\left(\sum_{\tau=1}^{t}\tilde{\xb}_{a_{\tau}}\tilde{\xb}_{a_{\tau}}^{\top}+\Ib_{K}\right)^{-1}}\\
&\le \sigma\sqrt{2\log\frac{\det(\sum_{\tau=1}^{t}\tilde{\xb}_{a_{\tau}}\tilde{\xb}_{a_{\tau}}^{\top}+\Ib_{K}){}^{1/2}}{\delta}}\\ %\left(\sw{\sigma\sqrt{2\log\frac{t\det(\sum_{\tau=1}^{t}\tilde{\xb}_{a_{\tau}}\tilde{\xb}_{a_{\tau}}^{\top}+\Ib_{K})^{1/2}}{\delta}}}\right)\\
&\le \sigma\sqrt{\log\frac{\det(\sum_{\tau=1}^{t}\tilde{\xb}_{a_{\tau}}\tilde{\xb}_{a_{\tau}}^{\top}+\Ib_{K})}{\delta}}
%\left(\sw{\sigma\sqrt{\log\frac{t\det(\sum_{\tau=1}^{t}\tilde{\xb}_{a_{\tau}}\tilde{\xb}_{a_{\tau}}^{\top}+\Ib_{K})}{\delta}}}\right).
\end{align*}
Applying Lemma 10 of~\citet{abbasi-yadkori2011improved} yields
\begin{align*}
\det\left(\sum_{\tau=1}^{t}\tilde{\xb}_{a_{\tau}}\tilde{\xb}_{a_{\tau}}^{\top}+\Ib_{K}\right)&\le \left\{ \frac{\text{Tr}\left(\sum_{\tau=1}^{t}\tilde{\xb}_{a_{\tau}}\tilde{\xb}_{a_{\tau}}^{\top}\right)+K}{K}\right\} ^{K}\\
&\le \left\{ \frac{t\max_{a\in[K]}\|\tilde{\xb}_{a_{\tau}}\|_{2}+K}{K}\right\} ^{K}\\
&\le \left\{ t+1\right\} ^{K},
\end{align*}
for all $t \ge 1$, where the last inequality holds by $\|\tilde{\xb}_{a_{\tau}}\|_{2}\le\sqrt{K}\|\tilde{\xb}_{a_{\tau}}\|_{\infty}\le K$.
Hence,
\begin{equation*}
\left\Vert \sum_{\tau=1}^{t}\tilde{\xb}_{a_{\tau}}\epsilon_{\tau}\right\Vert _{\Vb_{t}^{-1}}
%\sw{\le\left(\sigma\sqrt{\frac{t(t+1)^K}{\delta}}\right)}
\le\sigma\sqrt{K\log\frac{t+1}{\delta}},
\end{equation*}
which follows that,
\begin{align*}
\max_{a\in[K]}\abs{\tilde{\xb}_{a}^{\top}(\widehat{\mub}_{t}^R-{\mub}_{\star})}&\le \frac{\max_{a\in[K]}\|\tilde{\xb}_{a}\|_{\Vb_{t}^{-1}}}{1-\epsilon}\left(\frac{\sigma}{p}\sqrt{K\log\frac{t+1}{\delta}}+\left\Vert {\mub}_{\star}\right\Vert _{\Vb_{t}^{-1}}\right)\\
&\le \frac{1}{\sqrt{t}}\cdot\frac{1}{1-\epsilon}\left(\frac{\sigma}{p}\sqrt{K\log\frac{t+1}{\delta}}+\left\Vert {\mub}_{\star}\right\Vert _{\Vb_{t}^{-1}}\right).
\end{align*}
Since $\Vert {\mub}_{\star}\Vert_{\Vb_{t}^{-1}} \le \Vert {\mub}_{\star}\Vert_{2} \le \sqrt{K}$, choosing $\epsilon=1/2$ completes the proof.\hfill\qed

\subsection{Proof of \cref{thm:ridge_regret}}
\label{appendix:proof_thm5}
Similar to the proof of~\cref{thm:lasso_regret}~(\cref{appendix:proof_thm3}), the instantaneous regret is bounded above by $2$ for any $t\in[T]$, and the number of rounds for the exploration phase satisfies $\vert\Ecal_{T}\vert \le 32(1-p)^{-2}K^{2}\log({2dT^{2}}/{\delta})$. Consequently, the cumulative regret is bounded above as follows:
\begin{align}
\regret(T)
&\le 32(1-p)^{-2}K^{2}\log\frac{2dT^{2}}{\delta}+\sum_{t\in[T]\setminus\Ecal_{T}}\EE_{t-1}[y_{\star,t}]-\EE_{t-1}[y_{a_{t},t}] \notag \\
&\begin{aligned}
&= 32(1-p)^{-2}K^{2}\log\frac{2dT^{2}}{\delta}+\sum_{t\in[T]\setminus\Ecal_{T}}\left\{ \II(a_{t}=\widehat{a}_{t})\left(\EE_{t-1}[y_{\star,t}]-\EE_{t-1}[y_{a_{t},t}]\right)\right\} \\
& \quad+\sum_{t\in[T]\setminus\Ecal_{T}}\left\{ \II(a_t\neq\widehat{a}_t)\left(\EE_{t-1}[y_{\star,t}]-\EE_{t-1}[y_{a_{t},t}]\right)\right\}
\label{eq:ridge_regret_decomposition}.
\end{aligned}
\end{align}
We first consider the second term in~\cref{eq:ridge_regret_decomposition}.
On the event $\{a_{t}=\widehat{a}_{t}\}$, 
\begin{align*}
\EE_{t-1}[y_{\star,t}]-\EE_{t-1}[y_{a_{t},t}]&= \tilde{\xb}_{a_{\star}}^{\top}{\mub}_{\star}-\tilde{\xb}_{\widehat{a}_{t}}^{\top}{\mub}_{\star} \\
&= \tilde{\xb}_{a_\star}^\top \mub_\star + \tilde{\xb}_{a_\star}^\top \widehat{\mub}_{t-1}^{R} - \tilde{\xb}_{a_\star}^\top \widehat{\mub}_{t-1}^{R} + \tilde{\xb}_{\widehat{a}_t}^\top \widehat{\mub}_{t-1}^{R} - \tilde{\xb}_{\widehat{a}_t}^\top \widehat{\mub}_{t-1}^{R} -\tilde{\xb}_{\widehat{a}_{t}}^{\top}{\mub}_{\star} \\
&\le \left\vert\tilde{\xb}_{a_\star}^\top(\mub_\star - \widehat{\mub}_{t-1}^{R})\right\vert + \left\vert{\tilde{\xb}_{\widehat{a}_t}^\top(\mub_\star-\widehat{\mub}_{t-1}^{R})}\right\vert +\tilde{\xb}_{a_{\star}}^{\top}\widehat{\mub}_{t-1}^{R}-\tilde{\xb}_{\widehat{a}_{t}}^{\top}\widehat{\mub}_{t-1}^{R} \\
&\le 2\max_{a\in[K]}\left\vert\tilde{\xb}_{a}^{\top}\left({\mub}_{\star}-\widehat{\mub}_{t-1}^{R}\right)\right\vert+\tilde{\xb}_{a_{\star}}^{\top}\widehat{\mub}_{t-1}^{R}-\tilde{\xb}_{\widehat{a}_{t}}^{\top}\widehat{\mub}_{t-1}^{R} \\
&\le 2\max_{a\in[K]}\left\vert\tilde{\xb}_{a}^{\top}\left({\mub}_{\star}-\widehat{\mub}_{t-1}^{R}\right)\right\vert \\
&\le \frac{4}{\sqrt{t}}\left(\frac{\sigma}{p}\sqrt{K\log\frac{t}{\delta}}%\sw{\left(\sqrt{(K+1)\log\frac{t}{\delta}}\right)}
+\sqrt{K}\right),
\end{align*}
with probability at least $1-3\delta$, for all $t$ such that $t \ge |\Ecal_t|$ (\cref{thm:est_ridge_consistency}). 
Summing over $t$ gives, with probability at least $1-3\delta$,
\begin{equation}
\sum_{t\in[T]\setminus\Ecal_{T}}\left\{ \II(a_{t}=\widehat{a}_{t})\left(\EE_{t-1}[y_{\star,t}]-\EE_{t-1}[y_{a_{t},t}]\right)\right\} \le 8\sqrt{KT}\left(\frac{\sigma}{p}\sqrt{\log\frac{T}{\delta}}+1\right).
\label{eq:ridge_regret_second_term}
\end{equation}
For the last term in~\cref{eq:ridge_regret_decomposition}, we have that
\begin{align}
    &\sum_{t\in [T]\setminus \Ecal_T}\left\{ \II(a_t\neq\widehat{a}_t)\left(\EE_{t-1}[y_{\star,t}]-\EE_{t-1}[y_{a_{t},t}]\right)\right\} \notag \\
    &\le 2\left[\sum_{t\in [T]} \II(a_t\neq\widehat{a}_t)-\PP\left(a_{t}\neq\widehat{a}_{t}\right) + \PP\left(a_{t}\neq\widehat{a}_{t}\right)\right] \notag \\
    & \le 2  \sqrt{2T \log \frac{2}{\delta}} + \frac{4 \sqrt{T}}{K-1} + 4\delta, \label{eq:ridge_regret_last_term}
\end{align}
where the first term in~\cref{eq:ridge_regret_last_term} holds with probability at least $1-\delta$ by Hoeffding's inequality, and the remaining terms follow from~\cref{lem:exploration_prob_bound}.
Finally, by taking a union bound over~\cref{eq:ridge_regret_second_term} and~\cref{eq:ridge_regret_last_term}, we obtain, with probability at least $1-4\delta$,
\begin{equation*}
\regret(T)\le \frac{32K^{2}}{(1-p)^{2}}\log\frac{2dT^{2}}{\delta}+2 \sqrt{2T \log \frac{2}{\delta}} + \frac{4 \sqrt{T}}{K-1} + 4\delta+8\sqrt{KT}\left(\frac{\sigma}{p}\sqrt{\log\frac{T}{\delta}}+1\right),
\end{equation*} 
which completes the proof. \hfill\qed

\section{Algorithm-Agnostic Lower Bound of Regret Ignoring Unobserved Features}
\label{appendix:regret_lower_bound_general}
In this section, extending our argument in~\cref{thm:regret_linear_lower_bound}, we show that there exists a problem instance such that linear bandit algorithms relying solely on observed features can incur regret that grows linearly in $T$.
We begin by formally defining such algorithms.

\begin{definition}[Policy dependent on observed features]
\label{def:obs_dependent_policy}
For each $t \in [T]$, let $\pi_t:\RR^{d}\times\RR^{t-1}\to[0,1]$ be a policy that maps an observed feature vector $\xb\in\{\xb_a: a \in [K]\}$, given past reward observations $\{y_{a_s,s}: s \in [t-1]\}$, to a probability of selection.
Then the policy $\pi_t$ is dependent \emph{only} on observed features if, for any $y_{a_1,1},\ldots,y_{a_{t-1},t-1}$, it holds that $\xb_1 = \xb_2$ implies $\pi_t(\xb_1|y_{a_1,1},\ldots,y_{a_{t-1},t-1})=\pi_t(\xb_2|y_{a_1,1},\ldots,y_{a_{t-1},t-1})$.
\end{definition}

For instance, the UCB and Thompson sampling-based policies for linear bandits (with observed features), considered in~\cref{thm:regret_linear_lower_bound}, satisfy~\cref{def:obs_dependent_policy}, as they assign the same selection probability as long as the observed features are the same.
In contrast, policies in MAB algorithms (which disregard observed features) may assign different selection probabilities even when the observed features are equal, and thus are not dependent on the observed features.
In the theorem below, we particularly provide a lower bound for algorithms that employ policies that are dependent on the observed features.

\begin{theorem}[Regret Lower Bound under Policies Dependent on Observed Features]
\label{thm:regret_lower_bound_general}
    % Suppose that $ T \ge 4d^2 $. 
%We consider a set $\Xcal := \{\pm d/\sqrt{T} \}^d$ where observable features are generated.
%Furthermore, the parameter that determines the observable portion of the reward is defined as $\thetao = (1/3d, \dots, 1/3d)^\top\in\RR^d$.
%Let $a_{\star}$ denote the action that maximizes the expected reward based on the full feature vector, as defined in~\cref{eq:reward_decomposition}, while $ a_o $ represents the action that maximizes the observable portion of the expected reward, i.e., $a_o:=\argmax_{a\in\Acal}\dotp{\xb_a}{\thetao}$.
%In this scenario, using linear bandit algorithms that only consider observable features, the lower bound of the cumulative regret over the total horizon is given by:
For any algorithm $\Pi:=(\pi_1,\ldots,\pi_T)$ that consists of the policies $\{\pi_t:t\in[T]\}$ that are dependent on observed features, there exists a set of features $\{\zb_{1},\ldots,\zb_{K}\}$ and a parameter $\thetab_{\star} \in \RR^{d_z}$ such that the cumulative regret
\begin{equation*}
    \regret_{\Pi}(T, \thetab_{\star},\zb_1,\ldots,\zb_K) \ge \frac{T}{6}. %+ \Omega(d\sqrt{T}).
\end{equation*}
\end{theorem}

\begin{proof}
We start the proof by providing a detailed account of the scenario described in the theorem. 
Without loss of generality, we consider the case where $K = 3$.
As stated in the theorem, $ a_\star $ represents the index of the optimal action when considering the entire reward, including both observed and latent components. 
In contrast, $a_o$ denotes the index of the optimal action when considering only the observed components.
We introduce an additional notation, $ a' $, which refers to an action whose observed features are identical to those of $ a_\star $, but with a distinct latent component. 
Specifically, this implies that $ a' \ne a_\star $ and $\mathbf{z}_{a'} \ne \mathbf{z}_{a_\star} $, but $\mathbf{x}_{a'} = \mathbf{x}_{a_\star} $.
By definition of the policy $\pi_t$ that depends on the observed features, $\pi_t(\xb_{a_{\star}})=\pi_t(\xb_{a'})$ and the probability of selecting the optimal arm is $\pi_t(\xb_{a_{\star}})\le1/2$.

Taking this scenario into account, the observed part of the features associated with $a_\star$, $a'$, and $a_o$ are defined as follows:
\begin{equation*}
    \xb_{a_\star} := \left[-\frac{1}{2},\dots,-\frac{1}{2}\right]^\top, 
    \xb_{a'}:= \left[-\frac{1}{2}, \dots, -\frac{1}{2} \right]^\top, 
    \xb_{a_o}:= \left[\frac{1}{2},\dots, \frac{1}{2}\right]^\top.
\end{equation*}
Additionally, we define the unobserved feature vectors for actions $a_\star$, $a'$, and $a_o$ as follows:
\begin{equation*} 
\ub_{a_\star} := \left[1, \dots, 1\right]^\top, 
\ub_{a'} := \left[-1, \dots, -1\right]^\top, 
\ub_{a_o} := \left[-1, \dots, -1,1,\dots, 1\right]^\top, 
\end{equation*}
where in $\ub_{a_o}$, the number of 1's and -1's are equal.
% \wy{Since $T \ge 4d^2$, it can be observed that the supremum norms of $\zb_{a_\star}$, $\zb_{a'}$, and $\zb_{a_o}$ --- each constructed by concatenating the observable and corresponding unobserved parts --- do not exceed 1.}
This ensures that the scenario aligns with the assumption imposed on the feature vectors throughout this paper.
We further define the true parameter as follows: 
\begin{equation*} \thetab_{\star}:=\left[{\frac{1}{3d}},\dots,{\frac{1}{3d}}, {\frac{2}{3d_u}},\dots,{\frac{2}{3d_u}}\right]^\top\in\RR^{d_z},
\end{equation*}
thus it follows that $\thetao = [1/3d,\dots,1/3d]^\top\in\RR^{d}$ and $\thetau = [2/3d_u,\dots,2/3d_u]^\top\in\RR^{d_u}$.
Note that it is straightforward to verify that $|\langle \zb_a, \thetab_{\star} \rangle| \le 1$, thereby satisfying the assumption on the mean reward~(\cref{subsec:problem_formulation}).
With this established, we can also observe that the expected rewards for the three actions are defined as:
\begin{align*}
    &\dotp{\latent{\star}}{\thetab_{\star}} = \dotp{\obs{\star}}{\thetao} +\dotp{\ub_{a_\star}}{\thetau} = - \frac{1}{6} + \frac{2}{3} = \frac{1}{2}, \\
    &\dotp{{\zb}_{a'}}{\thetab_{\star}} = \dotp{\xb_{a'}}{\thetao} +\dotp{\ub_{a'}}{\thetau} = -\frac{1}{6} - \frac{2}{3} = -\frac{5}{6}, \\
    &\dotp{{\zb}_{a_o}}{\thetab_{\star}} = \dotp{\xb_{a_o}}{\thetao} + \dotp{\ub_{a_o}}{\thetau} = \frac{1}{6} + 0 = \frac{1}{6},
\end{align*}
respectively, and it is straightforward to verify that $\dotp{\latent{\star}}{\thetab_{\star}} - \dotp{{\zb}_{a_o}}{\thetab_{\star}} = 2/3 > 0$ and that $\dotp{\latent{\star}}{\thetab_{\star}} - \dotp{{\zb}_{a'}}{\thetab_{\star}} = 4/3 > 0$,
which confirms that $a_\star$ is optimal when considering the full feature set.

At each round $t\in[T]$ for any policy $\pi_t$ satisfying~\cref{def:obs_dependent_policy}, we have
\begin{equation*}
    \pi_t(\xb_{a_\star}|y_{a_1,1},\ldots,y_{a_{t-1},t-1}) = \pi_t(\xb_{a'}|y_{a_1,1},\ldots,y_{a_{t-1},t-1}),
\end{equation*}
which implies $\PP(a_t=a_{\star})=\PP(a_t=a')$ and
\[
\PP(a_t=a_\star) = 1 - \PP(a_t=a') - \PP(a_t=a_o) \le 1 - \PP(a_t=a') = 1 - \PP(a_t=a_{\star}),
\]
hence the probability of selecting an optimal arm cannot exceed $1/2$.
Thus, the expected regret,
\[
\regret_{\Pi}(T, \theta_{\star},\zb_{a_\star},\zb_{a'},\zb_{a_o}) \ge \Big(\frac{1}{2} - \frac{1}{6}\Big)\sum_{t=1}^T \PP(a_t \neq a_{\star}) \ge \frac{T}{6},
\]
which completes the proof.
\end{proof}

\section{Technical Lemmas}
\label{appendix:lemmas}
\begin{lemma}
\label{lem:exp_bound} (Exponential martingale inequality) If a martingale $(\Xb_{t};t\ge0)$, adapted to filtration $\Fcal_{t}$, satisfies $\condexp{\exp(\lambda \Xb_{t})}{\Fcal_{t-1}} \le\exp(\lambda^{2}\sigma_{t}^{2}/2)$
for some constant $\sigma_{t}$, for all $t$, then for any $a\ge0$,
\begin{equation*}
\PP\left(\abs{X_{T}-X_{0}}\ge a\right)\le2\exp\left(-\frac{a^{2}}{2\sum_{t=1}^{T}\sigma_{t}^{2}}\right).
\end{equation*}
Thus, with probability at least $1-\delta$,
\begin{equation*}
\abs{X_{T}-X_{0}}\le\sqrt{2\sum_{t=1}^{T}\sigma_{t}^{2}\log\frac{2}{\delta}}.
\end{equation*}
\end{lemma}

\subsection{A Hoeffding bound for Matrices}
\def\CE#1#2{\mathbb{E}\left[\left.#1\right\vert#2\right]}%
\def\CP#1#2{\mathbb{P}\left(\left.#1\right\vert#2\right)}%

\begin{lemma} 
\label{lem:matrix_hoeffding} 
Let $\{\Mb_{\tau}:\tau\in[t]\}$ be a $\mathbb{R}^{d\times d}$-valued stochastic process adapted to the filtration $\{\mathcal{F_{\tau}}:\tau\in[t]\}$, i.e., $\Mb_{\tau}$ is $\Fcal_{\tau}$-measurable for $\tau\in[t]$. 
Suppose that the matrix $\Mb_{\tau}$ is symmetric and the eigenvalues of the difference $\Mb_{\tau}-\condexp{\Mb_{\tau}}{\Fcal_{\tau-1}}$ lie in $[-b,b]$ for some $b>0$. 
Then for $x>0$, 
\begin{equation*}
\mathbb{P}\left(\left\Vert \sum_{\tau=1}^{t}\Mb_{\tau}-\condexp{\Mb_{\tau}}{\Fcal_{\tau-1}} \right\Vert _{2}\ge x\right)\le2d\exp\left(-\frac{x^{2}}{2tb^{2}}\right).
\end{equation*}

\end{lemma}
\begin{proof}
The proof adapts Hoeffding's inequality to a matrix stochastic process, following the argument of~\citet{tropp2012user-friendly}.
Let $\Db_{\tau}:=\Mb_{\tau}-\condexp{\Mb_{\tau}}{\Fcal_{\tau-1}}$. 
Then, for $x>0$, 
\begin{equation}
\mathbb{P}\left(\left\Vert \sum_{\tau=1}^{t}\Db_{\tau}\right\Vert _{2}\ge x\right)\le\mathbb{P}\left(\lambda_{\max}\left(\sum_{\tau=1}^{t}\Db_{\tau}\right)\ge x\right)+\mathbb{P}\left(\lambda_{\max}\left(-\sum_{\tau=1}^{t}\Db_{\tau}\right)\ge x\right).
\label{eq:hoeffding_matrix1}
\end{equation}
We bound the first term of~\cref{eq:hoeffding_matrix1}; a similar argument applies to the second term. 
For any $v>0$, 
\begin{equation*}
\mathbb{P}\left(\lambda_{\max}\left(\sum_{\tau=1}^{t}\Db_{\tau}\right)\ge x\right)\le\mathbb{P}\left(\exp\left\{ v\lambda_{\max}\left(\sum_{\tau=1}^{t}\Db_{\tau}\right)\right\} \ge e^{vx}\right)\le e^{-vx}\mathbb{E}\left[\exp\left\{ v\lambda_{\max}\left(\sum_{\tau=1}^{t}\Db_{\tau}\right)\right\} \right].
\end{equation*}
Since $\sum_{\tau=1}^{t}\Db_{\tau}$ is a real symmetric matrix, 
\begin{align*}
\exp\left\{ v\lambda_{\max}\left(\sum_{\tau=1}^{t}\Db_{\tau}\right)\right\} = & \lambda_{\max}\left\{ \exp\left(v\sum_{\tau=1}^{t}\Db_{\tau}\right)\right\} \le\text{Tr}\left\{ \exp\left(v\sum_{\tau=1}^{t}\Db_{\tau}\right)\right\} ,
\end{align*}
where the last inequality holds since $\exp(v\sum_{\tau=1}^{t}\Db_{\tau})$ has nonnegative eigenvalues. 
Taking expectations on both sides yields
\begin{align*}
\mathbb{E}\left[\exp\left\{ v\lambda_{\max}\left(\sum_{\tau=1}^{t}\Db_{\tau}\right)\right\} \right]
&\le  \mathbb{E}\left[\text{Tr}\left\{ \exp\left(v\sum_{\tau=1}^{t}\Db_{\tau}\right)\right\} \right]\\
&=  \text{Tr}\mathbb{E}\left[\exp\left(v\sum_{\tau=1}^{t}\Db_{\tau}\right)\right]\\
&= \text{Tr}\mathbb{E}\left[\exp\left(v\sum_{\tau=1}^{t-1}\Db_{\tau}+\log\exp(v\Db_{t})\right)\right].
\end{align*}
By Lieb's theorem~\citep{tropp2015introduction}, the map $\Db \mapsto \text{Tr}\exp(\Hb + \log \Db)$ is concave over the set of symmetric positive definite matrices for any symmetric positive definite $\Hb$.
Applying Jensen's inequality, we obtain

\begin{equation}
\label{eq:trace_jensen_ineq}
\text{Tr}\mathbb{E}\left[\exp\left(v\sum_{\tau=1}^{t-1}\Db_{\tau}+\log\exp(v\Db_{t})\right)\right]\le\text{Tr}\mathbb{E}\left[\exp\left(v\sum_{\tau=1}^{t-1}\Db_{\tau}+\log\CE{\exp(v\Db_{t})}{\Fcal_{t-1}}\right)\right].
\end{equation}

By convexity of $e^{vx}$ and Hoeffding's lemma, for all $x \in [-b,b]$,
\begin{equation*}
e^{vx}\le\frac{b-x}{2b}e^{-vb}+\frac{x+b}{2b}e^{vb}.
\end{equation*}
Since the eigenvalues of $\Db_{t}$ lie within $[-b,b]$, it follows that
\begin{align*}
\CE{\exp(v\Db_{t})}{\Fcal_{t-1}}
&\preceq \CE{\frac{e^{-vb}}{2b}\left(b\Ib_{d}-\Db_{t}\right)+\frac{e^{vb}}{2b}\left(\Db_{t}+b\Ib_{d}\right)}{\Fcal_{t-1}}\\
&= \frac{e^{-vb}+e^{vb}}{2}\Ib_{d}\\
&\preceq \exp(\frac{v^{2}b^{2}}{2})\Ib_{d}.
\end{align*}

Now we recursively upper bound~\cref{eq:trace_jensen_ineq} as follows:
\begin{align*}
\mathbb{E}\left[\exp\left\{ v\lambda_{\max}\left(\sum_{\tau=1}^{t}\Db_{\tau}\right)\right\} \right]&\le \text{Tr}\mathbb{E}\left[\exp\left(v\sum_{\tau=1}^{t-1}\Db_{\tau}+\log\CE{\exp(v\Db_{t})}{\Fcal_{t-1}}\right)\right]\\
&\le \text{Tr}\mathbb{E}\left[\exp\left(v\sum_{\tau=1}^{t-1}\Db_{\tau}+(\frac{v^{2}b^{2}}{2})\Ib_{d}\right)\right]\\
&\le \text{Tr}\mathbb{E}\left[\exp\left(v\sum_{\tau=1}^{t-2}\Db_{\tau}+(\frac{v^{2}b^{2}}{2})\Ib_{d}+\log\CE{\exp(v\Db_{t-1})}{\Fcal_{t-2}}\right)\right]\\
&\le \text{Tr}\mathbb{E}\left[\exp\left(v\sum_{\tau=1}^{t-2}\Db_{\tau}+(\frac{2v^{2}b^{2}}{2})\Ib_{d}\right)\right]\\
& \vdots \vdots\\
&\le \text{Tr}\exp\left((\frac{tv^{2}b^{2}}{2})\Ib_{d}\right)\\
&= \exp\left(\frac{tv^{2}b^{2}}{2}\right)\text{Tr}\left(\Ib_{d}\right)\\
&= d\exp\left(\frac{tv^{2}b^{2}}{2}\right).
\end{align*}
Thus we have 
\begin{equation*}
\mathbb{P}\left(\lambda_{\max}\left(\sum_{\tau=1}^{t}\Db_{\tau}\right)\ge x\right)\le d\exp\left(-vx+\frac{tv^{2}b^{2}}{2}\right).
\end{equation*}
Minimizing over $v>0$ gives $v=x/(tb^{2})$ and
\begin{equation*}
\mathbb{P}\left(\lambda_{\max}\left(\sum_{\tau=1}^{t}\Db_{\tau}\right)\ge x\right)\le d\exp\left(-\frac{x^{2}}{2tb^{2}}\right),
\end{equation*}
which proves the lemma.
\end{proof}

\subsection{A Bound for the Gram Matrix}
The Hoeffding bound for matrices (\cref{lem:matrix_hoeffding}) implies the following bound for the two Gram matrices $\Ab_t:=\sum_{\tau=1}^{t}\tilde{\xb}_{a_\tau}\tilde{\xb}_{a_\tau}^\top$ and $\Vb_t:=\sum_{\tau=1}^{t} \sum_{a\in[K]}\tilde{\xb}_{a}\tilde{\xb}_{a}^\top$
\begin{corollary}
\label{cor:Gram_matrix}
For any $\epsilon \in (0,1)$ and $t\ge8\epsilon^{-2}(1-p)^{-2}K^{2}\log({2Kt^{2}}/{\delta})$, with probability at least $1-\delta/t^{2},$
\begin{equation*}
\left\Vert\Ib_{K}-\Vb_{t}^{-1/2}\Ab_{t}\Vb_{t}^{-1/2}\right\Vert_2\le\epsilon.
\end{equation*}
\end{corollary}

\begin{proof}
Note that
\begin{equation*}
\Vb_{t}^{-1/2}\Ab_{t}\Vb_{t}^{-1/2}-\Ib_{K}=\Vb_{t}^{-1/2}\left\{ \sum_{\tau=1}^{t}\sum_{a\in[K]}\left(\frac{\II(\tilde{a}_{\tau}=a)}{\phi_{a,\tau}}-1\right)\mathbf{\tilde{x}}_{a}\mathbf{\tilde{x}}_{a}^{\top}\right\} \Vb_{t}^{-1/2},
\end{equation*}
and the martingale difference matrix for each $\tau\in[t]$,
\begin{align*}
\left\Vert \sum_{a\in[K]}\left(\frac{\II(\tilde{a}_{\tau}=a)}{\phi_{a,\tau}}-1\right)\Vb_{t}^{-1/2}\mathbf{\tilde{x}}_{a}\mathbf{\tilde{x}}_{a}^{\top}\Vb_{t}^{-1/2}\right\Vert _{2} \notag
&\le \left(\sum_{a\in[K]}\left\vert \frac{\II(\tilde{a}_\tau=a)}{\phi_{a,\tau}} -1 \right\vert\right) \max_{a\in[K]}\left\Vert \Vb_{t}^{-1/2}\mathbf{\tilde{x}}_{a}\mathbf{\tilde{x}}_{a}^{\top}\Vb_{t}^{-1/2}\right\Vert _{2}\\
&\le \left(\frac{K-1}{1-p}+K-2\right)\max_{a\in[K]}\left\Vert \Vb_{t}^{-1/2}\mathbf{\tilde{x}}_{a}\mathbf{\tilde{x}}_{a}^{\top}\Vb_{t}^{-1/2}\right\Vert _{2} \notag \\
&\le \frac{2K}{1-p}\max_{a\in[K]}\left\Vert \mathbf{\tilde{x}}_{a}\right\Vert _{\Vb_{t}^{-1}}^{2} \notag \\
&\le \frac{2K}{1-p}\cdot\frac{1}{t}. 
\end{align*}
Note that the second inequality holds under the assumption $p \in (1/2,1)$, which implies $\phi_{a,\tau}^{-1} \le (K-1)/(1-p)$, and the last inequality is obtained via the Sherman–Morrison formula.
By~\cref{lem:matrix_hoeffding}, for $x>0$, we have
\begin{equation*}
\PP\left(\left\Vert\Vb_{t}^{-1/2}\Ab_{t}\Vb_{t}^{-1/2}-\Ib_{K}\right\Vert_{2}>x\right)\le 2K\exp\left(-\frac{(1-p)^{2}tx^{2}}{8K^{2}}\right).
\end{equation*}
Setting $x=\epsilon\in(0,1)$, for
$t\ge8\epsilon^{-2}(1-p)^{-2}K^{2}\log({2Kt^{2}}/{\delta})$ with
probability at least $1-\delta/t^{2},$
\begin{equation*}
\left\Vert\Ib_{K}-\Vb_{t}^{-1/2}\Ab_{t}\Vb_{t}^{-1/2}\right\Vert_2\le\epsilon.
\end{equation*}
\end{proof}

\subsection{An error bound for the Lasso estimator}

\begin{lemma}[An error bound for the Lasso estimator with unrestricted minimum eigenvalue]
\label{lem:lasso}  
Let $\{\xb_{\tau}\}_{\tau \in [t]} \subset [-1,1]^{d}$ denote the covariates, and let $y_{\tau} = \xb_{\tau}^{\top} \bar{\wb} + e_{\tau}$ for some $\bar{\wb} \in \RR^{d}$ and $e_{\tau} \in \RR$.
For $\lambda > 0$, consider the estimator
\begin{equation*}
\widehat{\wb}_{t}=\argmin_{\wb}\sum_{\tau=1}^{t}\left(y_{\tau}-\xb_{\tau}^{\top}\wb\right)^{2}+\lambda\norm \wb_{1}.
\end{equation*}
Define $\bar{\Scal} := \{ i \in [d] : \bar{\wb}(i) \neq 0 \}$ and $\Sigmab_{t} := \sum_{\tau=1}^{t} \xb_{\tau} \xb_{\tau}^{\top}$.
Suppose that $\Sigmab_t$ has a strictly positive minimum eigenvalue and that $\left| \sum_{\tau=1}^{t} e_{\tau} \xb_{\tau} \right|_{\infty} \le \lambda/2$.
Then,
\begin{equation*}
\norm{\widehat{\wb}_{t}-\bar{\wb}}_{\Sigmab_t}\le\frac{2\lambda\sqrt{\abs{\bar{\Scal}}}}{\sqrt{\lambda_{\min}\left(\Sigmab_t\right)}}.
\end{equation*}
\end{lemma}

\begin{proof}
The proof follows a similar structure to Lemma B.4 in~\citet{kim2024doubly}, but we provide a new argument for the \emph{unrestricted} minimum eigenvalue condition.
Let $\Xb_t^\top := (\xb_1, \ldots, \xb_t) \in [-1,1]^{d \times t}$ and $\mathbf{e}_t^\top := (e_1, \ldots, e_t) \in \RR^t$.
We denote by $\Xb_t(j)$ the $j$-th column of $\Xb_t$ and by $\widehat{\wb}_t(j)$ the $j$-th entry of $\widehat{\wb}_t$.
By definition of $\widehat{\wb}_{t}$,
\begin{equation*}
\norm{\Xb_{t}\left(\bar{\wb}-\widehat{\wb}_{t}\right)+\mathbf{e}_{t}}_{2}^{2}+\lambda\norm{\widehat{\wb}_{t}}_{1}\le\norm{\mathbf{e}_{t}}_{2}^{2}+\lambda\norm{\bar{\wb}}_{1},
\end{equation*}
which implies
\begin{align*}
\norm{\Xb_{t}\left(\bar{\wb}-\widehat{\wb}_{t}\right)}_{2}^{2}+\lambda\norm{\widehat{\wb}_{t}}_{1}&\le 2\left(\widehat{\wb}_{t}-\bar{\wb}\right)^{\top}\Xb_{t}^{\top}\mathbf{e}_{t}+\lambda\norm{\bar{\wb}}_{1}\\
&\le 2\norm{\widehat{\wb}_{t}-\bar{\wb}}_{1}\norm{\Xb_{t}^{\top}\mathbf{e}_{t}}_{\infty}+\lambda\norm{\bar{\wb}}_{1}\\
&\le \lambda \norm{\widehat{\wb}_{t}-\bar{\wb}}_{1}+\lambda\norm{\bar{\wb}}_{1},
\end{align*}
where the last inequality uses the bound on $\lambda$. On the left
hand side, by triangle inequality,
\begin{align*}
\norm{\widehat{\wb}_{t}}_{1}& = \sum_{i\in\bar{\Scal}}\abs{\widehat{\wb}_{t}(i)}+\sum_{i\in[d]\setminus\bar{\Scal}}\abs{\widehat{\wb}_{t}(i)}\\
& \ge \sum_{i\in\bar{\Scal}}\abs{\widehat{\wb}_{t}(i)}-\sum_{i\in\bar{\Scal}}\abs{\widehat{\wb}_{t}(i)-\bar{\wb}(i)}+\sum_{i\in[d]\setminus\bar{\Scal}}\abs{\bar{\wb}(i)}\\
& = \norm{\bar{\wb}}_{1}-\sum_{i\in\bar{\Scal}}\abs{\widehat{\wb}_{t}(i)-\bar{\wb}(i)}+\sum_{i\in[d]\setminus\bar{\Scal}}\abs{\widehat{\wb}_{t}(i)},
\end{align*}
and for the right-hand side,
\begin{equation*}
\norm{\widehat{\wb}_{t}-\bar{\wb}}_{1}=\sum_{i\in\bar{\Scal}}\abs{\widehat{\wb}_{t}(i)-\bar{\wb}(i)}+\sum_{i\in[d]\setminus\bar{\Scal}}\abs{\widehat{\wb}_{t}(i)}.
\end{equation*}
Plugging in both sides and rearranging the terms,
\begin{equation}
\norm{\Xb_{t}\left(\bar{\wb}-\widehat{\wb}_{t}\right)}_{2}^{2} \le 2\lambda\sum_{i\in\bar{\Scal}}\abs{\widehat{\wb}_{t}(i)-\bar{\wb}(i)}.
\label{eq:w_basic}
\end{equation}
Because $\Xb_t^\top \Xb_t$ is positive definite,
\begin{align*}
\norm{\Xb_{t}\left(\bar{\wb}-\widehat{\wb}_{t}\right)}_{2}^{2}
& \ge \lambda_{\min}(\Xb_{t}^{\top}\Xb_{t})\sum_{i\in\bar{\Scal}}\abs{\widehat{\wb}_{t}(i)-\bar{\wb}(i)}^{2}\\
& \ge \frac{\lambda_{\min}(\Xb_{t}^{\top}\Xb_{t})}{\abs{\bar{\Scal}}}\left(\sum_{i\in\bar{\Scal}}\abs{\widehat{\wb}_{t}(i)-\bar{\wb}(i)}\right)^{2},
\end{align*}
where the last inequality holds by Cauchy-Schwarz inequality. Plugging in \cref{eq:w_basic} gives,
\begin{align*}
\norm{\Xb_{t}\left(\bar{\wb}-\widehat{\wb}_{t}\right)}_{2}^{2}
& \le2\lambda\sum_{i\in\bar{\Scal}}\abs{\widehat{\wb}_{t}(i)-\bar{\wb}(i)} \\
&\le2\lambda\sqrt{\frac{\abs{\bar{\Scal}}}{\lambda_{\min}(\Sigmab_{t})}}\norm{\Xb_{t}\left(\bar{\wb}-\widehat{\wb}_{t}\right)}_{2}\\
&\le \frac{2\lambda^{2}\abs{\bar{\Scal}}}{\lambda_{\min}(\Sigmab_{t})}+\frac{1}{2}\norm{\Xb_{t}\left(\bar{\wb}-\widehat{\wb}_{t}\right)}_{2}^{2},
\end{align*}
where the last inequality uses $ab\le a^{2}/2+b^{2}/2$. 
Rearranging the terms,
\begin{equation*}
\norm{\Xb_{t}\left(\bar{\wb}-\widehat{\wb}_{t}\right)}_{2}^{2}\le\frac{4\lambda^{2}\abs{\bar{\Scal}}}{\lambda_{\min}(\Sigmab_{t})},
\end{equation*}
which proves the result. 
\end{proof}

\subsection{Eigenvalue bounds for the Gram matrix.}

\begin{lemma} 
\label{lem:eig_bound} 
For $a\in[K]$, let $\tilde{\mathbf{x}}_{a}:=[\mathbf{x}_{a}^{\top},\mathbf{e}_{a}^{\top}\mathbf{p}_{1},\cdots,\mathbf{e}_{a}^{\top}\mathbf{p}_{K-d}]^{\top}\in\mathbb{R}^{d}$
denote augmented features. 
Then, an eigenvalue of $\sum_{a\in[K]}\tilde{\mathbf{x}}_{a}\tilde{\mathbf{x}}_{a}^{\top}$
is in the following interval
\begin{equation*}
    \left[\min\left\{\lambda_{\min}\left(\sum_{a\in[k]}\mathbf{x}_{a}\mathbf{x}_{a}^{\top}\right),1\right\},\max\left\{\lambda_{\max}\left(\sum_{a\in[K]}\mathbf{x}_{a}\mathbf{x}_{a}^{\top}\right),1\right\}\right].
\end{equation*}

\end{lemma} 
\begin{proof}
Let $\mathbf{P}:=(\mathbf{p}_{1},\ldots,\mathbf{p}_{K-d})\in\mathbb{R}^{K\times(K-d)}$.
Because the columns in $\mathbf{P}$ are orthogonal each other and
to $\mathbf{x}_{1},\ldots,\mathbf{x}_{K}$, 
\begin{align*}
\sum_{a\in[K]}\tilde{\mathbf{x}}_{a}\tilde{\mathbf{x}}_{a}^{\top}= & \begin{bmatrix}\sum_{a\in[K]}\mathbf{x}_{a}\mathbf{x}_{a}^{\top} & \sum_{a\in[K]}\mathbf{x}_{a}\mathbf{e}_{a}^{\top}\mathbf{P}\\
\sum_{a\in[K]}\mathbf{P}^{\top}\mathbf{e}_{a}\mathbf{x}_{a}^{\top} & \mathbf{P}^{\top}\mathbf{P}
\end{bmatrix}\\
= & \begin{bmatrix}\sum_{a\in[K]}\mathbf{x}_{a}\mathbf{x}_{a}^{\top} & \sum_{a\in[K]}\mathbf{x}_{a}\mathbf{e}_{a}^{\top}\mathbf{P}\\
\sum_{a\in[K]}\mathbf{P}^{\top}\mathbf{e}_{a}\mathbf{x}_{a}^{\top} & I_{K-d}
\end{bmatrix}\\
= & \begin{bmatrix}\sum_{a\in[K]}\mathbf{x}_{a}\mathbf{x}_{a}^{\top} & \sum_{a\in[K]}\mathbf{X}\mathbf{e}_{a}\mathbf{e}_{a}^{\top}\mathbf{P}\\
\sum_{a\in[K]}\mathbf{P}^{\top}\mathbf{e}_{a}\mathbf{e}_{a}^{\top}\mathbf{X} & I_{K-d}
\end{bmatrix}\\
= & \begin{bmatrix}\sum_{a\in[K]}\mathbf{x}_{a}\mathbf{x}_{a}^{\top} & \mathbf{X}\mathbf{P}\\
\mathbf{P}^{\top}\mathbf{X}^{\top} & I_{K-d}
\end{bmatrix}\\
= & \begin{bmatrix}\sum_{a\in[K]}\mathbf{x}_{a}\mathbf{x}_{a}^{\top} & \mathbf{O}\\
\mathbf{O} & I_{K-d}
\end{bmatrix}.
\end{align*}
Thus, for any $\lambda\in\RR$, $\det(\sum_{a\in[K]}\tilde{\mathbf{x}}_{a}\tilde{\mathbf{x}}_{a}^{\top}-\lambda\mathbf{I}_{K})=\det(\sum_{a\in[K]}\mathbf{x}_{a}\mathbf{x}_{a}^{\top}-\lambda\mathbf{I}_{d})(1-\lambda)^{K-d}$.
Solving $\det(\sum_{a\in[K]}\mathbf{x}_{a}\mathbf{x}_{a}^{\top}-\lambda\mathbf{I}_{d})(1-\lambda)^{K-d}=0$
gives the eigenvalues and the lemma is proved.
\end{proof}

\subsection{A Bound for the Probability of Exploration}
\begin{lemma}
\label{lem:exploration_prob_bound}
For each $t$, let $\widehat{a}_t = \argmax_{a\in[K]} \tilde{\xb}_a^\top \widehat{\mub}_t$ denote the maximizing action based on the estimator $\widehat{\mub}_t$.
Then the action $a_t$ chosen by the resampling scheme in \cref{alg:RoLF} and \cref{alg:RoLF_ridge} satisfies,
\[
\sum_{t=1}^{T}\PP(a_t \neq \widehat{a}_t) \le \frac{2\sqrt{T}}{(K-1)}+2\delta.
\]
\end{lemma}
\begin{proof}
Since the algorithm resamples at most $\rho_{t}$ times, for a fixed $t \in [T]$,
\begin{align}
\PP(a_t \ne \hat{a}_t) 
&= \PP\left(\{\tilde{a}_t = a_t\} \cap \{a_t \ne \hat{a}_t\}\right) + \PP\left(\{\tilde{a}_t \ne a_t\} \cap \{a_t \ne \hat{a}_t \}\right) \notag \\
&\le\PP\left(\{\tilde{a}_t = a_t\} \cap \{a_t \ne \hat{a}_t \}\right)  + \PP(\tilde{a}_t \ne a_t) \notag \\
&= \PP\left(\{ \tilde{a}_t = a_t\} \cap \{a_t = k\} \right) + \underbrace{\PP(\tilde{a}_t \ne a_t)}_{\text{Failure of resampling}} \quad \text{for } k \ne \hat{a}_t, \label{eq:lemma5_resample_term}
\end{align}
where $\mathbb{P}(a_t = k) = t^{-1/2}/(K - 1)$ and $\mathbb{P}(\tilde{a}_t = a_t) = p$, defined by~\cref{alg:RoLF} and~\cref{eq:pseudo_distribution}, respectively.
For the first term in~\cref{eq:lemma5_resample_term}, by applying the union bound, we obtain
\begin{align*}
\PP\left(\{ \tilde{a}_t = a_t\} \cap \{a_t = k\} \right)
&= \PP\left(\bigcup_{m=1}^{\rho_t} \{ \text{Resampling success at trial } m \} \cap \{a_t = k\} \right) \\
&\le \sum_{m=1}^{\rho_t} \PP\left(\{ \text{Resampling success at trial } m\} \cap \{a_t = k\} \right)
\\
&\le \PP(a_t=k) \\
&= \frac{1}{\sqrt{t}(K-1)},
%&= \sum_{m=1}^{\rho_t} \{ \text{resampling fail until } m-1 \cap \text{resampling success at } m-1 \cap (a_t = k) \}.
\end{align*}
which, combined with~\cref{eq:lemma5_resample_term}, gives
\begin{equation*}
    \PP(a_t \ne \hat{a}_t)  \le 
    \frac{1}{\sqrt{t}(K-1)} + \PP(\text{Resampling failure})
    %\sum_{m=1}^{\rho_t} \PP\left(\{ \text{Resampling success at trial } m\} \cap \{a_t = k\} \right) + \PP(\text{Resampling failure}).
\end{equation*}
By the definition of $\rho_t$, the probability that the resampling fails is bounded by $\delta/(t+1)^2$.
Thus, the probability of the event
$\{a_{t}\neq\widehat{a}_{t}\}$ is
\[
  \PP(a_t \ne \hat{a}_t)  \le 
    \frac{1}{\sqrt{t}(K-1)} + \frac{\delta}{(t+1)^2}
\]
%\begin{align*}
%\PP\left(a_{t}\neq\widehat{a}_{t}\right) \le & \sum_{m=1}^{\rho_{t}}\frac{p}{(K-1)\sqrt{t}} \cdot \PP(\text{Resample succeed at trial }m) +\PP(\text{Resample fails})\\
%\le& \sum_{m=1}^{\rho_{t}}\frac{p}{(K-1)\sqrt{t}}\cdot \PP(\text{Resample succeed at trial }m)+ \frac{\delta}{(t+1)^2}\\
%= & \frac{p}{(K-1)\sqrt{t}}\left(1-\frac{\delta}{(t+1)^2}\right) + \frac{\delta}{(t+1)^2} \\
%\le & \frac{p}{(K-1)\sqrt{t}}+ \frac{\delta}{(t+1)^2},
%\end{align*}
%where the equality holds by the definition of $\rho_t$.
Summing up over $t\in[T]$ completes the proof.
\end{proof}

%%%%%%%%%%%%%%%%%%%%%%%%%%%%%%%%%%%%%%%%%%%%%%%%%%%%%%%%%%%%%%%%%%%%%%%%%%%%%%%
%%%%%%%%%%%%%%%%%%%%%%%%%%%%%%%%%%%%%%%%%%%%%%%%%%%%%%%%%%%%%%%%%%%%%%%%%%%%%%%

\end{document}